%% file: main_ubicomp.tex
\documentclass[acmlarge]{acmart}
\AtBeginDocument{%
  }

\usepackage{xcolor}
\usepackage{xurl}
\usepackage{booktabs}
\usepackage[ruled]{algorithm2e}
\usepackage{algpseudocode}
\usepackage{enumerate}
\usepackage{enumitem}
\usepackage[english]{babel}
\usepackage{blindtext}
\usepackage[caption=false,font=footnotesize,subrefformat=parens]{subfig}
\usepackage{amsmath}

\usepackage{amssymb}
\usepackage{xspace}
\usepackage{multirow}
\usepackage{threeparttable}
\usepackage{float}
\usepackage{flafter}
\usepackage{comment}
\usepackage[skip=12pt]{caption}
\usepackage{graphicx}
\usepackage{array}
\usepackage{arydshln}
\usepackage{amsmath,amssymb,stmaryrd}   
\usepackage{arydshln,fp,xstring,siunitx,xcolor,tikz}

\usepackage{amssymb}
\usepackage{pifont}
\newlength\maxlentime

\newcommand\pesqtimebarstd[3][red!16]{%
  \edef\valuewithstd{#3}
  \StrBefore{\valuewithstd}{±}[\mainvalue]
  \StrBehind{\valuewithstd}{±}[\stdvalue]
  \FPeval\result{round((\mainvalue-0)/#2:4)}
  \rlap{\textcolor{#1}{\hspace*{\dimexpr-\tabcolsep+.5\arrayrulewidth - 6pt}%
        \rule[-.05\ht\strutbox]{\result\maxlentime}{.95\ht\strutbox}}}%
  \makebox[\dimexpr\maxlentime-0.2\tabcolsep+\arrayrulewidth][c]{%
    \ensuremath{\mainvalue\!\pm\!\stdvalue}
  }%
}

\newcommand\pesqtimebarstdunderline[3][red!16]{%
  \edef\valuewithstd{#3}
  \StrBefore{\valuewithstd}{±}[\mainvalue]
  \StrBehind{\valuewithstd}{±}[\stdvalue]
  \FPeval\result{round((\mainvalue-0)/#2:4)}
  \rlap{\textcolor{#1}{\hspace*{\dimexpr-\tabcolsep+.5\arrayrulewidth - 6pt}%
        \rule[-.05\ht\strutbox]{\result\maxlentime}{.95\ht\strutbox}}}%
  \makebox[\dimexpr\maxlentime-0.2\tabcolsep+\arrayrulewidth][c]{%
    \ensuremath{\underline{\mainvalue \!\pm\! \stdvalue}}
  }%
}

\newcommand\stoitimebarstd[3][cyan!20]{%
  \edef\valuewithstd{#3}
  \StrBefore{\valuewithstd}{±}[\mainvalue]
  \StrBehind{\valuewithstd}{±}[\stdvalue]
  \FPeval\result{round((\mainvalue-0)/#2:4)}
  \rlap{\textcolor{#1}{\hspace*{\dimexpr-\tabcolsep+.5\arrayrulewidth - 6pt}%
        \rule[-.05\ht\strutbox]{\result\maxlentime}{.95\ht\strutbox}}}%
  \makebox[\dimexpr\maxlentime-0.2\tabcolsep+\arrayrulewidth][c]{%
    \ensuremath{\mainvalue\!\pm\!\stdvalue}
  }%
}

\newcommand\stoitimebarstdunderline[3][cyan!20]{%
  \edef\valuewithstd{#3}
  \StrBefore{\valuewithstd}{±}[\mainvalue]
  \StrBehind{\valuewithstd}{±}[\stdvalue]
  \FPeval\result{round((\mainvalue-0)/#2:4)}
  \rlap{\textcolor{#1}{\hspace*{\dimexpr-\tabcolsep+.5\arrayrulewidth - 6pt}%
        \rule[-.05\ht\strutbox]{\result\maxlentime}{.95\ht\strutbox}}}%
  \makebox[\dimexpr\maxlentime-0.2\tabcolsep+\arrayrulewidth][c]{%
    \ensuremath{\underline{\mainvalue \!\pm\! \stdvalue}}
  }%
}

\def\headertime{New}
\def\fig{Fig.\xspace}
\def\eqn{Eq.\xspace}
\def\sec{Sec.\xspace}
\def\tab{Tab.\xspace}

\def\ie{{\textit{i.e.}\xspace}} 
\def\eg{{\textit{e.g.}\xspace}}

\newcommand{\head}[1]{{\noindent \textbf{#1:}}}
\newcommand{\term}[1]{{\textit{#1}}}

\graphicspath{{./figures/}}
\graphicspath{{./pdf/}}
\DeclareGraphicsExtensions{.pdf,.jpeg,.png}

\ifodd 0
\newcommand{\rev}[1]{{\color{blue}#1}} 
\newcommand{\com}[1]{\textbf{\color{red}(COMMENT: #1)}} 
\newcommand{\todo}[1]{\textbf{{\color{orange}(TODO: #1)}}}
\newcommand{\unused}[1]{}
\else
\newcommand{\rev}[1]{#1}
\newcommand{\com}[1]{}
\newcommand{\todo}[1]{}
\fi

\ifodd 0
\newcommand{\orev}[1]{{\color{blue}#1}} 
\else
\newcommand{\orev}[1]{#1}
\fi

\setcopyright{acmlicensed}
\copyrightyear{2026}
\acmYear{2026}
\acmDOI{XXXXXXX.XXXXXXX}

\def\sysname{\textsc{RF-CRATE}\xspace}

\acmSubmissionID{4683}

\begin{document}

\title[\sysname]{Towards White-Box Deep Wireless Sensing}

\author{Xie Zhang}
\affiliation{%
  \institution{The University of Hong Kong}
  \city{Hong Kong SAR}
  \country{China}}
\email{zhangxie@connect.hku.hk}
\orcid{0000-0003-4103-6256}

\author{Yina Wang}
\authornote{This work was performed while Y. Wang was a summer intern at HKU.}
\affiliation{%
    \institution{MIT}
    \city{Cambridge}
    \country{USA}
}
\email{yina1123@mit.edu}

\author{Chenshu Wu}
\authornote{Corresponding author.}
\affiliation{%
  \institution{The University of Hong Kong}
  \city{Hong Kong SAR}
  \country{China}}
\email{chenshu@cs.hku.hk}
\orcid{0000-0002-9700-4627}

\renewcommand{\shortauthors}{\sysname}

\begin{abstract}
The empirical success of deep learning has spurred its application to the radio-frequency (RF) domain, leading to significant advances in Deep Wireless Sensing (DWS). 
However, most existing DWS models remain black boxes, with ad-hoc architectures and learned representations lacking explicit physical and mathematical grounding, which limits their reliability and generalizability in real-world deployments.
We present \sysname, an early step towards white-box DWS grounded in the complex sparse rate reduction principle.
Using the $\mathbb{CR}$-Calculus framework, we derive a fully complex-valued transformer with mathematically interpretable self-attention and residual modules.
To address labeled data scarcity, we introduce subspace regularization to enhance representation diversity, yielding a 19.98\% average improvement.
We evaluate \sysname across heterogeneous RF modalities and human sensing tasks, including activity, gait, and gesture recognition, pose estimation, and respiration monitoring.
Experiments on five datasets show that \sysname remains competitive with strong black-box models while providing mathematically interpretable architectures and representations.
Moreover, the complex-valued design achieves a 3.39\% gain in classification accuracy and a 10.34\% reduction in regression error.
Our results demonstrate that mathematically grounded models can achieve strong performance in wireless sensing, offering a promising step towards physically aligned white-box DWS systems.
\end{abstract}

\setcopyright{acmlicensed}
\acmJournal{IMWUT}
\acmYear{2026} \acmVolume{10} \acmNumber{3} \acmArticle{191}
\acmMonth{9} \acmDOI{10.1145/3831998}




\maketitle

\input{body_v3}

\bibliographystyle{ACM-Reference-Format}
\bibliography{refs}



\newpage
\input{appendix}

\end{document}

%% file: body_v3.tex

\section{Introduction}
\label{sec:intro}
Wireless sensing has emerged as a powerful paradigm for human-centric perception, enabling device-free understanding of human activities and physiological states through radio-frequency (RF) signals in everyday environments.
By leveraging the ubiquity of wireless infrastructure and the expressive power of deep learning \cite{jumperHighlyAccurateProtein2021a, biAccurateMediumrangeGlobal2023, fawzi2022discovering}, deep wireless sensing (DWS) \cite{zhangCrossSenseCrossSiteLargeScale2018, zhengMoReFiMotionrobustFinegrained2021, dingRFnetUnifiedMetalearning2020, yangSLNetSpectrogramLearning2023, liangMmStressDistillingHuman2023} has demonstrated remarkable progress across a wide range of applications, including activity recognition \cite{dingRFnetUnifiedMetalearning2020, yangSLNetSpectrogramLearning2023}, gesture interaction \cite{zhengZeroEffortCrossDomainGesture2019}, gait analysis \cite{wuGaitWayMonitoringRecognizing2021}, pose estimation \cite{renGoPose3DHuman2022, zhao2018through, zhao2018rf, Lee_2023_WACV_hupr}, and vital-sign monitoring \cite{zhengMoReFiMotionrobustFinegrained2021, guoContactlessFinegrainedCardiac2025, wangRFGymCareIntroducingRespiratory2024}.

\begin{figure*}[t]
    \centering
    \includegraphics[width=0.85\textwidth]{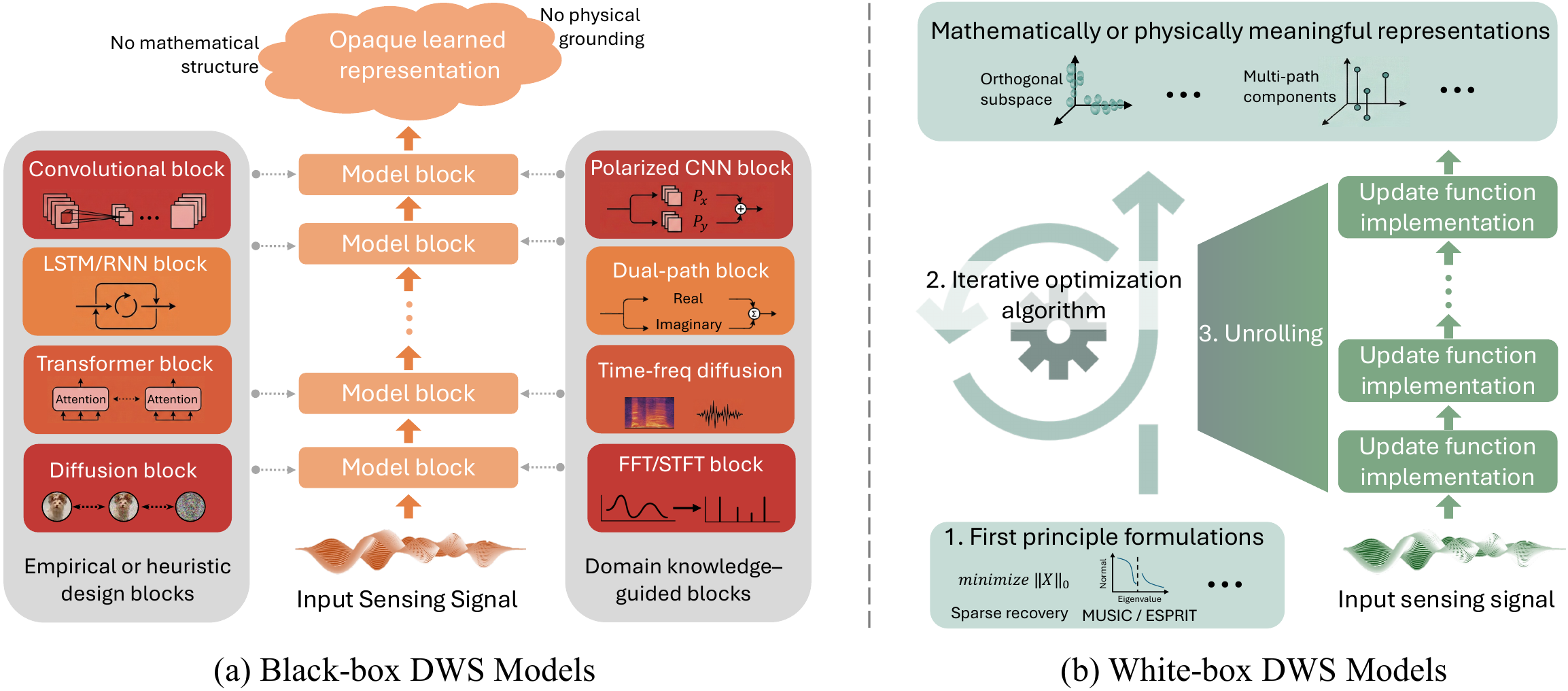}
    \vspace{-0.1in}
    \caption{Black-box versus white-box paradigms in deep wireless sensing. 
    \rm{
    (a) Black-box DWS models combine \textit{empirically designed} deep-learning modules with domain knowledge-guided components. Their internal representations are opaque and lack explicit mathematical or physical structure.
    (b) White-box DWS models arise from mathematically specified first-principles formulations. The model is \textit{derived} by unrolling the iterative optimization algorithm, and its learned representations correspond to explicit mathematical objectives and potentially physically meaningful structures, enabling post hoc analysis and diagnosability.
    }
    }
    \label{fig:black_box_vs_white_box}
\end{figure*}

Despite these advances, most existing DWS systems are built upon deep neural networks whose internal mechanisms remain opaque.
Their architectures are largely designed empirically, and the learned representations often lack explicit physical or mathematical meaning.
This stands in contrast to classical model-based signal processing approaches \cite{wangUnderstandingModelingWiFi2015a, qianWidar2PassiveHuman2018}, where each computational step and representation is grounded in analytical formulations of signal propagation.
As illustrated in \fig\ref{fig:black_box_vs_white_box}(a), contemporary DWS models typically combine generic deep-learning architectures such as CNNs \cite{heDeepResidualLearning2016, huangDenselyConnectedConvolutional2018}, LSTMs \cite{chung2014empirical, zhengZeroEffortCrossDomainGesture2019}, and transformers \cite{dosovitskiyImageWorth16x162021, liuSwinTransformerHierarchical2021b, liuSwinTransformerV22022} with domain-inspired modules, including STFT blocks \cite{yaoSTFNetsLearningSensing2019}, dual-path real–imaginary processing \cite{dingRFnetUnifiedMetalearning2020}, polarized convolutions \cite{yangSLNetSpectrogramLearning2023}, and time-frequency diffusion blocks \cite{chi2024rf}.
Although these designs improve performance, they remain heuristic compositions of \textbf{black-box} layers and cannot be rigorously integrated with model-based signal processing frameworks \cite{wuGaitWayMonitoringRecognizing2021, wuRFbasedInertialMeasurement2019, qianWidar2PassiveHuman2018, wangEeyesDevicefreeLocationoriented2014, zengWiWhoWifibasedPerson2016, puWholehomeGestureRecognition2013}.
Consequently, existing DWS systems often exhibit limited interpretability, fragile generalization, and unpredictable behavior under domain shifts \cite{dingRFnetUnifiedMetalearning2020}, environmental changes \cite{zhengZeroEffortCrossDomainGesture2019, zhangWiFibasedCrossDomainGesture2021}, or adversarial perturbations \cite{xu2022wicam}.
Such opacity poses fundamental challenges for deploying DWS systems in safety- and trust-sensitive domains, such as healthcare \cite{zhengMoReFiMotionrobustFinegrained2021} and autonomous environments \cite{heVIMapInfrastructureAssistedRealTime2023, shiVIPSRealTimePerception2022}.

Recently, a new class of deep learning models—referred to as \textbf{white-box} models—has emerged in computer vision and natural language processing \cite{yuWhiteBoxTransformersSparse2023a, yuEmergenceSegmentationMinimalistic2023, zlahticTransferringBlackBoxDecision2024, yangScalingWhiteBoxTransformers2024a, luoGloballyPredictableKSpace2026}.
Unlike black-box architectures, white-box models are constructed directly from mathematically specified optimization programs, where the model architecture corresponds to an unrolled optimization algorithm and the learned representations are governed by explicit objectives.
As shown in \fig\ref{fig:black_box_vs_white_box}(b), this paradigm enables deep models whose internal operations admit clear mathematical interpretations and may facilitate physically grounded analysis.
Such a paradigm is particularly appealing for wireless sensing.
First, mathematically grounded models can naturally interface with RF propagation theory, optimization frameworks, and classical signal processing methods such as sparse recovery \cite{yuWhiteBoxTransformersSparse2023a}, multi-path modeling \cite{tseFundamentalsWirelessCommunication2005a}, and MUSIC \cite{wangPhaseBeatExploitingCSI2017a, chenTRBREATHTimeReversalBreathing2018}.
Second, their representations can, in favorable cases, be related to physically meaningful structures such as subspaces, propagation paths, or multi-path components, enabling diagnosability and system-level reasoning in real-world sensing deployments.
These properties suggest that white-box modeling may provide a principled pathway towards more interpretable and analyzable DWS systems.

In this paper, we present \sysname, the first complex-valued, mathematically grounded white-box framework tailored for wireless sensing.
\sysname is inspired by CRATE \cite{yuWhiteBoxTransformersSparse2023a}, a white-box transformer whose architecture is derived from iterative sparse rate reduction optimization and achieves competitive performance with empirically engineered transformers such as ViT \cite{dosovitskiyImageWorth16x162021}.
CRATE’s success in vision, together with the promise of white-box modeling, motivates us to explore whether such mathematically grounded architectures can be extended to the fundamentally different domain of RF sensing.
Building on this insight, we develop \sysname as a complex-valued white-box model grounded in a mathematically defined complex sparse rate reduction framework.

As an early step towards white-box DWS, \sysname focuses on establishing mathematical interpretability as a foundational capability, \orev{\textit{while paving the way for future models that may further achieve physical interpretability}}.
Yet, designing a mathematically grounded white-box framework for DWS is fundamentally different from existing vision-domain white-box models and introduces several unique challenges.

\head{\ding{182} \rev{Preserving physical semantics in the complex domain}}
RF measurements such as WiFi CSI and radar waveforms encode both amplitude and phase, which correspond to physically meaningful quantities such as propagation delay, Doppler shifts, and multi-path interference \cite{tseFundamentalsWirelessCommunication2005a}.
To preserve these physical semantics, a white-box DWS model must operate natively in the complex domain, rather than relying on real-valued approximations that decouple or discard phase information.
However, extending white-box models to the complex domain is non-trivial, as the core optimization objective underlying sparse rate reduction becomes non-holomorphic and cannot be handled by standard complex calculus.

\head{\ding{183} \rev{Mathematically grounding residual architectures}}
Residual feed-forward blocks with skip connections are critical for stabilizing optimization and enhancing expressiveness in transformer models \cite{dosovitskiyImageWorth16x162021, liuSwinTransformerV22022}, and are widely adopted in existing DWS systems \cite{dingRFnetUnifiedMetalearning2020, liUniTSShortTimeFourier2021}.
\rev{CRATE \cite{yuWhiteBoxTransformersSparse2023a} effectively realizes a white-box transformer model, yet it exhibits a significant gap where the introduction of residual structures via orthogonal dictionaries lacks a rigorous mathematical derivation.}
Bridging this gap is essential for a fully white-box DWS: architectures should remain fully derivable while preserving the performance gains that residual connections empirically provide.

\head{\ding{184} \rev{Training principled models under labeled data constraints}}
While white-box transformers such as CRATE benefit from pre-training on massive datasets like ImageNet, large-scale labeled RF datasets remain scarce.
For example, the widely used Widar3.0 dataset \cite{zhengZeroEffortCrossDomainGesture2019} contains only 258,575 samples—nearly 55 times fewer than ImageNet-21K.
More recent multi-modal RF datasets are even smaller, such as OctoNet \cite{yuanOctoNetLargeScaleMultiModal2025} with 8.76k samples, XRF55 \cite{wangXRF55RadioFrequency2024} with 42.9k samples, and mmFi \cite{yangMMFiMultiModalNonIntrusive2023} with 1,080 samples.
These constraints pose a fundamental challenge: how to train mathematically principled yet expressive white-box models under limited data regimes.

To address these challenges, we develop \sysname through three key design innovations.

\noindent$\blacksquare$ First, we derive a complex-valued self-attention mechanism tailored for RF sensing.
By extending sparse rate reduction to the complex domain and leveraging the $\mathbb{CR}$-Calculus framework \cite{kreutz_delgadoComplexGradientOperator2009}, we rigorously derive an \emph{RF self-attention module} 
that directly handles complex-valued signals, resolving objective non-holomorphicity to yield a mathematically grounded attention mechanism.

\noindent$\blacksquare$ Second, we derive a principled residual feed-forward structure in the complex domain.
By treating the metric tensor as a learnable parameter during derivation, we obtain a fully interpretable complex-valued residual MLP module, termed \emph{RF-MLP}.
In contrast to the ad hoc residual blocks in the original CRATE model, RF-MLP arises directly from the underlying optimization program, enabling a fully derivable, residual-enabled transformer-like architecture for DWS.

\noindent$\blacksquare$ Third, we present a Subspace Regularization (SSR) strategy to enable effective training under limited RF data.
SSR promotes balanced feature allocation across learned subspaces, enhancing representation diversity and improving generalization without altering the theoretically derived architecture.
Importantly, SSR is enabled by the white-box nature of \sysname, which provides direct access to subspace-level representations for analysis and optimization.

\begin{figure*}[t]
    \centering
    \includegraphics[width=0.85\textwidth]{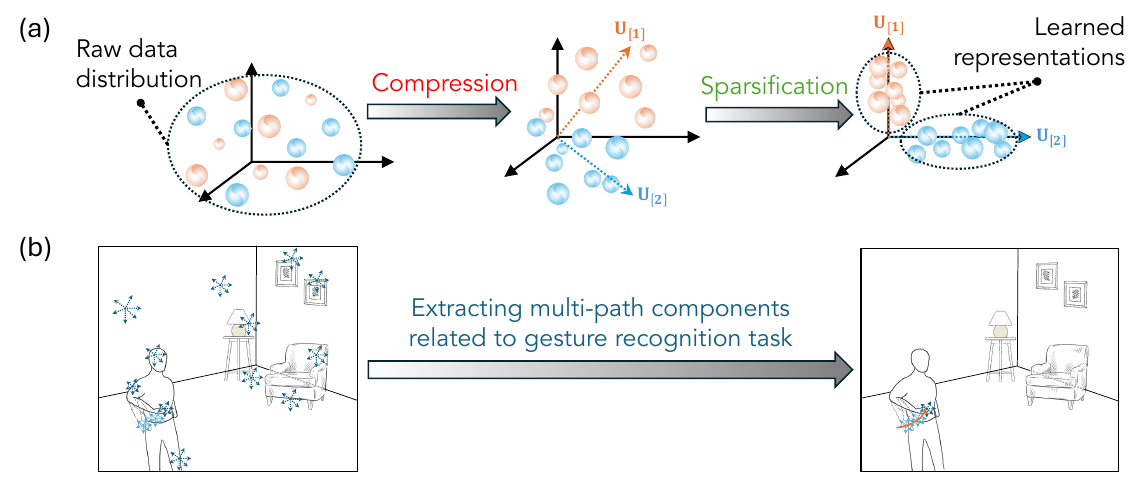}
    \vspace{-0.1in}
    \caption{
    \orev{Parsimony in representation learning and wireless sensing. \rm{(a) Representation space: The sparse rate reduction principle identifies structured representations by compressing features into incoherent, low-dimensional subspaces and enforcing sparsity, which facilitates semantic learning and enhances model generalization. 
    (b) Sensing signal space: Intrinsic parsimony in wireless sensing, exemplified by WiFi gesture recognition. Here, the sensing process resolves specific multi-path components associated with hand motion, projecting high-dimensional RF signals into physically meaningful, low-dimensional 3D temporal trajectories.}}
    }
    \label{fig:principle_parsimony}
\end{figure*}

We evaluate \sysname across heterogeneous RF modalities and diverse human sensing tasks, including activity, gait, and gesture recognition, as well as pose estimation and respiration monitoring.
Our evaluation spans four public datasets and a self-collected WiFi dataset, covering WiFi CSI, UWB radar, and FMCW mmWave radar signals.
Across tasks and modalities, \sysname achieves competitive performance against strong black-box baselines while providing explicit mathematical transparency.
Extending CRATE to the complex RF domain yields an average performance gain of 6.86\% over CRATE, and integrating SSR further improves performance by 19.98\% on average across sensing tasks, although the magnitude of these gains varies with task difficulty and performance headroom.
Beyond performance metrics, we conduct in-depth analyses of learned representations, providing qualitative evidence that the empirical behavior is broadly consistent with the theoretical design.
Overall, \sysname represents a principled early step towards more interpretable DWS.
By bridging deep learning with mathematical learning theory, it enables architectures that are fully derivable and interpretable from a rigorous mathematical perspective, while establishing a foundation for future work on more physically grounded sensing models.
\textit{\sysname is open-sourced at} {\color{blue}\url{https://github.com/aiot-lab/RF-CRATE}}.

In summary, we make the following core contributions:
\begin{itemize}[noitemsep, topsep=0pt, leftmargin=*]
\item \textbf{Mathematically grounded white-box DWS framework:} We propose \sysname, a complex-valued, mathematically grounded white-box framework for deep wireless sensing, whose architecture is fully derived from complex sparse rate reduction rather than empirically engineered.
\item \textbf{Complex transformer derivation:} We extend the white-box transformer CRATE to wireless sensing by deriving complex-valued self-attention and residual MLP modules under the $\mathbb{CR}$-Calculus framework, establishing a mathematically interpretable transformer architecture for complex RF signals.
\item \textbf{Data-efficient training via subspace regularization:} We introduce Subspace Regularization (SSR) to improve representation diversity and enable effective training under limited RF data, without altering the principled model structure.
\item \textbf{Comprehensive evaluation across modalities and tasks:} We implement \sysname as a fully complex-valued system and evaluate it across five datasets and three RF modalities, demonstrating competitive performance against strong black-box baselines while offering explicit mathematical transparency.
\end{itemize}

\section{Preliminary}
\label{sec:primer}

This section establishes the conceptual and physical foundations of our approach.
We first review the core idea of white-box transformers through CRATE \cite{yuWhiteBoxTransformersSparse2023a}.
We then show that wireless sensing inherently exhibits sparse and low-dimensional structure, making it naturally compatible with the white-box paradigm.
Finally, we explain why a complex-valued formulation is essential for wireless sensing and why extending CRATE to the complex domain is both necessary and challenging.

\subsection{White-box Transformers}
\label{white_box_t}
Representation learning aims to learn a continuous mapping $f: \mathbb{R}^{D} \to \mathbb{R}^{d}$ that transforms a data point $\mathbf{x} \in \mathbb{R}^{D}$ into a representation vector $\mathbf{z} \in \mathbb{R}^{d}$, where $D$ and $d$ are the dimensions of the original data space and the representation space, respectively.

A fundamental challenge in representation learning is not only to achieve high performance, but also to understand what structure the learned representations encode.
CRATE addresses this challenge by presenting a white-box transformer architecture that prioritizes parsimony \cite{yuWhiteBoxTransformersSparse2023a}. 
Rather than treating neural networks as black-box function approximators, CRATE interprets representation learning as a structured process of compressing input data into sparse, low-dimensional subspaces.
Specifically, when learned representations can be modeled as mixtures of low-dimensional Gaussian distributions supported on incoherent subspaces, their quality can be characterized by a unified objective termed \textit{sparse rate reduction} \cite{yuWhiteBoxTransformersSparse2023a}. 
As illustrated in \fig\ref{fig:principle_parsimony}(a), sparse rate reduction jointly performs dimensionality reduction and sparsification, yielding representations that are compact, structured, and interpretable.

By viewing each network layer as an iterative optimization step of the sparse rate reduction objective, CRATE derives transformer-like architectures whose internal operations admit explicit mathematical interpretation.
These white-box transformers have demonstrated strong empirical performance while preserving transparency and scalability \cite{yuWhiteBoxTransformersSparse2023a, yang2024scaling}.
However, existing studies of CRATE have focused primarily on computer vision, leaving open the question of whether such mathematically grounded architectures can capture the structure of wireless sensing signals.

\subsection{Intrinsic Parsimony of RF Sensing}

Wireless sensing signals are inherently high-dimensional.
They are shaped by high sampling rates, multiple frequency subcarriers, and multi-antenna configurations.
For example, two seconds of WiFi CSI with 30 subcarriers and three links sampled at 1000 Hz results in 180,000 dimensions \cite{zhengZeroEffortCrossDomainGesture2019}.

Yet, in human sensing scenarios, only a small fraction of these dimensions carry information about human motion or physiological activity.
Most signal variations arise from static environmental reflections, noise, or irrelevant multi-path components, while task-relevant information is concentrated in a few propagation paths affected by human movement \cite{yangSLNetSpectrogramLearning2023, adib3DTrackingBody2014}.
\fig\ref{fig:principle_parsimony}(b) illustrates this phenomenon in WiFi-based gesture recognition.
Although raw CSI captures reflections from all surrounding objects, only a limited number of multi-path components induced by hand motion are essential for recognizing gestures.
These components can be interpreted as trajectories of a small number of hand joints, forming a low-dimensional temporal structure embedded in a high-dimensional RF signal space.
More broadly, wireless human sensing can be viewed as the problem of extracting sparse, structured latent factors—corresponding to human motion, body dynamics, and environmental interactions—from complex RF measurements.
This implies that the underlying representations of RF sensing tasks naturally reside in sparse, low-dimensional subspaces.

This intrinsic structure closely aligns with the sparse rate reduction principle underlying CRATE.
Motivated by this alignment, we extend the white-box paradigm from vision to wireless sensing and develop \sysname, a mathematically grounded, complex-valued white-box framework for DWS.

\subsection{Why a Complex Extension is Needed}

Wireless sensing is fundamentally governed by complex-valued signal models \cite{adibSeeWallsWiFi2013a, zhangFresnelZoneBased2021, zhangWiDetectRobustMotion2019, tseFundamentalsWirelessCommunication2005a}.
RF signals are naturally represented in the complex domain, where amplitude and phase jointly encode physically meaningful quantities such as signal attenuation, propagation delay, Doppler shifts, and multi-path interference.

For wireless human sensing, these complex-valued characteristics are not merely mathematical artifacts but carry essential physical information about human motion and environmental dynamics.
Therefore, models that aim to achieve interpretable and physically grounded representations must operate natively in the complex domain, rather than relying on real-valued approximations that separate or discard phase information.
Accordingly, \sysname is designed as a complex-valued white-box framework whose derivation begins directly in the complex domain, with complex-valued inputs, parameters, and operators.
This design choice is critical for maintaining consistency with classical RF signal models and for preserving the physical meaning embedded in learned representations.
However, extending CRATE to the complex domain is fundamentally challenging\footnote{Although CRATE employs Hermitian notation, its derivation is restricted to real-valued inputs.}.
When expressed in complex variables, the sparse rate reduction objective becomes non-holomorphic: it does not satisfy the Cauchy–Riemann conditions and cannot be differentiated using standard complex calculus.
Since CRATE derives its architecture by unrolling optimization steps that depend on gradients of this objective, its original real-valued formulation cannot be directly applied to RF signals.

To address this challenge, \sysname adopts the $\mathbb{CR}$-Calculus framework \cite{kreutz_delgadoComplexGradientOperator2009}, which provides generalized gradients for non-holomorphic complex functions.
This framework enables a rigorous derivation of complex-valued self-attention and residual MLP modules, forming the theoretical foundation of \sysname.
Such a complex-domain formulation is essential not only for mathematical correctness but also for enabling interpretable and physically grounded wireless sensing.

\section{\sysname Design}
\label{sec:design}

In this section, we present the key design ideas behind \sysname and summarize its main results from a \emph{wireless-sensing} perspective. 
At a high level, \sysname is a complex-valued, white-box DWS architecture derived by unrolling an optimization procedure for \emph{complex sparse rate reduction}.
To preserve readability of the main text, we summarize the underlying intuition and the resulting architectural modules, while relegating the full $\mathbb{CR}$-Calculus-based derivations \cite{kreutz_delgadoComplexGradientOperator2009} to the Appendix.

\head{Notations} 
Throughout this paper, vectors are denoted by bold lower-case letters (\eg, $\mathbf{x}$), while matrices and tensors are represented by bold upper-case letters (\eg, $\mathbf{Z}$). Calligraphic letters denote number fields (\eg, $\mathbb{R}$ and $\mathbb{C}$). Furthermore, $\mathbf{Z}^{\top}$ and $\mathbf{Z}^{H}$ indicate the transpose and conjugate transpose of a matrix $\mathbf{Z}$, respectively. The conjugate of a complex number $z$ is denoted by $\overline{z}$. 
We use the terms "complex" and "complex-valued" interchangeably.

\begin{figure*}[t]
    \centering
    \includegraphics[width=0.9\textwidth]{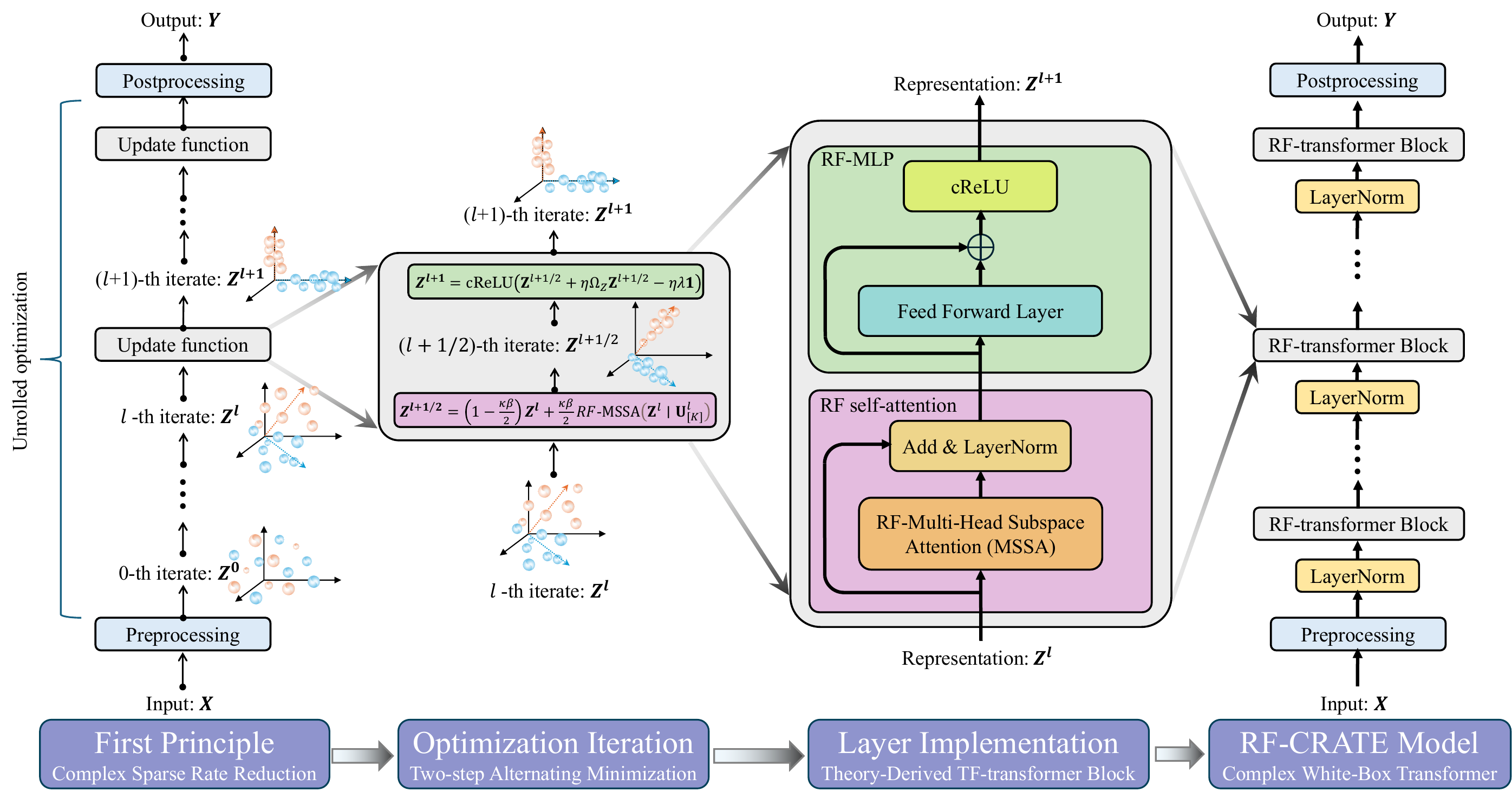}
    \vspace{-0.1in}
    \caption{Overview of \sysname: from a sensing principle to a complex-valued white-box transformer. 
    \sysname unrolls an alternating optimization process for complex sparse rate reduction into repeated RF-transformer blocks, each consisting of (i) an RF self-attention module for \emph{compression into subspaces} and (ii) an RF-MLP module for \emph{sparsification in the complex domain}.}
    \label{fig:model_overview}
\end{figure*}

\subsection{\sysname Overview}
Wireless sensing signals (\eg, CSI) are high-dimensional and complex-valued, but the task-relevant structure is often parsimonious: only a few multi-path components modulated by the human body carry the key information \cite{adibSeeWallsWiFi2013a, adib3DTrackingBody2014, zhangWiDetectRobustMotion2019, zhangFresnelZoneBased2021, tseFundamentalsWirelessCommunication2005a}. 
\sysname is built on this premise and aims to learn representations that are simultaneously \emph{compressed} (low-dimensional) and \emph{sparse} (few active factors), while remaining physically meaningful by operating natively in the complex domain.

As shown in \fig\ref{fig:model_overview}, \sysname is derived from a complex sparse rate reduction objective that quantifies how well extracted features can be (i) encoded compactly and (ii) organized into multiple low-dimensional subspaces with sparse activations. 
We optimize this objective via a two-step alternating procedure. 
Each iteration contains:
(1) a \textbf{\term{compression}} step that projects features into multiple subspaces and selects informative components; and 
(2) a \textbf{\term{sparsification}} step that suppresses irrelevant components to promote parsimonious representations. 
Unrolling these two steps yields an RF-transformer block with two corresponding complex-valued modules:
a \textbf{\term{RF self-attention module}} (compression) and a \textbf{\term{RF-MLP module}} (sparsification). 
Stacking multiple blocks produces a hierarchical model that progressively refines RF representations for downstream sensing tasks.

\subsection{Complex Sparse Rate Reduction}

\head{Setup}
Let $\mathbf{X} = [\mathbf{x}_1, \dots, \mathbf{x}_{N}] \in \mathbb{C}^{D \times N}$ be a sequence of $N$ complex-valued tokens (\eg, temporal segments or frequency-time patches of CSI), where each token $\mathbf{x}_i \in \mathbb{C}^D$. 
We map $\mathbf{X}$ to representations $\mathbf{Z} = [\mathbf{z}_1, \dots, \mathbf{z}_{N}] \in \mathbb{C}^{d \times N}$ with $\mathbf{z}_i \in \mathbb{C}^d$ and $d \ll D$.
To capture the multi-factor nature of RF propagation, we model representations using $K$ orthonormal subspaces $\mathbf{U}_{[K]} = \{\mathbf{U}_1, \dots, \mathbf{U}_K\}$, where $\mathbf{U}_k \in \mathbb{C}^{d \times p}$ and columns of $\mathbf{U}_k$ are orthonormal. 
Each token is assumed to be generated as $\mathbf{z}_i \triangleq \mathbf{U}_{s_i}\boldsymbol{\alpha}_i$, where $s_i \in [K]$ indexes the active subspace and $\boldsymbol{\alpha}_i \in \mathbb{C}^{p}$ is a circularly-symmetric complex Gaussian coefficient.

\head{Objective}
To quantify the parsimony of $\mathbf{Z}$, we extend sparse rate reduction to the complex domain:
\begin{equation}
    L(\mathbf{Z}; \mathbf{U}_{[K]}) 
    = R(\mathbf{Z}) - R^c(\mathbf{Z}\mid \mathbf{U}_{[K]}) - \lambda \|\mathbf{Z}\|_0,
\label{eq:csrr}
\end{equation}
where $\lambda \in \mathbb{R}$ balances \emph{rate reduction} $\Delta R(\mathbf{Z}\mid \mathbf{U}_{[K]}) = R(\mathbf{Z})-R^c(\mathbf{Z}\mid \mathbf{U}_{[K]})$ and sparsity $\|\mathbf{Z}\|_0$. 
Rate reduction measures information gain and has been used to characterize representation quality in white-box deep learning \cite{chanReduNetWhiteboxDeep2022}.

\begin{figure}[t]
    \centering
    \includegraphics[width=0.65\textwidth]{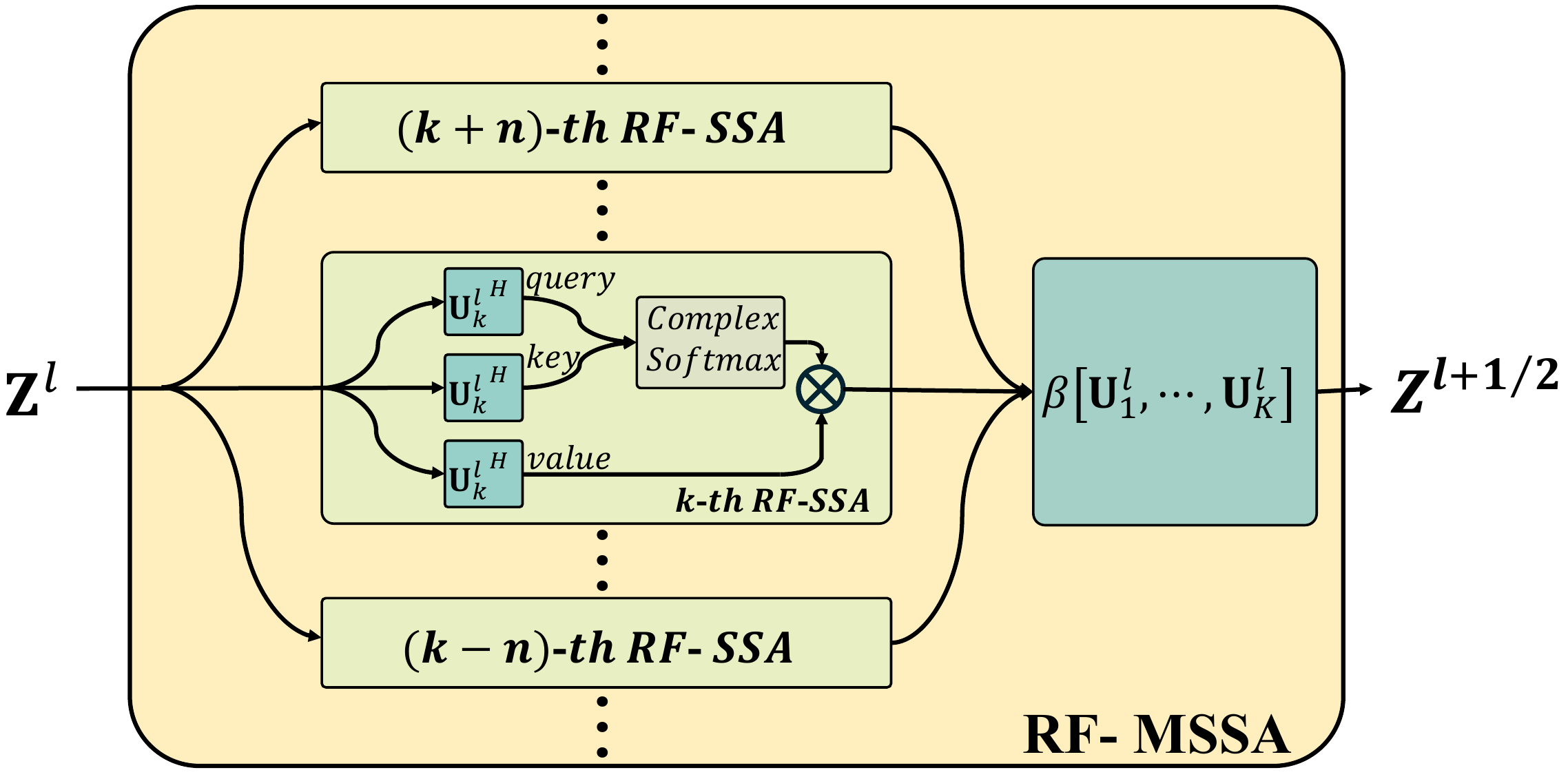}
    \vspace{-0.1in}
    \caption{RF Multi-Head Subspace Attention (RF-MSSA). 
    \rm{Derived from minimizing the subspace coding rate in \eqn\eqref{eq:csrr_optimize_1}, RF-MSSA projects representations into multiple subspaces and performs complex-valued attention within each subspace.}}
    \label{fig:rf_mssa}
\end{figure}

We use the following lossy coding rate estimators:
\begin{equation}
    R(\mathbf{Z}) \triangleq \frac{1}{2} \log \det (\mathbf{I} + \alpha \mathbf{Z}^H \mathbf{Z}),
\label{eq:CRZ}
\end{equation}
\begin{equation}
    R^c(\mathbf{Z} \mid \mathbf{U}_{[K]}) \triangleq \frac{1}{2} \sum_{k=1}^{K} \log \det \Big(\mathbf{I} + \beta (\mathbf{U}_k^H \mathbf{Z})^H (\mathbf{U}_k^H \mathbf{Z})\Big),
\label{eq:CRcZ}
\end{equation}
where $\alpha \overset{\cdot}{=} \frac{d}{N\epsilon^2}$ and $\beta \overset{\cdot}{=} \frac{p}{N\epsilon^2}$ with quantization precision $\epsilon > 0$. 
Since the matrices inside $\log\det(\cdot)$ are Hermitian, both $R(\mathbf{Z})$ and $R^c(\mathbf{Z}\mid\mathbf{U}_{[K]})$ are real-valued, preserving their interpretation as coding rates in the complex field.
At the same time, $L(\cdot)$ is \emph{non-holomorphic} (real-valued with complex arguments), so standard complex derivatives do not apply; \sysname therefore relies on $\mathbb{CR}$-Calculus \cite{kreutz_delgadoComplexGradientOperator2009} for a principled complex-domain derivation.

\head{Learning as an optimization program}
We view representation learning as:
\begin{equation}
    \max_{f \in \mathcal{F}} \ 
    \mathbb{E}_{\mathbf{Z}=f(\mathbf{X})}
    \Big[ R(\mathbf{Z}) - R^c(\mathbf{Z}\mid \mathbf{U}_{[K]}) - \lambda \|\mathbf{Z}\|_0 \Big].
\label{eq:csrr_optimize}
\end{equation}

\head{Unrolling via alternating updates}
Following the forward-construction view in CRATE \cite{yuWhiteBoxTransformersSparse2023a}, we optimize \eqn\eqref{eq:csrr_optimize} with two alternating updates:
\begin{equation}
    \min_{f_1 \in \mathcal{F}_1} \ 
    \mathbb{E}_{\mathbf{Z}^{l+1/2}=f_1(\mathbf{Z}^{l})}
    \Big[ R^c(\mathbf{Z}^{l+1/2} \mid \mathbf{U}_{[K]}^{l}) \Big],
\label{eq:csrr_optimize_1}
\end{equation}
\begin{equation}
    \min_{f_2 \in \mathcal{F}_2} \ 
    \mathbb{E}_{\mathbf{Z}^{l+1}=f_2(\mathbf{Z}^{l+1/2})}
    \Big[ -R(\mathbf{Z}^{l+1}) + \lambda \|\mathbf{Z}^{l+1}\|_0 \Big].
\label{eq:csrr_optimize_2}
\end{equation}
We interpret each RF-transformer block as one iteration of these updates: 
\eqn\eqref{eq:csrr_optimize_1} becomes an RF self-attention module (compression into subspaces), and 
\eqn\eqref{eq:csrr_optimize_2} becomes an RF-MLP module (sparsification). 
We next describe the resulting modules.

\subsection{RF Self-Attention Module Derivation}
\label{subsec:rf-attention-derive}
To solve the optimization program in \eqn\eqref{eq:csrr_optimize_1}, we utilize an approximate gradient descent step to minimize the coding rate $R^{c}(\mathbf{Z}^{l+1/2} \mid \mathbf{U}_{[K]}^{l})$, \ie, $\mathbf{Z}^{l + 1/2} \approx \mathbf{Z}^{l}- \kappa \nabla R^c(\mathbf{Z}^{l} \mid \mathbf{U}_{[K]}^{l})$ with $\kappa > 0$. 
Notably, since we are operating in the complex field and $R^{c}(\cdot)$ is not complex differentiable, we need to generalize the approximate gradient descent to the complex field via the $\mathbb{CR}$-Calculus framework \cite{kreutz_delgadoComplexGradientOperator2009}. 
Specifically, the approximate gradient is derived as follows, omitting the subscript $l$:
\begin{equation} 
\label{eq:rf_attention}
\nabla R^c(\mathbf{Z} \mid \mathbf{U}_{[K]})
 \approx \frac{1}{2} \beta \sum_{k = 1}^{K} \mathbf{U}_k \left( \mathbf{U}_k^H\mathbf{Z} - \beta\mathbf{U}_k^H\mathbf{Z} \left( (\mathbf{U}_k^H\mathbf{Z})^H(\mathbf{U}_k^H\mathbf{Z}) \right) \right),
\end{equation}
where the first equality is based on the conjugate gradient definition and the chain rule in the $\mathbb{CR}$-Calculus framework \cite{kreutz_delgadoComplexGradientOperator2009} and the last approximation uses the von Neumann approximation \cite{krishnan2017neumann}. 

After obtaining the gradient, we perform a gradient descent step with a learning rate $\kappa > 0$ to minimize $R^{c}(\mathbf{Z}^{l+1/2} \mid \mathbf{U}_{[K]}^{l})$, resulting in the following update function:
\begin{equation} 
\label{eq:rf_self_attention}
\begin{split}
Z^{l + 1/2} = (1- \kappa \beta /2)Z^{l} + \kappa \beta/2 \textit{ RF-MSSA}(\mathbf{Z}^{l} \mid \mathbf{U}_{[K]}^{l}).
\end{split}
\end{equation}
$\textit{RF-MSSA}$ (Multi-Head Subspace Attention) is defined as:
\begin{equation}
\label{eq:rf_mssa}
    \mathrm{RF-MSSA}(\mathbf{Z}^{l} \mid \mathbf{U}_{[K]}^{l}) \triangleq \beta
    \begin{bmatrix}
    \mathbf{U}_1^{l \top} \\
    \vdots \\
    \mathbf{U}_K^{l \top}
    \end{bmatrix}^{\top}
    \begin{bmatrix}
    \textit{RF-SSA}(\mathbf{Z}^{l} \mid \mathbf{U}_1^{l}) \\
    \vdots \\
    \textit{RF-SSA}(\mathbf{Z}^{l} \mid \mathbf{U}_K^{l})
    \end{bmatrix},
\end{equation}
where $\textit{RF-SSA}$ (Subspace Attention) operator is defined as:
\begin{equation}
\label{eq:rf_ssa}
\mathrm{RF-SSA}(\mathbf{Z}^{l} \mid \mathbf{U}_{k}^{l}) \triangleq \underbrace{{\mathbf{U}^{l}_k}^H\mathbf{Z}^{l}}_{\textbf{Value}} \mathcal{S}( \underbrace{{(\mathbf{U}^{l}_k}^H\mathbf{Z}^{l})^H}_{\textbf{Query}} \underbrace{{(\mathbf{U}^{l}_k}^H\mathbf{Z}^{l})}_{\textbf{Key}}).
\end{equation}
The $\mathcal{S}(\cdot)$ is the complex softmax function defined as follows to preserve the internal geometry of the vector space \cite{scardapaneComplexValuedNeuralNetworks2020}:
$\mathcal{S}(z_i) \triangleq (e^{\Re(z_i)^2 + \Im(z_i)^2})/(\sum_{j = 1}^{K} e^{\Re(z_j)^2 + \Im(z_j)^2})$
where $z_i$ denotes the element in the token matrix $\mathbf{Z}$.
As shown in \fig\ref{fig:rf_mssa}, the implementation of \eqn\eqref{eq:rf_mssa} corresponds to the RF-MSSA module, which serves as the self-attention mechanism specifically derived for RF data in the \sysname model.

\begin{figure}[t]
    \centering
    \includegraphics[width=0.65\textwidth]{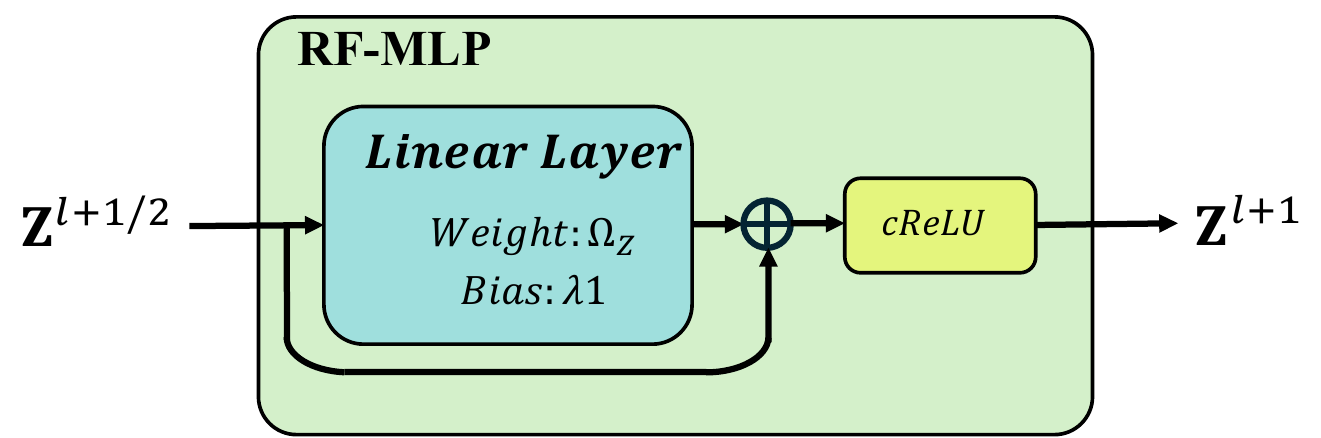}
    \vspace{-0.1in}
    \caption{RF-MLP module. 
    \rm{Derived from the sparsification update in \eqn\eqref{eq:csrr_optimize_2}, RF-MLP applies a complex linear transform with an analytically motivated skip connection and cReLU sparsification.}}
    \label{fig:rf_mlp}
\end{figure}

\subsection{RF-MLP Module Derivation}
\label{subsec:rf-mlp-derive}
To address the optimization program \eqn\eqref{eq:csrr_optimize_2}, we first relax it to a non-negative LASSO program \cite{yuWhiteBoxTransformersSparse2023a}.
Then, within the \(\mathbb{CR}\)-Calculus framework, we derive the quadratic upper bound of the smooth part, i.e., \( -R(\mathbf{Z}^{l+1}) \). 
Subsequently, we minimize the quadratic upper bound and take a proximal step on the non-smooth part, i.e., \(\lambda \|\mathbf{Z}^{l+1}\|_0\).
Finally, we obtain the RF-MLP module with a skip connection by treating the metric tensor as a learnable parameter during the derivation.

Following the approach in CRATE \cite{yuWhiteBoxTransformersSparse2023a}, we relax the $l^0$ "norm" to the $l^1$ norm and impose a nonnegativity constraint in the original optimization program \eqn\eqref{eq:csrr_optimize_2}. This leads to the following program:
\begin{equation}
\label{eq:lasso}
\begin{aligned}
    \min_{f_2 \in \mathcal{F}_2} \mathbb{E}_{\mathbf{Z}^{l+1}} &\left[ -R(\mathbf{Z}^{l+1}) + \lambda \|\mathbf{Z}^{l+1}\|_1 + \chi_{\{\mathbf{Z}^{l+1} \geq 0\}}(\mathbf{Z}^{l+1}) \right],
\end{aligned}
\end{equation}
where $\chi_{\{\mathbf{Z} \geq 0\}}$ denotes the characteristic function for the set of elementwise-nonnegative matrices $\mathbf{Z}^{l+1}$ and $\lambda > 0$.

To minimize \(-R(\mathbf{Z})\), we first derive its quadratic upper bound $Q(\mathbf{Z} \mid \mathbf{Z}^{l+1/2})$ in the neighborhood of the current iteration \(\mathbf{Z}^{l+1/2}\), considering its non-holomorphic property under the \(\mathbb{CR}\)-Calculus framework:
\begin{equation}
\begin{aligned}
&Q(\mathbf{Z} \mid \mathbf{Z}^{l+1/2}) \\&= -R(\mathbf{Z}^{l+1/2}) + \operatorname{Tr} \left( 2\Re \left\{-\nabla_{\mathbf{z}} R(\mathbf{Z}^{l+1/2})^{H} ( \mathbf{Z} - \mathbf{Z}^{l+1/2} )  \right\} \right) \\
& \quad +\frac{9 \alpha}{16} \operatorname{Tr} \left( \Re \left\{\left(\mathbf{Z} - \mathbf{Z}^{l + 1/2} \right)^{H} \left(\mathbf{Z} - \mathbf{Z}^{l + 1/2}\right)\right\} \right),
\end{aligned}
\end{equation}
where $\nabla_{\mathbf{Z}} R(\mathbf{Z}^{l+1/2})$ denotes the gradient. Then, the optimal $\mathbf{Z}_{opt}$ regards the smooth term \(-R(\mathbf{Z})\) is obtained by setting the partial derivative of $Q(\mathbf{Z} \mid \mathbf{Z}^{l+1/2})$ equal to 0 under the assumption that the columns of  $Z^{l+1/2}$ are normalized:
\begin{equation}
    \mathbf{Z}_{opt} = \left( 1+ \frac{16}{9(1+ \alpha)} \Omega_{Z} \right) \mathbf{Z}^{l + 1/2},
\end{equation}
where $\Omega_Z$ is a Hermitian, positive-definite $d \times d$ metric tensor.

To address the remaining term, $\lambda \|\mathbf{Z}^{l+1}\|_1 + \chi_{\{\mathbf{Z}^{l+1} \geq 0\}}(\mathbf{Z}^{l+1})$, we treat the proximal operator of the sum of \(\chi_{\{Z \geq 0\}}\) and \(\lambda \|\cdot\|_1\) as a one-sided soft-thresholding operator:
\begin{equation}
    \text{prox}_{\chi_{\{z \geq 0\}} + \lambda \|\cdot\|_1}(Z) = \max \{ Z - \lambda \mathbf{1}, \mathbf{0} \},
\end{equation}
where the maximum is applied elementwise. To solve the non-smooth part \(\lambda \|\mathbf{Z}^{l+1}\|_1 + \chi_{\{\mathbf{Z}^{l+1} \geq 0\}}(\mathbf{Z}^{l+1})\), we take a proximal majorization-minimization step:
\begin{equation}
\label{eq:ista_theory}
    Z^{l + 1} = \text{cReLU} \left( \left( 1 + \frac{16}{9(1 + \alpha)} \Omega_{Z} \right) \mathbf{Z}^{l + 1/2} - \frac{16 \lambda}{9 \alpha} \mathbf{1} \right),
\end{equation}
where the cReLU function is defined as $\rho(\Re(z)) + i \cdot \rho(\Im(z))$, with $z \in \mathbb{C}$ and $\rho$ representing the real-valued ReLU function.

\begin{figure*}[t]
  \begin{minipage}{0.48\textwidth}
    \centering
    \includegraphics[width=1\textwidth]{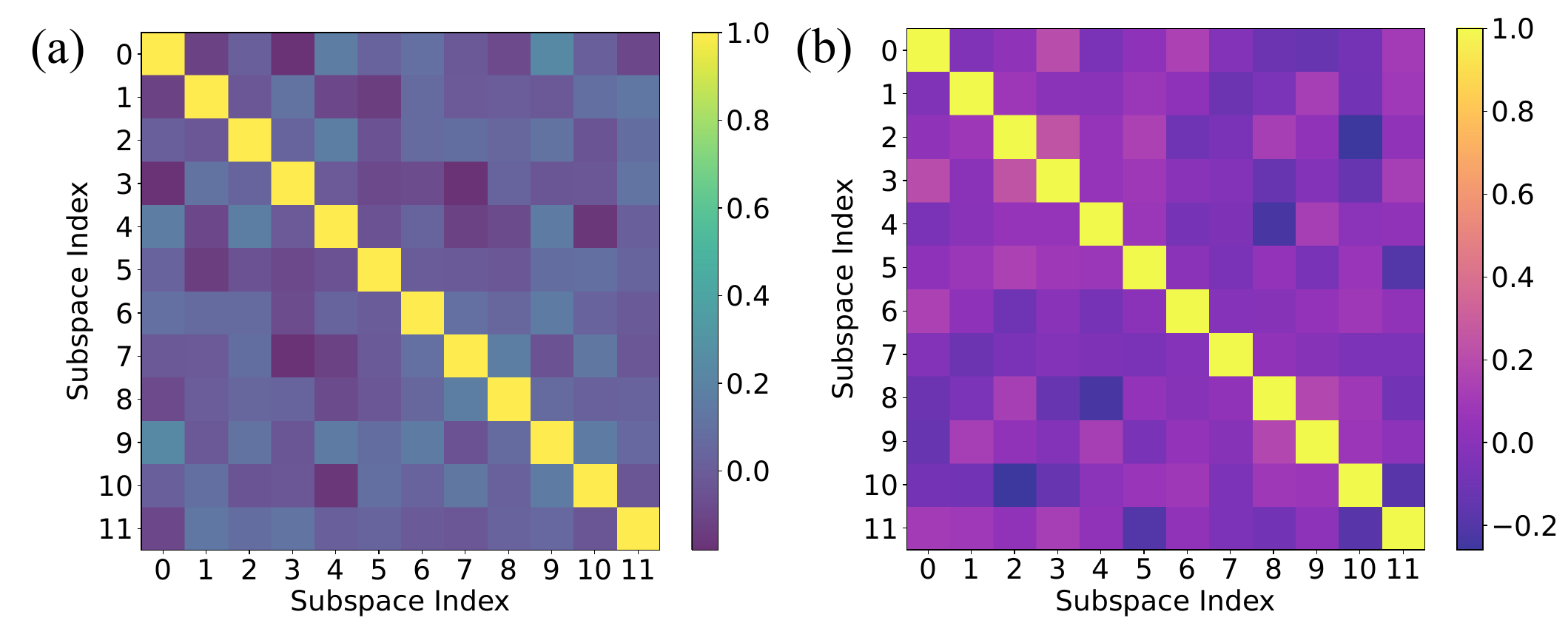}
    \vspace{-0.3in}
    \caption{Subspace correlation heatmaps. 
    \rm{A well-trained \sysname exhibits low inter-subspace correlation, indicating that different subspaces capture complementary factors.}}
    \label{fig:subspace_uncorrelated}
  \end{minipage}
  \hfill
  \begin{minipage}{0.48\textwidth}
    \centering
    \includegraphics[width=1\textwidth]{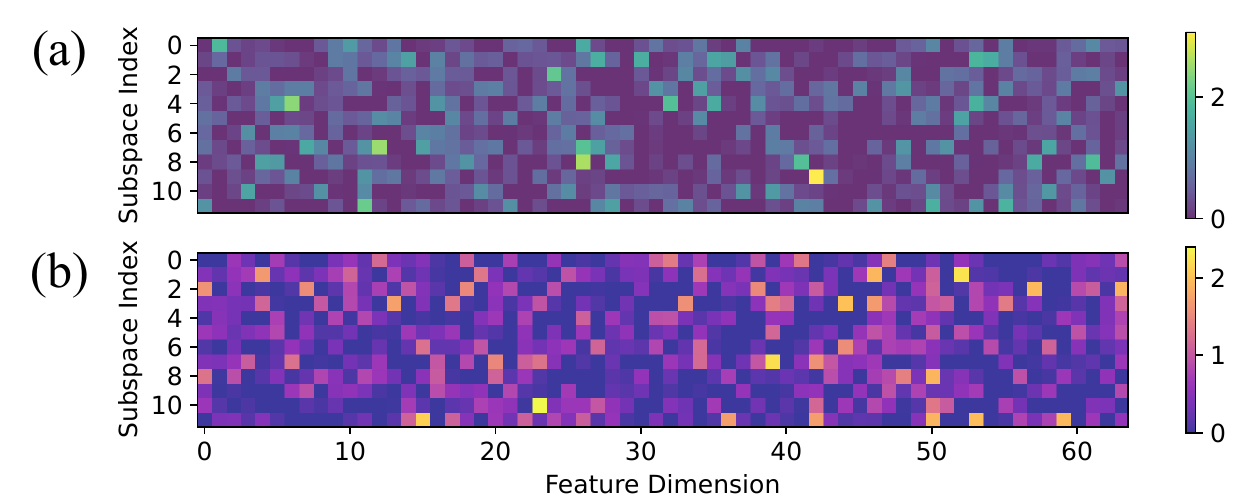}
    \vspace{-0.3in}
    \caption{Feature sparsity in \sysname. 
    \rm{Only a few dimensions in each subspace carry significant magnitude, consistent with the sparsification objective.}}
    \label{fig:sparse_features}
  \end{minipage}
  \hfill
\end{figure*}

Finally, by treating the metric tensor $\Omega_Z$ as a learnable weight matrix and relaxing the constraints that it must be Hermitian and positive definite, we define the $\textit{RF-MLP}$ operator as an approximation to \eqn\eqref{eq:ista_theory}:
\begin{equation}
\label{eq:rf_ista1}
\mathrm{RF-MLP}(\mathbf{Z}^{l + 1/2}) \triangleq \text{cReLU} \left( \mathbf{Z}^{l + 1/2} + \eta \Omega_{Z} \mathbf{Z}^{l + 1/2} - \eta \lambda \mathbf{1} \right),
\end{equation}
where \(\eta = \frac{16}{9(1 + \alpha)}\) is the learning rate. 
Importantly, the skip connection arises naturally from the derivation, making $\textit{RF-MLP}$ a mathematically complete and interpretable design.  
As shown in \fig\ref{fig:rf_mlp}, the implementation of RF-MLP consists of a complex linear transformation and a skip connection.

\head{Discussion on a direct complex adaptation of CRATE}
A direct adaptation of the CRATE derivation introduces an orthogonal dictionary $\mathbf{D} \in \mathbb{C}^{d\times d}$ using the invariance $R(\mathbf{D}\mathbf{Z}) = R(\mathbf{Z})$, leading to:
\begin{equation}
\label{eq:rf_ista2}
\mathrm{RF\text{-}ISTA}(\mathbf{Z}^{l + 1/2})
=
\mathrm{cReLU}\!\left(
\mathbf{Z}^{l + 1/2} + \eta \mathbf{D}^H(\mathbf{Z}^{l + 1/2} - \mathbf{D}\mathbf{Z}^{l + 1/2})
- \eta \lambda \mathbf{1}
\right).
\end{equation}
While \eqn\eqref{eq:rf_ista2} resembles a proximal gradient step, the resulting skip connection is not fully justified in the complex-domain derivation. 
In contrast, RF-MLP in \eqn\eqref{eq:rf_ista1} retains competitive performance while preserving a complete white-box interpretation (\S\ref{sec:performance}).

\subsection{\sysname Model Construction}
As shown in \fig\ref{fig:model_overview}, \sysname model features a hierarchical structure with an RF-transformer block at the core. 

\head{RF-transformer block}
An RF-transformer block corresponds to one iteration of \eqn\eqref{eq:csrr_optimize}. 
It first applies the RF self-attention module (compression; \eqn\eqref{eq:rf_self_attention}) and then a complex LayerNorm. 
The normalized output is processed by RF-MLP (sparsification; \eqn\eqref{eq:rf_ista1}) to produce the block output. 

\head{Hierarchical model}
\sysname stacks multiple RF-transformer blocks to form a hierarchical representation learner: each block further compresses and sparsifies RF representations, progressively improving their utility for the downstream sensing task. 
A lightweight task head (classification or regression) is attached on top depending on the application. 
Importantly, unlike existing DWS models that rely heavily on black-box architectural choices, \sysname can be derived end-to-end from the complex sparse rate reduction principle, yielding the mathematically grounded complex-valued white-box transformer for wireless sensing.

\orev{
\head{Discussion on theoretical relaxations}
To bridge theory and practice, \sysname employs three key relaxations. 
We use a convergent von Neumann approximation for the matrix inversion in \eqn\ref{eq:rf_attention}, relax the NP-hard $\ell_0$-norm to a convex $\ell_1$-norm in \eqn\ref{eq:lasso} for tractable sparse recovery, and treat the metric tensor $\Omega_Z$ in \eqn\ref{eq:rf_ista1} as a learnable weight matrix to adapt to non-stationary RF environments. 
These approximations are essential for practical implementation, ensuring both computational tractability and optimization stability. Future work can further explore constrained neural network architectures to reintegrate these mathematical priors more strictly.
}

\begin{figure}[t]
\centering
    \includegraphics[width=0.7\textwidth]{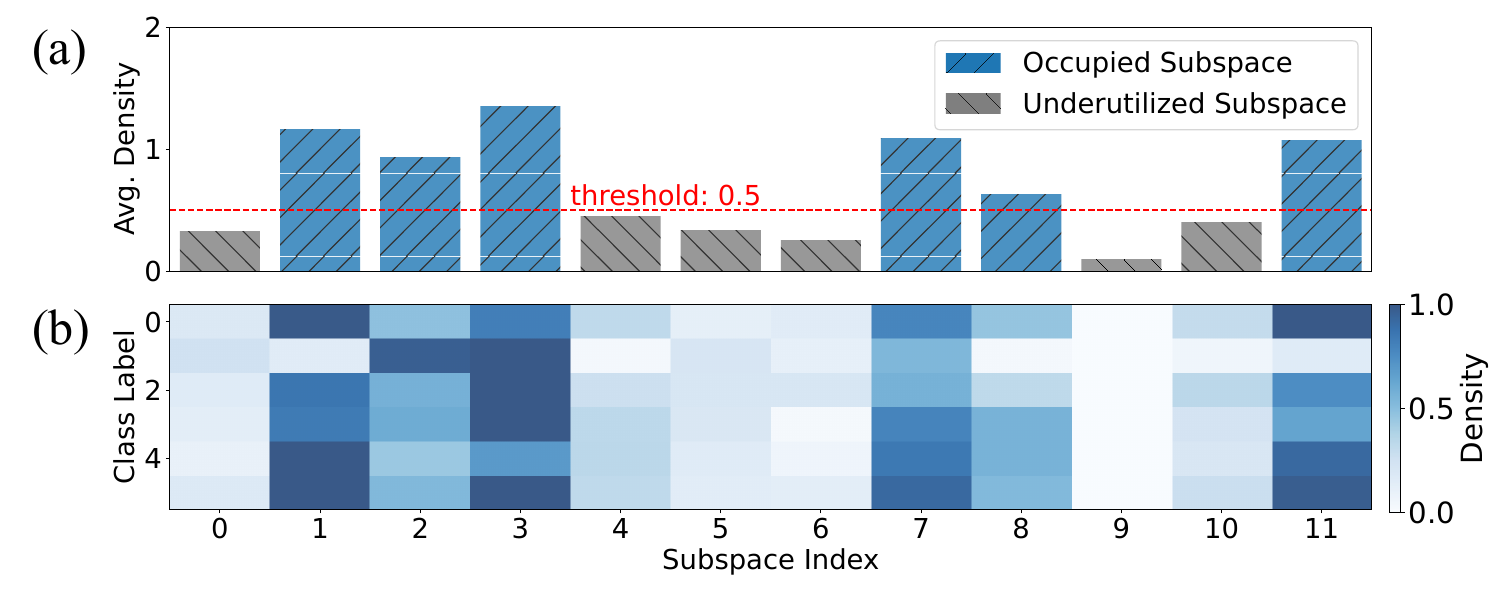}
    \vspace{-0.1in}
    \caption{Subspace occupancy w.r.t.\ gesture categories.
    \rm{Different classes activate subspaces with different densities, revealing how \sysname allocates representation capacity across subspaces.}}
    \label{fig:subspace_class}    
\end{figure}

\subsection{\sysname Model Analysis and SSR}
\label{ssec:model_analysis_ssr}
\sysname is explicitly designed to learn \emph{uncorrelated subspaces} with \emph{sparse activations}. 
To validate that these properties emerge in practice, we inspect a trained \sysname model on WiFi CSI-based gesture recognition with six gesture classes and analyze (i) subspace correlation, (ii) feature sparsity, and (iii) subspace occupancy. 
These observations motivate our Subspace Regularization (SSR).

\head{Subspace uncorrelation} 
The compression step in \eqn\eqref{eq:csrr_optimize_1} maps features into multiple subspaces $\mathbf{U}_{[K]} = (\mathbf{U}_k)_{k\in[K]} \in (\mathbb{C}^{d\times p})^K$. 
When training succeeds, different subspaces should become complementary rather than redundant.
To verify this, we extract the RF-SSA outputs in the final RF-transformer block and compute inter-subspace feature correlations. 
As shown in \fig\ref{fig:subspace_uncorrelated}, correlations concentrate near the diagonal, suggesting that subspaces are largely uncorrelated.

\head{Feature sparsity}
The sparsification step in \eqn\eqref{eq:csrr_optimize_2} encourages sparse activations within each subspace. 
We compute the average magnitude of each feature dimension and observe that only a few dimensions retain significant values while others remain near zero (\fig\ref{fig:sparse_features}), consistent with the intended parsimonious design.

\head{Subspace occupancy and SSR} 
We further examine how different gesture classes occupy the learned subspaces.
Let the \emph{subspace density} be the average magnitude of complex features within a subspace, and the \emph{subspace occupancy} be the average density aggregated over a batch.
\fig\ref{fig:subspace_class} shows that only a subset of subspaces consistently exhibits high occupancy, implying that representation capacity may become imbalanced across subspaces.
Motivated by this observation, we propose Subspace Regularization (SSR) to encourage a more balanced allocation of feature energy across subspaces, promoting diversity in representation:
\begin{equation}
\label{eq:ssr}
\mathrm{SSR} = \sum_{k=1}^{K} (\rho_k - \rho_{\min}),
\end{equation}
where $\rho_k = \frac{1}{N}\sum_{i=1}^{N} |\mathbf{z}_{k,i}|$ denotes the density of the $k$-th subspace over a batch of $N$ features, and $\rho_{\min} = \min_{k\in[K]} \rho_k$ is the minimum density among all subspaces.
We evaluate SSR in \S\ref{sec:performance} and show that it improves robustness by preventing over-concentration on a small subset of subspaces.

\begin{figure}[t]
    \centering
    \includegraphics[width=0.7\textwidth]{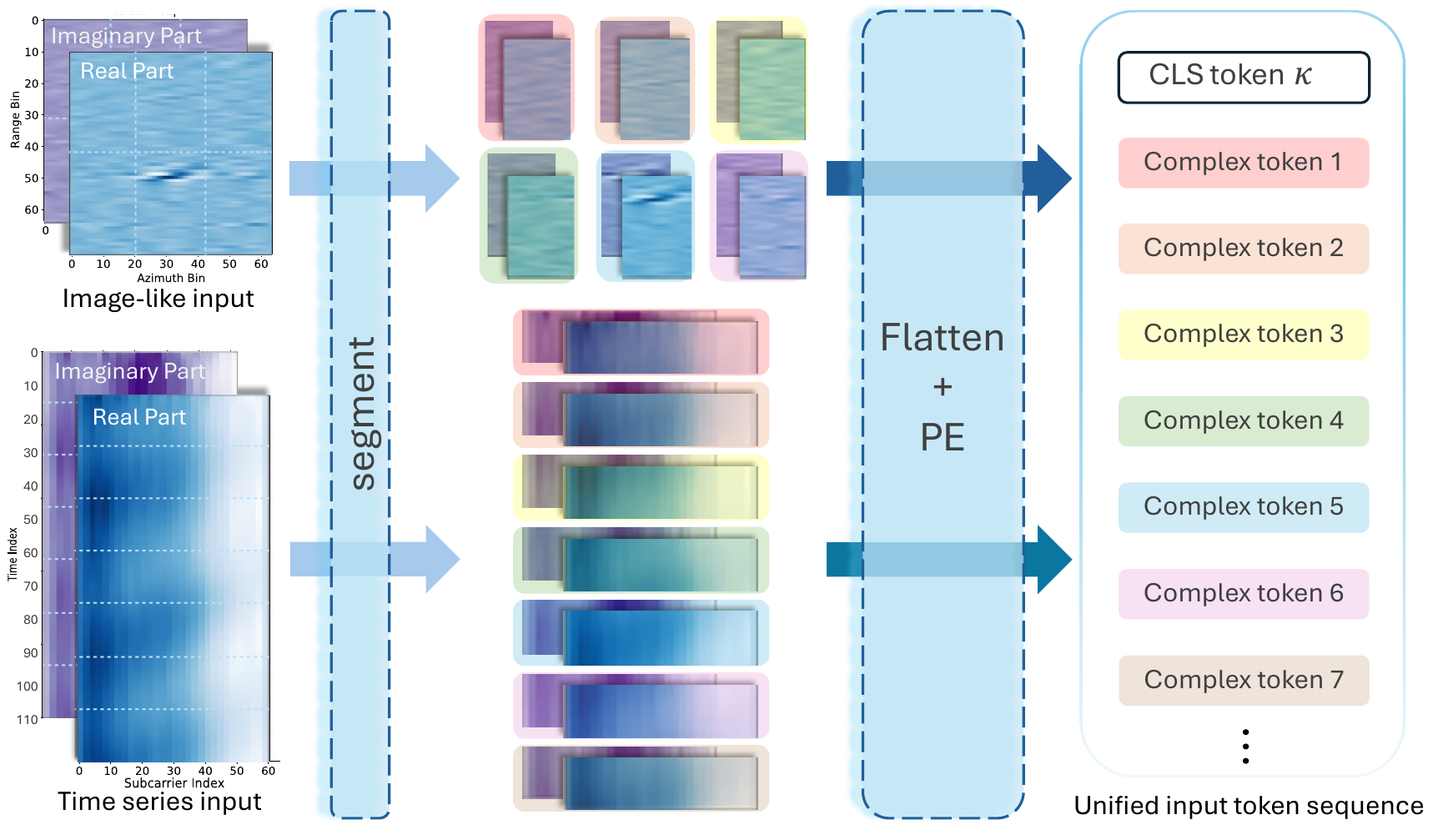}
    \vspace{-0.1in}
    \caption{Patching preprocessing method.}
    \label{fig:patch_preprocess}
\end{figure}

\section{Implementation}
\label{sec:impl}

We implement \sysname as a fully complex-valued network in PyTorch and release the code at {\color{blue}\url{https://github.com/aiot-lab/RF-CRATE}}.
All layers follow complex-domain algebra, and we optimize the model using conjugate Wirtinger derivatives to correctly handle non-holomorphic objectives \cite{kreutz_delgadoComplexGradientOperator2009, trabelsiDeepComplexNetworks2018}.
This section summarizes (i) modality-specific signal preprocessing, (ii) a unified patching pipeline that converts heterogeneous sensing signals into complex token sequences, (iii) task heads and losses, and (iv) the training setup.

\head{Raw signal preprocessing}
\textit{WiFi CSI.}
We follow Widar3.0 preprocessing \cite{zhengZeroEffortCrossDomainGesture2019} and apply conjugate multiplication between CSI streams from two antennas on the same NIC to cancel random phase offsets \cite{li2017indotrack}.
\textit{mmWave radar.}
We perform FFTs along all raw IF-signal dimensions to obtain a range--Doppler--azimuth--elevation representation, and then average over Doppler to yield a 3D spectrum volume \cite{Lee_2023_WACV_hupr}.
\textit{IR-UWB.}
We remove the first twenty CIR samples to mitigate the direct-path component and reduce saturation effects \cite{wuMSenseMobileMaterial2020}.

\head{Unified patching to complex tokens}
Wireless sensing modalities differ in shape (time-series vs.\ image-like tensors), but \sysname requires a shared token interface.
As shown in \fig\ref{fig:patch_preprocess}, we use a unified patching strategy.
For time-series inputs, we segment the complex sequence into overlapping patches along time (sliding windows), which reduces effective sequence length while preserving local dynamics.
For image-like inputs, we patchify across spatial dimensions following the vision transformer paradigm \cite{dosovitskiyImageWorth16x162021}.
Each patch is flattened and mapped to a complex token of dimension $d$, augmented with a positional embedding.
We prepend a complex CLS token $\kappa \sim \mathcal{N}(0,1) \in \mathbb{C}^{1 \times 1 \times d}$ to represent the whole sequence and feed the resulting token stream into \sysname.
This design keeps the model interface consistent across WiFi, radar, and UWB, while allowing the preprocessing to remain lightweight and modality-aware.

\head{Task head and post-processing}
\sysname outputs complex-valued tokens; downstream tasks require real-valued predictions.
We use the final CLS token and compute its modulus to obtain a real-valued feature vector, which is then fed into a linear head.
We use cross-entropy loss for classification tasks (\eg, gesture, activity, gait recognition) and mean squared error for regression tasks (\eg, pose estimation).
The output dimension depends on the task (\eg, six logits for six-class gesture recognition).

\head{Training scheme}  
\sysname is trained on an NVIDIA GeForce RTX 4090 GPU.
We use the AdamW optimizer \cite{loshchilov2019decoupledweightdecayregularization} with an initial learning rate of \( 5 \times 10^{-5} \) and a cosine learning rate scheduler. The training procedure is set for 100 epochs with an early-stopping strategy with patience of 10 epochs.

\section{Experimental Setup}
\label{sec:exp}

In this section, we describe the datasets, baseline methods, and evaluation metrics used to assess \sysname across heterogeneous RF modalities and human sensing tasks.

\begin{figure}[t]
    \centering
    \includegraphics[width=0.7\textwidth]{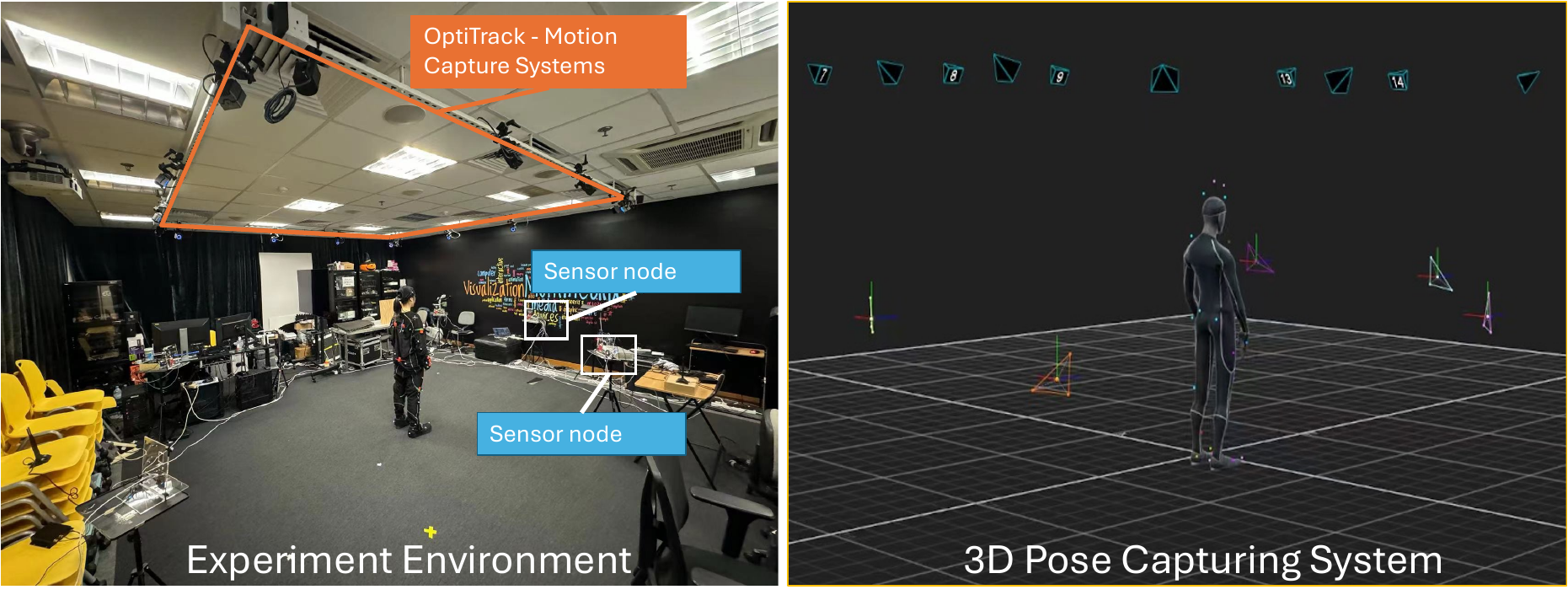}
    \vspace{-0.1in}
    \caption{Dataset collection setup for our self-collected WiFi dataset with motion-capture ground truth.}
    \label{fig:dataset_collection}
\end{figure}

\subsection{Datasets}
We evaluate \sysname on four public datasets and one self-collected dataset, spanning three representative RF modalities: WiFi CSI, FMCW mmWave radar, and IR-UWB.
Together, these datasets cover both \emph{classification} tasks (gesture, activity, and gait identification) and \emph{regression} tasks (2D/3D human pose estimation and respiratory rate estimation).
\fig\ref{fig:dataset_collection} illustrates the experimental setup for our in-house WiFi dataset.

\noindent$\blacksquare$ \textit{GaitID (WiFi CSI, gait identification)} \cite{zhangGaitIDRobustWiFi2020}.
The GaitID dataset contains CSI recordings from 10 subjects walking at varied speeds across two indoor environments.
It is designed to evaluate whether wireless sensing models can capture identity-related gait patterns under environmental variation and mobility diversity.

\noindent$\blacksquare$ \textit{Widar3.0 (WiFi CSI, gesture recognition)} \cite{zhengZeroEffortCrossDomainGesture2019}.
Widar3.0 is a widely used benchmark dataset consisting of CSI data from 16 users across multiple rooms and body orientations.
We use two subsets: Widar3G6 (6 common human–computer interaction gestures) and Widar3G16 (16 gestures including digit-writing motions).
Due to its multi-environment and multi-orientation design, Widar3.0 is particularly suitable for evaluating cross-domain generalization in wireless sensing.

\noindent$\blacksquare$ \textit{OPERAnet (IR-UWB, activity recognition)} \cite{bocusOPERAnetMultimodalActivity2022}.
The OPERAnet dataset provides approximately 8 hours of recordings from 6 participants.
We use its IR-UWB modality to benchmark \sysname on coarse-grained daily activity recognition, where signals are influenced by environmental multi-path, subject diversity, and device-dependent characteristics.

\noindent$\blacksquare$ \textit{HuPR (FMCW mmWave, 2D human pose estimation)} \cite{Lee_2023_WACV_hupr}.
HuPR includes approximately 4 hours of synchronized radar and camera data from 6 subjects performing three types of movements.
We follow the official evaluation protocol to assess radar-based 2D pose estimation, which requires fine-grained spatial inference from RF signals.

\noindent$\blacksquare$ \textit{WiP-pose (WiFi CSI, 3D human pose estimation).}
We built a WiFi sensing testbed integrated with an OptiTrack motion capture system to obtain high-precision 3D pose ground truth (\fig\ref{fig:dataset_collection}).
The dataset contains 10 participants (2 females, 8 males) performing 62 activities covering locomotion, arm movements, torso bending, and multi-limb actions.
Data were collected over 12 days, totaling approximately 8 hours of synchronized CSI and 3D skeletal annotations.
All experimental protocols were reviewed and approved by the local IRB prior to data collection.
WiP-pose evaluates whether \sysname can learn physically meaningful representations for dense, continuous-valued 3D pose regression.

\noindent$\blacksquare$ \textit{WiP-breath (WiFi CSI, respiratory rate estimation).}
We additionally collected a 2-hour dataset for respiratory rate estimation.
Ground truth was recorded using a Polar H10 sensor as an external reference.
WiP-breath evaluates \sysname on subtle periodic micro-motions and low-SNR physiological sensing signals.

\noindent\textbf{Ethics and IRB approval.}
All data collection protocols involving human participants were reviewed and approved by our university’s Institutional Review Board (IRB), and informed consent was obtained from all participants prior to data collection.

\begin{table*}[t]
\caption{Overall Performance. \rm{\orev{Relative performance is visualized via color bars: red denotes higher-is-better metrics (Accuracy), while blue denotes lower-is-better metrics (MPJPE/BPME).}
\underline{Underline} indicates the best \textit{DWS}-specialized baseline; transformer baselines are reported separately to contextualize performance against more generic backbone models.
MPJPE units: centimeters for WiFi 3D Pose, pixels for mmWave Pose. N/A means the dataset input is not supported by the model.}}

  \centering
  \small
    \setlength{\tabcolsep}{3.2pt}  
    \renewcommand{\arraystretch}{1.05}
  \begin{tabular}{cccccccc}
  \toprule
  Model Type & Models    & \begin{tabular}[c]{@{}c@{}}WiFi Gesture \\  (Acc. $\uparrow$)\end{tabular} & \begin{tabular}[c]{@{}c@{}}WiFi Gait \\ (Acc. $\uparrow$)\end{tabular} & \begin{tabular}[c]{@{}c@{}}WiFi 3D Pose\\ (MPJPE $\downarrow$)\end{tabular} & \begin{tabular}[c]{@{}c@{}}WiFi Breath\\ (BPME $\downarrow$)\end{tabular} & \begin{tabular}[c]{@{}c@{}}mmWave Pose\\ (MPJPE $\downarrow$)\end{tabular} & \begin{tabular}[c]{@{}c@{}}UWB Activity\\ (Acc. $\uparrow$)\end{tabular} \\ \hline
  \multirow{3}{*}{Transformer} 
  & Swin-T    & \pesqtimebarstd{100}{84.25±9.96}    & \pesqtimebarstd{100}{96.86 ± 3.11}& \stoitimebarstd{29}{28.70 ± 26.23}  & \stoitimebarstd{5}{2.44 ± 0.86}  & \stoitimebarstd{63}{10.61 ± 6.02}  & \pesqtimebarstd{100}{67.02 ± 9.16}   \\
  & Swin-T V2 & \pesqtimebarstd{100}{55.65±18.45}   & \pesqtimebarstd{100}{92.70 ± 4.90}& \stoitimebarstd{29}{28.90 ± 25.99}  & \stoitimebarstd{5}{2.44 ± 0.77}  & \stoitimebarstd{63}{10.40 ± 6.54}  & \pesqtimebarstd{100}{62.61 ± 6.83}   \\
  & CRATE     &\pesqtimebarstd{100}{75.20±12.92}   & \pesqtimebarstd{100}{98.96 ± 1.83}& \stoitimebarstd{29}{28.66 ± 26.28}  & \stoitimebarstd{5}{2.44 ± 0.85}  & \stoitimebarstd{63}{24.93 ± 11.96} & \pesqtimebarstd{100}{57.92 ± 8.97}  \\ \hdashline
  \multirow{5}{*}{DWS}   
  & RF-Net    & \pesqtimebarstd{100}{55.33 ± 15.91}   & \pesqtimebarstdunderline{100}{99.00 ± 1.88}& N/A& \stoitimebarstd{5}{3.57 ± 0.98}  & \stoitimebarstdunderline{63}{8.37 ± 5.91}   & \pesqtimebarstd{100}{57.87 ± 5.28}   \\
  & STFNets   & \pesqtimebarstd{100}{50.61 ± 17.48}   & \pesqtimebarstd{100}{95.30 ± 4.14}& N/A& N/A    & \stoitimebarstd{63}{47.20 ± 21.44} & \pesqtimebarstd{100}{58.59 ± 7.59}    \\
  & SLNet     & \pesqtimebarstd{100}{73.67 ± 13.11}   & \pesqtimebarstd{100}{94.79 ± 4.14}& N/A& N/A    & N/A     & N/A     \\
  & Widar3.0  & \pesqtimebarstd{100}{50.45 ± 18.29}  & \pesqtimebarstd{100}{82.01 ± 6.53}& N/A& N/A    & \stoitimebarstd{63}{62.40 ± 15.70} & N/A   \\
  & \textbf{\sysname}  & \pesqtimebarstdunderline{100}{82.10 ± 10.41}  & \pesqtimebarstd{100}{98.98 ± 1.91}& \stoitimebarstdunderline{29}{28.12 ± 27.02}  & \stoitimebarstdunderline{5}{2.44 ± 0.83}  & \stoitimebarstd{63}{17.09 ± 9.13}  & \pesqtimebarstdunderline{100}{60.07 ± 10.41}  \\ 
  \bottomrule
\end{tabular}
\label{tab:results}
\end{table*}

\subsection{Baselines}
We compare \sysname with seven baseline models that represent two major design paradigms in DWS:
(i) general-purpose transformer backbones adapted to RF inputs, and
(ii) RF-specialized architectures designed for domain-specific sensing tasks or domain shifts.
To ensure a fair comparison, all baselines are adapted to the same input representation (after modality-specific preprocessing) and trained under matched optimization settings whenever possible, including consistent data splits, training epochs, early-stopping criteria, and comparable parameter budgets.

\noindent \ding{182} \textit{General-purpose Transformers.}
We include Swin Transformer (Swin-T) \cite{liuSwinTransformerHierarchical2021b}, Swin Transformer V2 (Swin-T V2) \cite{liuSwinTransformerV22022}, and CRATE \cite{yuWhiteBoxTransformersSparse2023a}.
These models evaluate whether the performance gains of \sysname arise from generic transformer capacity or from its complex-domain, sparsity-driven white-box design.

\noindent \ding{183} \textit{DWS specialized models.}
We include STFNets \cite{yaoSTFNetsLearningSensing2019} (spatio-temporal-frequency modeling),
Widar3.0 \cite{zhengZeroEffortCrossDomainGesture2019} (cross-domain CSI gesture recognition),
RF-Net \cite{dingRFnetUnifiedMetalearning2020} (meta-learning for domain adaptation),
and SLNet \cite{yangSLNetSpectrogramLearning2023} (spectrogram-based RF representation learning).
These baselines embody established RF-specific inductive biases, including time–frequency structure, meta-learning for domain shift, and spectrogram-based feature extraction.

\subsection{Metrics}
We use task-appropriate evaluation metrics and report averages over the official test splits (or subject-disjoint splits for WiP datasets).

\noindent \ding{182} \textit{Classification tasks.}
For gesture recognition, activity recognition, and gait identification, we report \textbf{accuracy} (\%), defined as the fraction of correctly classified samples.
For datasets involving cross-domain or cross-environment evaluation (\eg, Widar3.0), we strictly follow the official protocols to ensure that reported results reflect generalization rather than memorization.

\noindent \ding{183} \textit{Regression tasks.}
For pose estimation, we report \textbf{MPJPE} (Mean Per-Joint Position Error), which measures the average Euclidean distance between estimated joints $\hat{\mathbf{j}}_i$ and ground-truth joints $\mathbf{j}_i$ over $J$ joints:
\begin{equation}
\text{MPJPE} = \frac{1}{J} \sum_{i=1}^{J} \left\| \hat{\mathbf{j}}_i - \mathbf{j}_i \right\|_2.
\end{equation}
For respiratory rate estimation, we report \textbf{BPME} (Breaths Per Minute Error), defined as the absolute difference between predicted and ground-truth breathing rates (in breaths/min).
Lower MPJPE and BPME indicate better performance.

\section{Performance}
\label{sec:performance}

To validate the effectiveness and generality of \sysname, we conduct comprehensive experiments across diverse sensing modalities and human-centric tasks, including gesture recognition, gait identification, activity recognition, 2D/3D pose estimation, and respiratory rate estimation.

\subsection{Performance across RF Modalities}

\head{WiFi CSI} 
We evaluate \sysname on four WiFi-based sensing tasks using three datasets.
\sysname directly processes complex-valued Doppler frequency shift (DFS) spectra, enabling it to fully exploit amplitude–phase coupling in RF signals.
In contrast, most baseline models rely on amplitude-only representations or real-valued phase–amplitude decompositions, which inevitably discard structural information and break the physical coherence of complex RF signals.

\noindent\textit{(1) Gesture recognition}: 
As shown in \tab\ref{tab:results} (1st column), \sysname performs competitively in gesture recognition on the Widar3.0 dataset. 
It outperforms most dedicated DWS baselines and remains competitive with strong transformer baselines, demonstrating robustness to diverse gesture patterns and cross-domain variations.

\noindent\textit{(2) Gait recognition}: As shown in \tab\ref{tab:results} (2nd column), \sysname achieves an accuracy of 98.98\%, closely matching the best-performing baseline, RF-Net (99.00\%).
This result highlights \sysname's ability to capture fine-grained identity-related motion patterns while maintaining architectural transparency.

\noindent\textit{(3) 3D Pose estimation}: As shown in \tab\ref{tab:results} (3rd column), \sysname achieves performance comparable to strong baselines on WiP-pose.
Notably, several DWS baselines cannot be applied to this task due to strict input-size assumptions or modality-specific architectures.
In contrast, \sysname generalizes naturally to continuous-valued regression tasks, demonstrating its flexibility across heterogeneous sensing scenarios.

\begin{figure*}[t]
  \begin{minipage}{0.34\textwidth}
    \centering
    \includegraphics[width=1.0\textwidth]{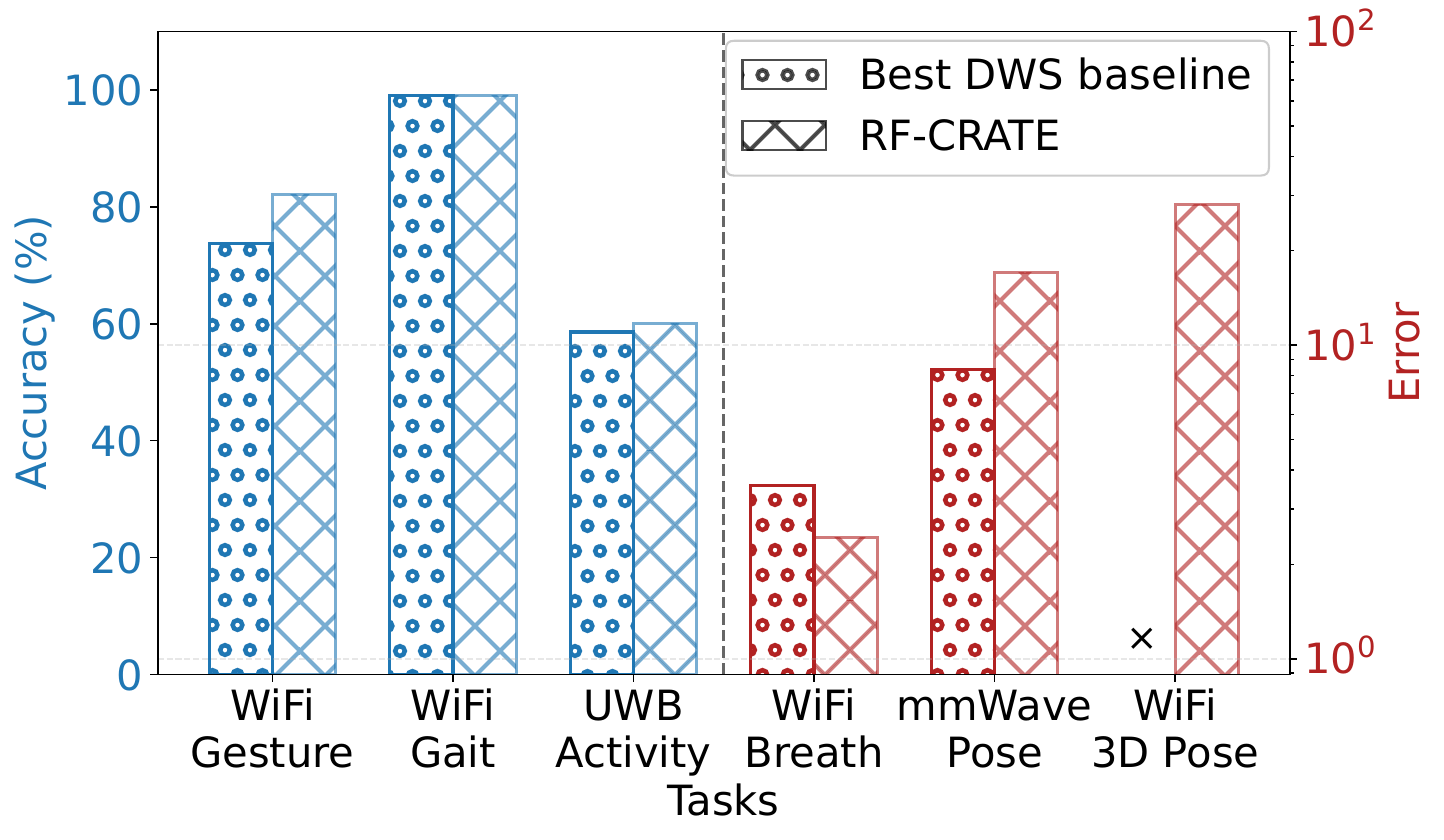}
 \vspace{-1.5\baselineskip}
  \caption{\sysname vs. DWS baselines. 
  }
\label{fig:V2_compare_to_baselines}
\end{minipage}
\hfill
\begin{minipage}{0.3\textwidth}
  \centering
  \includegraphics[width=1.0\textwidth]{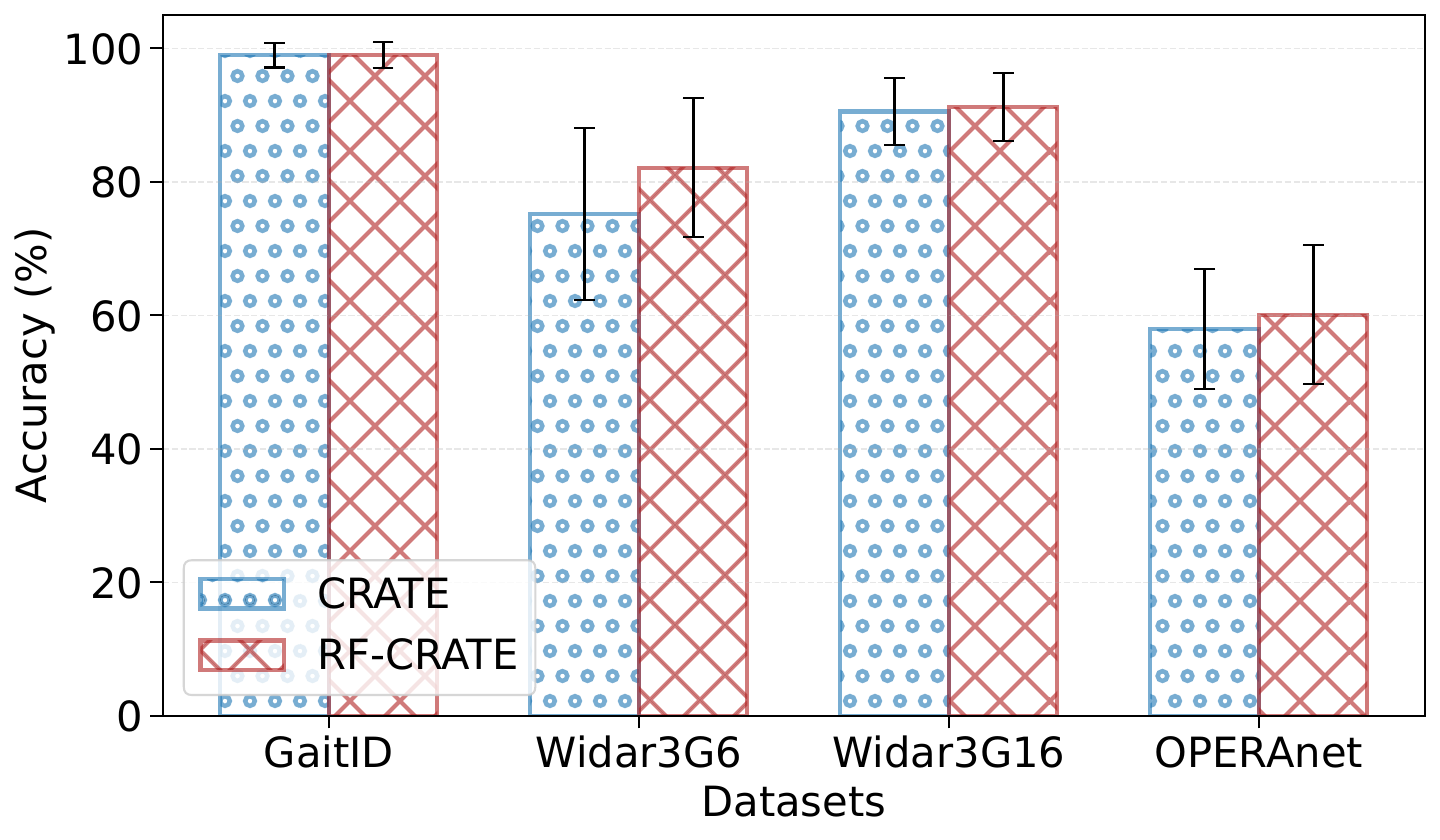}
  \vspace{-1.5\baselineskip}
  \caption{\sysname vs. CRATE on classification tasks.}
  \label{fig:V2_compare_classification}
\end{minipage}
\hfill
\begin{minipage}{0.3\textwidth}
\centering
  \includegraphics[width=1.0\textwidth]{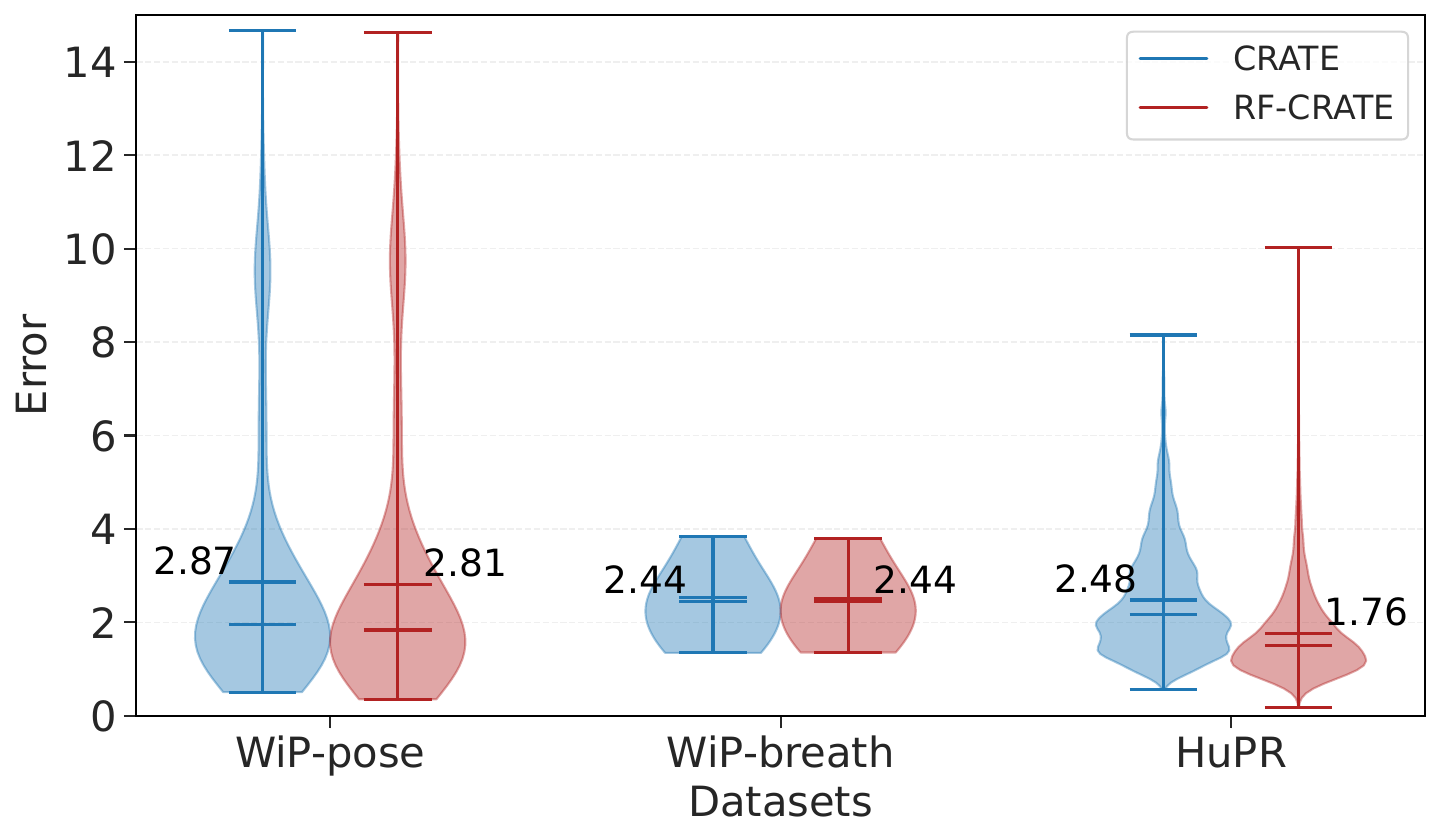}
  \vspace{-1.5\baselineskip}
  \caption{\sysname vs. CRATE on regression tasks.}
  \label{fig:V2_compare_regression}
\end{minipage}
\end{figure*}

\noindent\textit{(4) Respiratory rate estimation}: As shown in \tab\ref{tab:results} (4th column), \sysname achieves performance comparable to the best baseline, Swin-T V2.
This result supports \sysname's effectiveness in capturing subtle periodic physiological signals from noisy WiFi measurements.

\head{FMCW mmWave} 
We evaluate \sysname on mmWave-based 2D pose estimation using the HuPR dataset.
As shown in \tab\ref{tab:results} (5th column), \sysname outperforms CRATE, STFNets, and Widar3.0 in terms of MPJPE, demonstrating superior spatial inference capability from high-frequency radar signals.
These results highlight the advantage of complex-domain modeling for mmWave sensing, where phase information is tightly coupled with spatial geometry.

\head{IR-UWB}
We further evaluate \sysname on UWB-based activity recognition using the OPERAnet dataset.
As shown in the last column of \tab\ref{tab:results}, \sysname achieves competitive accuracy and outperforms several baselines.
This confirms that \sysname generalizes beyond WiFi and mmWave to impulse-radio sensing modalities, despite significant differences in signal structure and noise characteristics.

\begin{figure*}[t]
  \begin{minipage}{0.63\textwidth}
    \centering
    \includegraphics[width=0.95\textwidth]{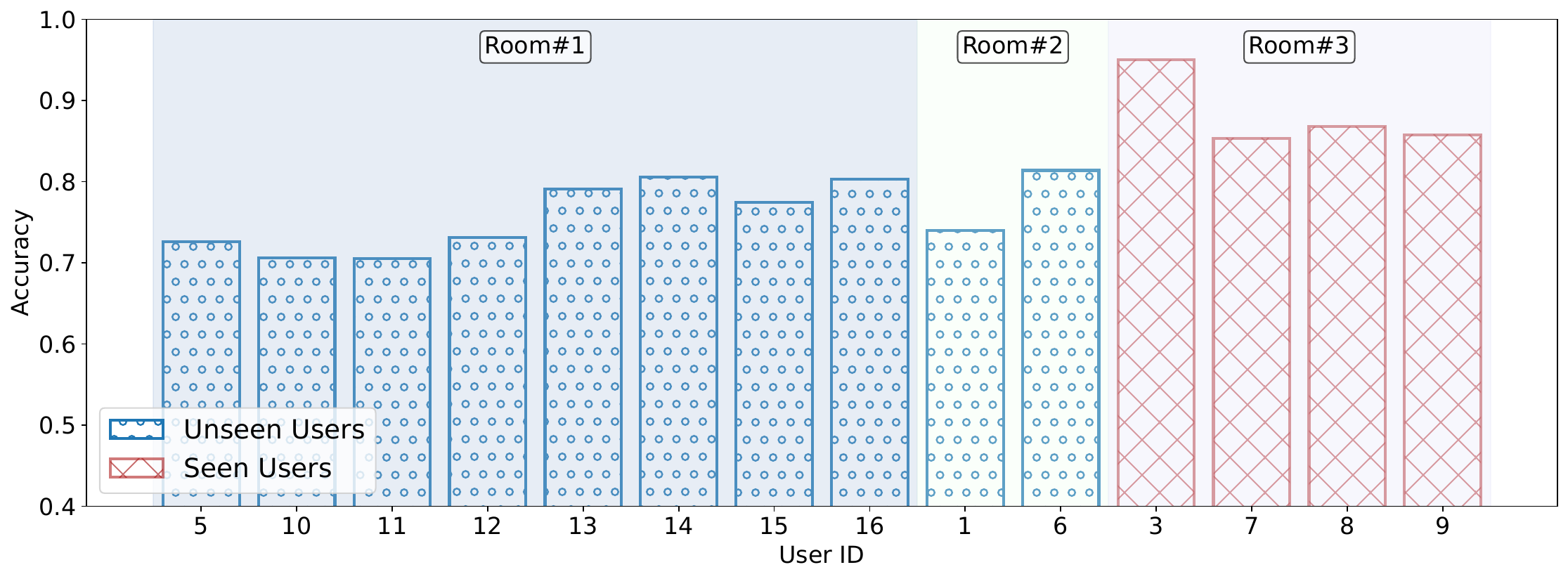}
    \caption{Performance evaluation of \sysname across multiple users and environments.}
    \label{fig:V2_cross_user_room}
  \end{minipage}
  \hfill
  \begin{minipage}{0.32\textwidth}
  \centering
    \includegraphics[width=1.0\textwidth]{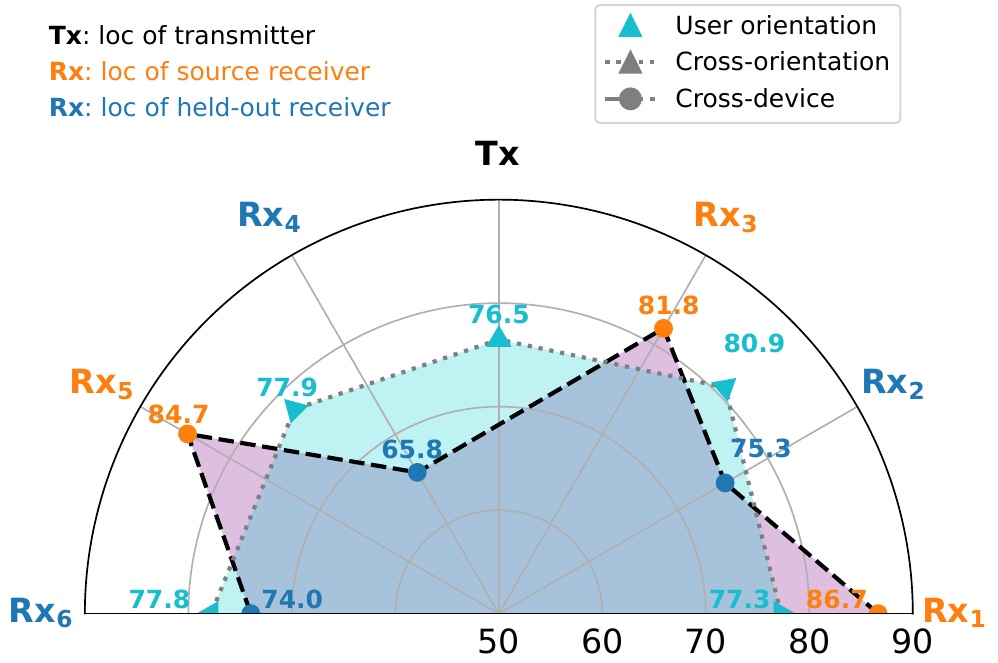}
    \caption{\sysname performance across different devices and orientations.}
    \label{fig:V2_cross_device_orientation}
  \end{minipage}
\end{figure*}

\subsection{Key Observations}
\label{ssec:key_observation}
\noindent\textbf{(1) Competitive performance with a white-box architecture.}
\sysname consistently achieves competitive performance while maintaining a fully white-box architecture.
As shown in \fig\ref{fig:V2_compare_to_baselines}, \sysname matches or exceeds several dedicated DWS baselines and remains in the performance range of strong generic transformer backbones across a wide range of sensing tasks.
These findings suggest that mathematical transparency need not preclude strong empirical performance, and that optimization-derived white-box architectures can serve as a viable alternative to empirically engineered deep models in wireless sensing.

\noindent\textbf{(2) Performance gains from complex-domain modeling.}
Extending \sysname to the complex domain yields substantial improvements over the original CRATE across all evaluated tasks.
As shown in \fig\ref{fig:V2_compare_classification} and \fig\ref{fig:V2_compare_regression}, \sysname achieves an average gain of $3.39\%$ in classification accuracy and reduces regression error by $10.34\%$.
\orev{While \sysname remains consistently competitive, the magnitude of these improvements is inherently dependent on task complexity and the available performance headroom. On near-saturated datasets such as GaitID, the baseline CRATE already approaches ceiling performance, leaving limited room for measurable gains. In contrast, the advantages of coherent complex structures become more pronounced in challenging scenarios where domain shifts and fine-grained regression are prevalent, such as cross-domain testing on Widar3G6 and human pose regression on HuPR. We further acknowledge that complex domain modeling is not free; the increased representational capacity can make optimization more difficult under severe data scarcity, and performance gains are not universally guaranteed, a finding supported by our ablation study on the complex extension of the Swin-T model (detailed in \sec\ref{ssc:ablation_study}). 
}
In addition, although complex arithmetic introduces a modest $\sim\!2\times$ increase in FLOPs, the parameter count remains nearly identical to CRATE, preserving memory efficiency.
These consistent gains outweigh the computational overhead and demonstrate the practical value of complex-domain modeling for wireless sensing.
Overall, the results position \sysname as an effective and principled white-box framework for DWS.

\subsection{Robustness Evaluation}
Wireless sensing systems are inherently affected by multi-path propagation, where RF signals interact with both the target subject and surrounding environments.
Such interactions introduce domain shifts across users, environments, devices, and orientations, posing fundamental challenges to reliable wireless sensing \cite{zhengZeroEffortCrossDomainGesture2019, zhangWiFibasedCrossDomainGesture2021}.
To assess whether \sysname can generalize beyond controlled conditions, we systematically evaluate its robustness under multiple domain-shifting factors using the Widar3.0 dataset.

\head{Cross-user and cross-environment performance}
We first evaluate \sysname under user and environment shifts.
As shown in \fig\ref{fig:V2_cross_user_room}, \sysname maintains consistently high accuracy when tested on previously unseen users.
Moreover, although trained exclusively on data from Room \#3, \sysname generalizes well to two unseen rooms with only minor performance degradation.
These results indicate that \sysname captures motion-relevant signal structures that are invariant to individual differences and environmental multi-path patterns, demonstrating strong cross-user and cross-environment robustness.

\head{Cross-device and cross-orientation performance}
We further evaluate hardware and spatial robustness by training \sysname on data from three source receivers ($\text{Rx}_1$, $\text{Rx}_3$, $\text{Rx}_5$) and testing it on all receivers, including held-out devices ($\text{Rx}_2$, $\text{Rx}_4$, $\text{Rx}_6$).
As illustrated in \fig\ref{fig:V2_cross_device_orientation}, where receivers and the transmitter are symmetrically distributed around the user, \sysname achieves strong performance on both source and unseen devices with minimal degradation.
In addition, \sysname exhibits stable accuracy across diverse user orientations, indicating robustness to spatial configuration changes.
Together, these results demonstrate that \sysname is resilient to hardware heterogeneity and positional variation, supporting its applicability in real-world deployment scenarios.

\head{Discussion}
The observed robustness can be attributed to \sysname's complex-domain modeling and sparsity-driven representation learning.
By preserving amplitude–phase coupling and explicitly organizing features into structured subspaces, \sysname learns representations that are less sensitive to domain-specific perturbations.
This property distinguishes \sysname from conventional real-valued or amplitude-only models and highlights the practical advantage of mathematically grounded white-box architectures in wireless sensing.

\begin{figure*}[t]
  \begin{minipage}{0.3\textwidth}
    \centering
    \includegraphics[width=1\textwidth]{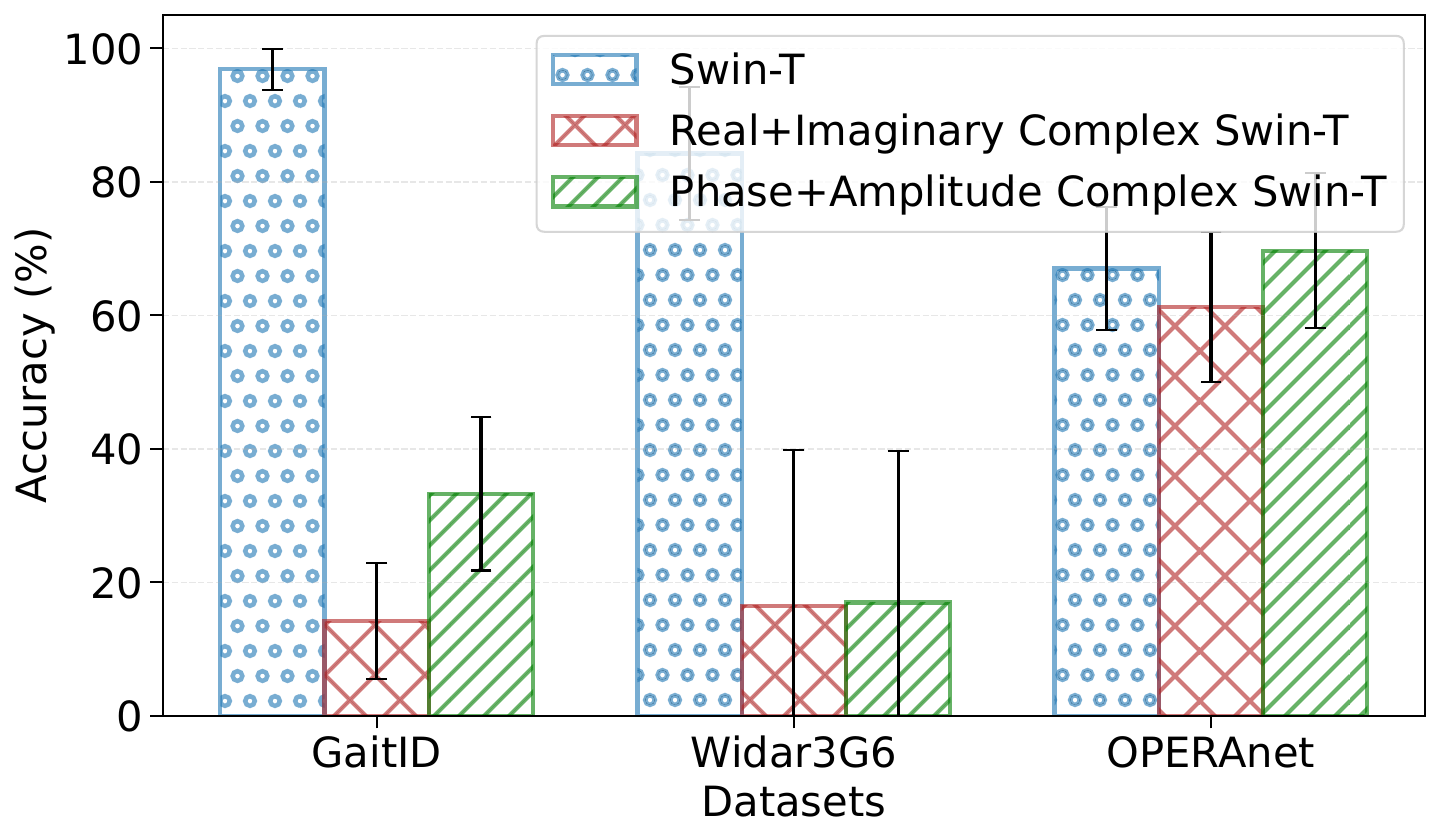}
    \vspace{-0.3in}
    \caption{\orev{Swin-T vs. Complex Swin-T on classification tasks.}}
    \label{fig:ablation_swint_class}
  \end{minipage}
  \hfill
  \begin{minipage}{0.3\textwidth}
    \centering
    \includegraphics[width=1\textwidth]{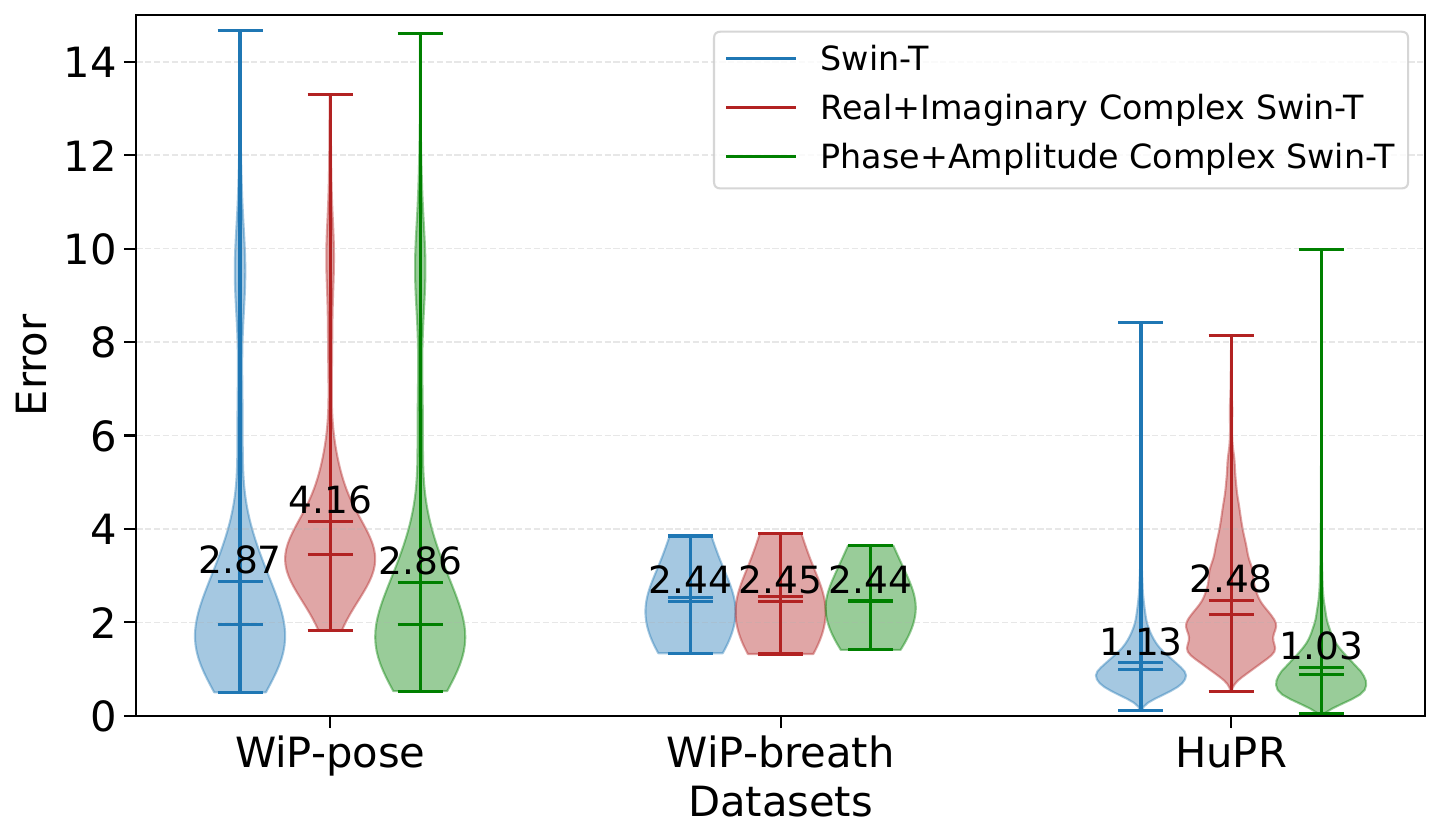}
    \vspace{-0.25in}
    \caption{\orev{Swin-T vs. Complex Swin-T on regression tasks.}}
    \label{fig:ablation_swint_regre}
  \end{minipage}
  \hfill
  \begin{minipage}{0.3\textwidth}
    \centering
    \includegraphics[width=1\textwidth]{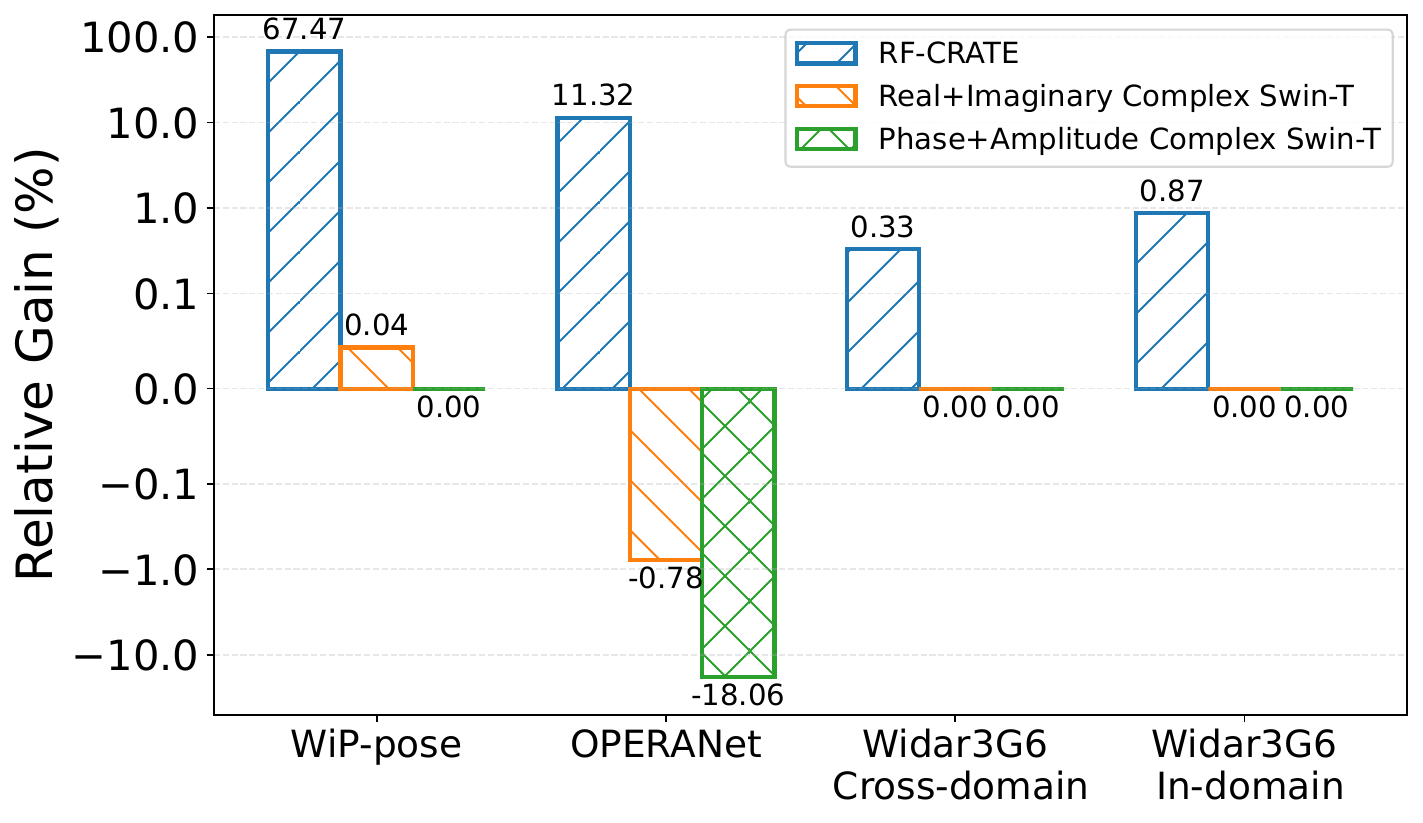}
    \vspace{-0.3in}
    \caption{\orev{Performance gains with SSR.}}
    \label{fig:ablation_ssr_complete}
  \end{minipage}
\end{figure*}

\subsection{Ablation Study}
\label{ssc:ablation_study}
We conduct systematic ablation studies to quantify the contribution of key design components in \sysname and to understand how architectural and signal-processing choices affect performance.

\orev{
\head{Impact of complex-valued design}
To determine whether the performance gains reported in \sec\ref{ssec:key_observation} stem merely from the use of complex-valued inputs or also account for the complex white-box structure of \sysname, we conduct experiments with two complex domain adaptations of the standard black-box Swin Transformer (Swin-T). 
The first variant, the Real+Imaginary Complex Swin-T, extends the architecture to handle complex inputs via a dual-path design that processes real ($\Re$) and imaginary ($\Im$) components simultaneously, deriving the final output from the magnitude of the feature logits.
The second variant, the Phase+Amplitude Complex Swin-T, encodes complex inputs into three real-valued channels representing amplitude ($|x|$) and the sine and cosine of the wrapped phase ($\sin(\angle x), \cos(\angle x)$) for numerical stability.
We evaluate these models across all datasets and compare their results with those of the original Swin-T, as summarized in \fig\ref{fig:ablation_swint_class} and \fig\ref{fig:ablation_swint_regre}. These results lead to the following conclusion: \textit{the complex domain alone does not guarantee performance gains}. While prior works (\eg, SLNet \cite{yangSLNetSpectrogramLearning2023}, RF-Diffusion \cite{chi2024rf}) suggest that phase or complex-aware modeling is beneficial, our controlled experiments indicate that operating natively in the complex domain without an appropriate architectural design does not guarantee improvements. As shown in \fig\ref{fig:ablation_swint_class}, both complex-extended Swin-T variants significantly underperform compared to the standard real-valued Swin-T on classification tasks such as gait recognition (GaitID) and WiFi gesture recognition (Widar3G6). In the regression tasks shown in \fig\ref{fig:ablation_swint_regre}, the Phase+Amplitude Complex Swin-T matches or slightly exceeds the performance of standard Swin-T, whereas the Real+Imaginary Complex Swin-T consistently demonstrates performance degradation. 
}

\begin{figure*}[t]
  \begin{minipage}{0.3\textwidth}
    \centering
    \includegraphics[width=1\textwidth]{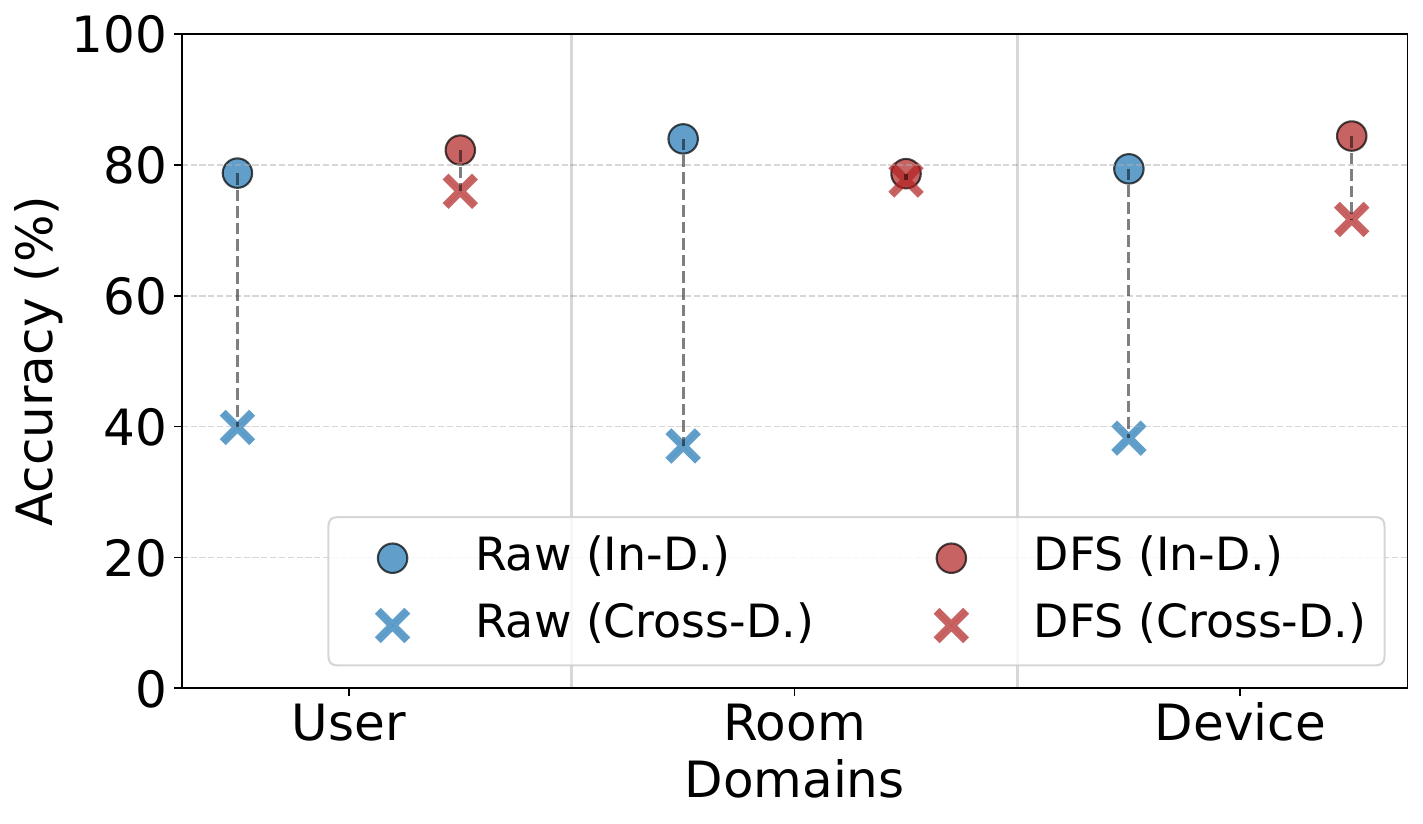}
    \vspace{-0.3in}
    \caption{\sysname performances with raw signal input or DFS spectrum.}
    \label{fig:ablation_raw_dfs}
  \end{minipage}
  \hfill
  \begin{minipage}{0.3\textwidth}
    \centering
    \includegraphics[width=1\textwidth]{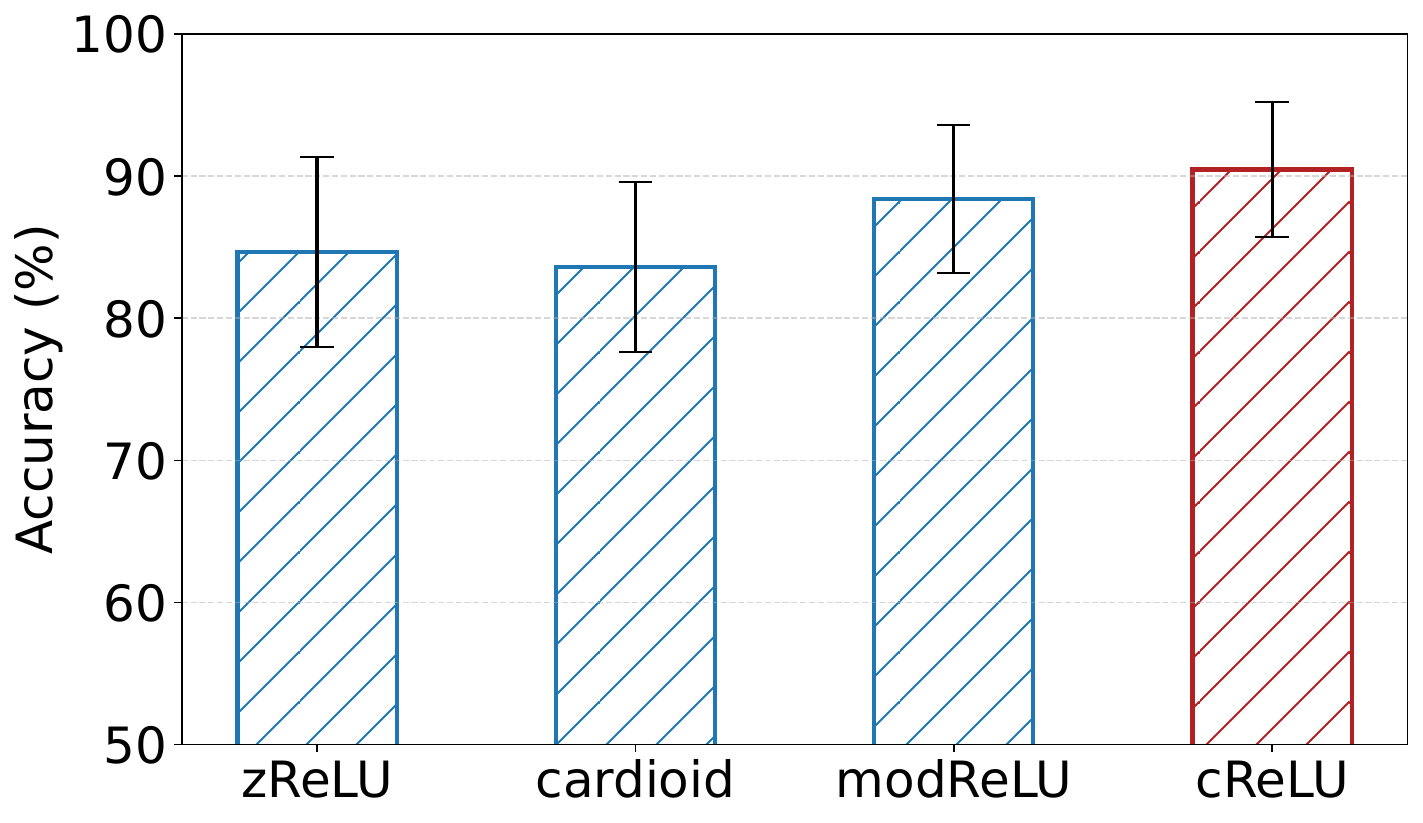}
    \vspace{-0.25in}
    \caption{\sysname performance w.r.t. complex ReLU functions.}
    \label{fig:ablation_relu}
  \end{minipage}
  \hfill
  \begin{minipage}{0.3\textwidth}
    \centering
    \includegraphics[width=1\textwidth]{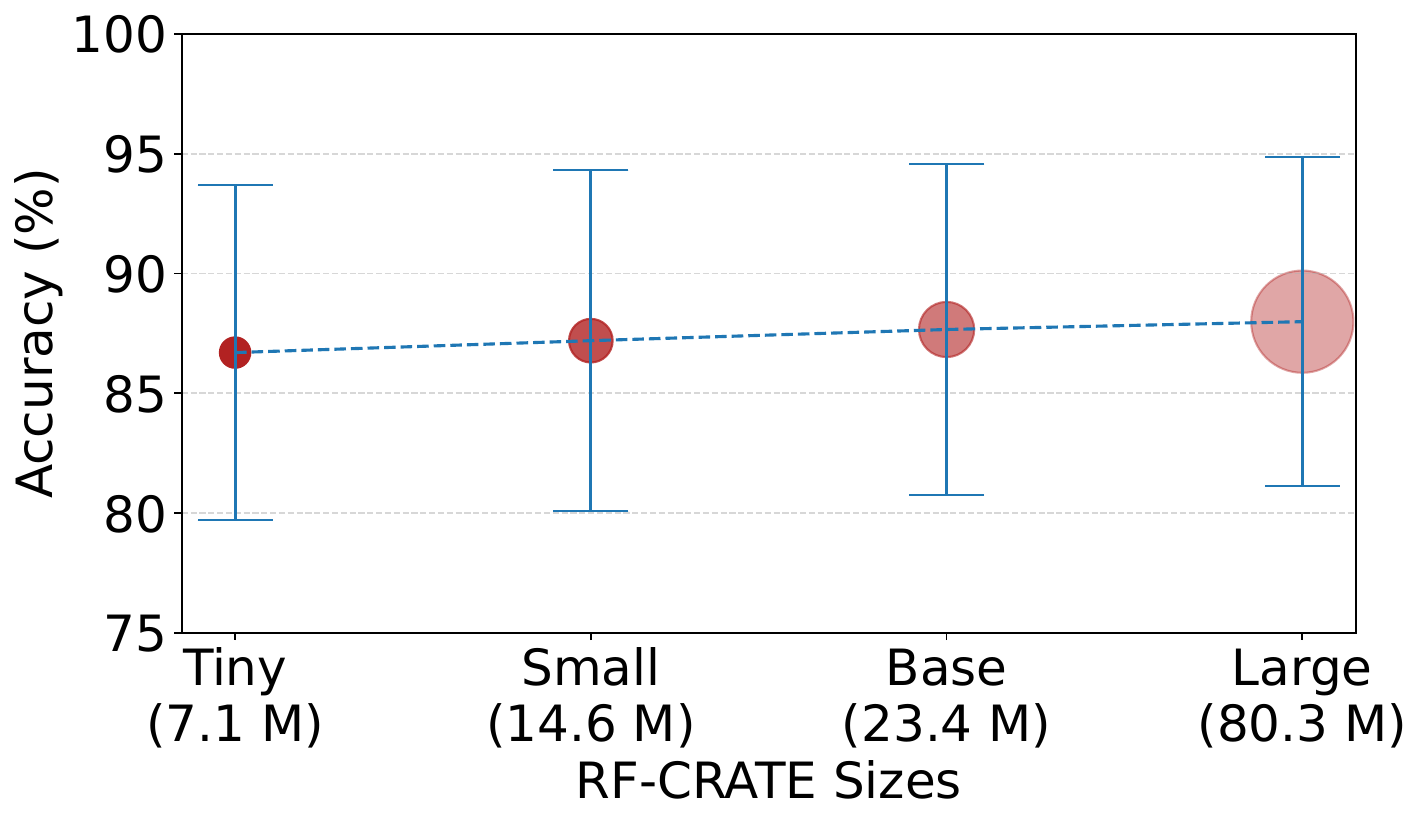}
    \vspace{-0.3in}
    \caption{\sysname performance w.r.t. different model sizes.}
    \label{fig:ablation_size}
  \end{minipage}
\end{figure*}

\orev{
\head{Impact of subspace regularization (SSR)}
We evaluate the impact of SSR on both \sysname and the complex-extended Swin-T baselines. As shown in \fig\ref{fig:ablation_ssr_complete}, incorporating SSR into \sysname yields a significant average performance improvement of 19.98\% across four datasets. In contrast, the effect of SSR on the complex-extended Swin-T variants is either neutral or detrimental.
The efficacy of SSR is grounded in the white-box theory of \sysname rather than mere empirical observation. \sysname is derived from the principle of complex sparse rate reduction, which mathematically dictates that the token embedding dimension be partitioned into independent "subspaces." In this framework, as discussed in \sec\ref{ssec:model_analysis_ssr}, SSR acts as a theory-based constraint that ensures these subspaces are utilized efficiently and prevents "subspace collapse." 
Conversely, in complex-extended Swin-T variants, features are constructed through a hierarchical process of patch-merging and localized windowed attention. Consequently, the "subspaces" targeted by SSR are merely post-hoc partitions of a flattened feature vector rather than functionally distinct architectural components. Because SSR lacks structural alignment with the natural decomposition of the complex-extended Swin-T backbone, it acts as a mis-specified constraint that can hinder optimization.
In summary, these results establish SSR as a \textit{theory-based} regularization strategy that is distinct from generic heuristics. Its success stems from the mathematical alignment between the regularization objective and the subspace structure inherent to complex sparse rate reduction. This finding highlights a core advantage of the white-box paradigm: its mathematically observable internal structure provides principled "hooks" for targeted regularization that directly enhance representation diversity and model generalization.
}

\head{Raw CSI vs. DFS spectrum}
We analyze the impact of input representations using the Widar3.0 dataset.
As shown in \fig\ref{fig:ablation_raw_dfs}, models trained on raw CSI exhibit significant performance degradation under domain shifts (Cross-D.) compared to in-domain (In-D.) settings.
In contrast, the Doppler frequency shift (DFS) spectrum obtained via STFT preserves stable performance across domains.
This result confirms that frequency-domain representations capture motion-relevant signal structures that are more invariant to environmental multi-path and device-specific distortions.
Importantly, this benefit persists even for deep models capable of processing raw signals, indicating that principled signal transformations remain valuable in DWS.

\head{Complex activation functions.}
We evaluate the effect of different complex-valued activation functions on the Widar3.0 gesture recognition task.
As shown in \fig\ref{fig:ablation_relu}, the theoretically derived cReLU function,
\( \rho(\Re(z)) + i \cdot \rho(\Im(z)) \),
achieves the highest accuracy of 90.47\%.
In comparison, zReLU \cite{gubermanComplexValuedConvolutional2016}, modReLU \cite{arjovskyUnitaryEvolutionRecurrent2016a}, and cardioid \cite{virtueBetterRealComplexvalued2017} achieve 84.64\%, 83.61\%, and 88.40\%, respectively.
These results indicate that treating real and imaginary components symmetrically, as dictated by the model derivation, is crucial for preserving expressive power in complex-valued RF representations.

\head{Model capacity.}
We evaluate \sysname under four model scales: Tiny (7.1M parameters), Small (14.6M), Base (23.4M), and Large (80.3M).
As shown in \fig\ref{fig:ablation_size}, performance increases only marginally with model size: 86.70\% (Tiny), 87.20\% (Small), 87.66\% (Base), and 87.99\% (Large) on Widar3G6.
This suggests that wireless sensing tasks are constrained more by signal structure than model capacity.
Considering the modest gains and computational overhead of larger models, we adopt the Small configuration as the default setting, achieving a favorable efficiency–performance trade-off.

\begin{figure*}[t]
    \centering
    \includegraphics[width=0.95\textwidth]{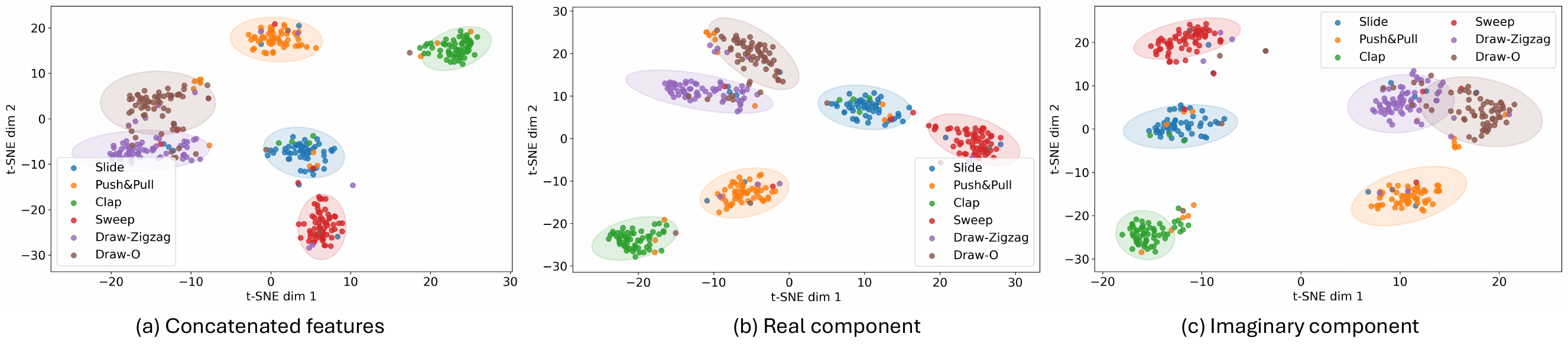}
    \vspace{-0.1in}
    \caption{\orev{t-SNE visualization of learned complex-valued features. \rm{(a) Distribution of concatenated complex features ($\Re \Vert \Im$) across gesture categories. The high separability demonstrates that the model successfully captures semantic nuances. (b, c) Independent visualizations of the real and imaginary components. Both dimensions exhibit category-centered clusters, confirming that the full complex representation is essential for discriminative sensing.}}}
    \label{fig:feature_distribution}
\end{figure*}

\begin{figure*}[t]
    \centering
    \includegraphics[width=0.95\textwidth]{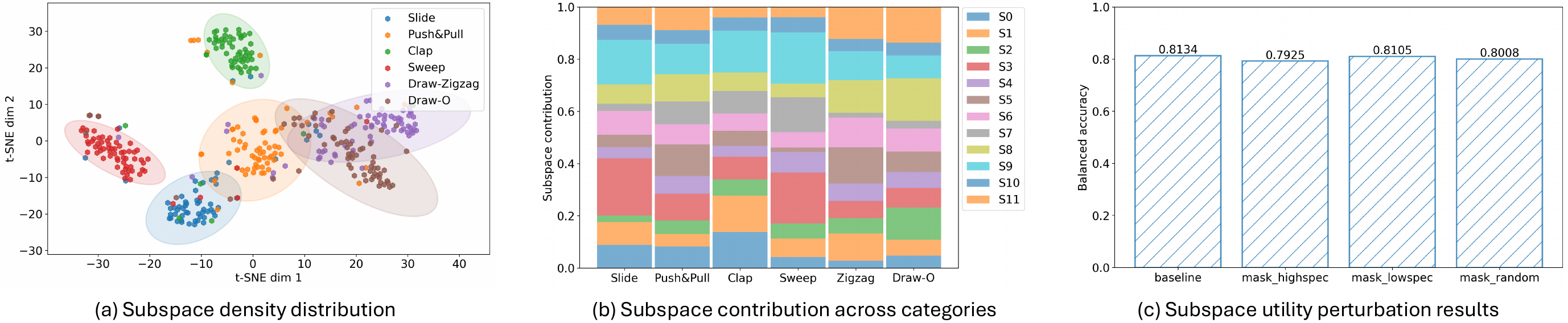}
    \vspace{-0.1in}
    \caption{\orev{Analysis of subspace density, specialization, and utility. \rm{(a) t-SNE of subspace-density vectors $\boldsymbol{\rho}_{n}$, revealing continuous region-based distributions that preserve task semantics. (b) Class-level subspace contribution, demonstrating that gestures rely on a distributed combination of subspaces rather than single-path detectors. (c) Perturbation analysis: masking highly specialized subspaces causes a sharper performance drop than masking least-specialized ones, providing functional evidence of subspace utility.}}}
    \label{fig:subspace_analysis}
\end{figure*}

\head{RF-MLP vs. RF-ISTA.}
Finally, we compare the proposed RF-MLP module with RF-ISTA, a direct complex-valued extension of CRATE \cite{yuWhiteBoxTransformersSparse2023a}.
On the HuPR dataset, RF-MLP achieves an MPJPE of 24.93, nearly identical to RF-ISTA (24.94).
This result confirms that RF-MLP preserves performance while offering a fully principled derivation and simpler implementation.
Unlike RF-ISTA, where residual connections are introduced heuristically, RF-MLP emerges naturally from the optimization formulation, reinforcing the interpretability and completeness of \sysname's architecture.

\orev{
\subsection{Post-training Model Analysis}
To bridge the theoretical derivation of \sysname with the intrinsic parsimony of wireless sensing, we analyze the trained model using the WiFi-based gesture recognition task (Widar3G6). 
Our analysis focuses on three dimensions: semantic feature distribution, internal subspace dynamics, and qualitative physical attention alignment.
}

\orev{
\head{Feature distribution and semantic}
We first investigate whether the learned representations encode categorical semantics. 
As shown in \fig\ref{fig:feature_distribution}(a), we apply t-SNE to the complex-valued CLS tokens from the final transformer block. 
The visualization reveals clear categorical clusters with minimal overlap.
Notably, as shown in \fig\ref{fig:feature_distribution}(b, c), both the real and imaginary components are independently separable.
This suggests that \sysname effectively utilizes the full complex domain to preserve task-relevant information.
At the same time, we emphasize that t-SNE is only a qualitative dimensionality-reduction tool: it approximately preserves local neighborhoods but can distort global geometry. Accordingly, these plots should be interpreted as illustrative evidence of separability rather than proof that the underlying latent features are compact, causal, or inherently explainable.
The minor overlap between the \textit{Draw-Zigzag} and \textit{Draw-O} classes is physically consistent with the inherent similarity in their DFS spectrum, as further discussed in the attention analysis.
}

\orev{
\head{Subspace density, specialization, and utility analysis}
We further analyze the internal subspace states to understand how \sysname organizes its parsimonious representations.

\noindent\ding{182} \textit{Semantic structure in subspace-density space.}
We define the subspace density vector for a sample $n$ as $\boldsymbol{\rho}_{n} = [\rho_{n,1}, \rho_{n,2}, \ldots, \rho_{n,K}] \in \mathbb{R}^{K}$, where $\rho_{n,k} = \sqrt{\sum_{i=1}^{D} |\mathbf{z}_{k,i}|^{2}}$ represents the $L_2$ energy of the $k$-th subspace and $D$ is the feature dimension. 
As shown in \fig\ref{fig:subspace_analysis}(a), these vectors form \emph{region-based} clusters.
This pattern indicates that the density representation captures both task semantics (gesture-specific regions) and natural intra-class variations (the spread within regions caused by differences in execution speed or amplitude).

\noindent\ding{183} \textit{Distributed subspace contribution.}
To determine if subspaces are specialized for specific gestures, we compute the class-level subspace mass $M_{c,k} = \frac{1}{|\mathcal{D}_{c}|}\sum_{n \in \mathcal{D}_{c}} \rho_{n,k}$, where $\mathcal{D}_c$ is the set of samples in class $c$.
The normalized contribution $P(k \mid c) = \frac{M_{c,k}}{\sum_{j=1}^{K} M_{c,j}}$ shown in \fig\ref{fig:subspace_analysis}(b) demonstrates that feature mass is distributed across nearly all subspaces for every class.
This suggests that \sysname encodes task semantics in a distributed manner, where multiple subspaces jointly contribute to the final representation rather than acting as isolated class detectors.

\noindent\ding{184} \textit{Subspace utility via perturbation.}
We evaluate the functional importance of these subspaces by masking specific dimensions of $\boldsymbol{\rho}_{n}$ and measuring the drop in nearest-centroid classification accuracy.
We define a specialization score $\mathrm{SpecScore}_k = D_k \cdot (1 - H_k)$, where $D_k = \max_{c} P(c \mid k)$ is the dominance and $H_k = - \frac{1}{\log C}\sum_{c=1}^{C} P(c \mid k)\log P(c \mid k)$ is the normalized entropy.
As shown in \fig\ref{fig:subspace_analysis}(c), masking the top-2 specialized subspaces causes a significantly larger accuracy drop ($0.8134 \to 0.7925$) compared to masking the bottom-2 ($0.8134 \to 0.8105$). 
This confirms that while subspaces are largely shared, those with higher specialization play a more critical role in preserving discriminative features.
}

\begin{figure*}[t]
    \centering
    \includegraphics[width=0.95\textwidth]{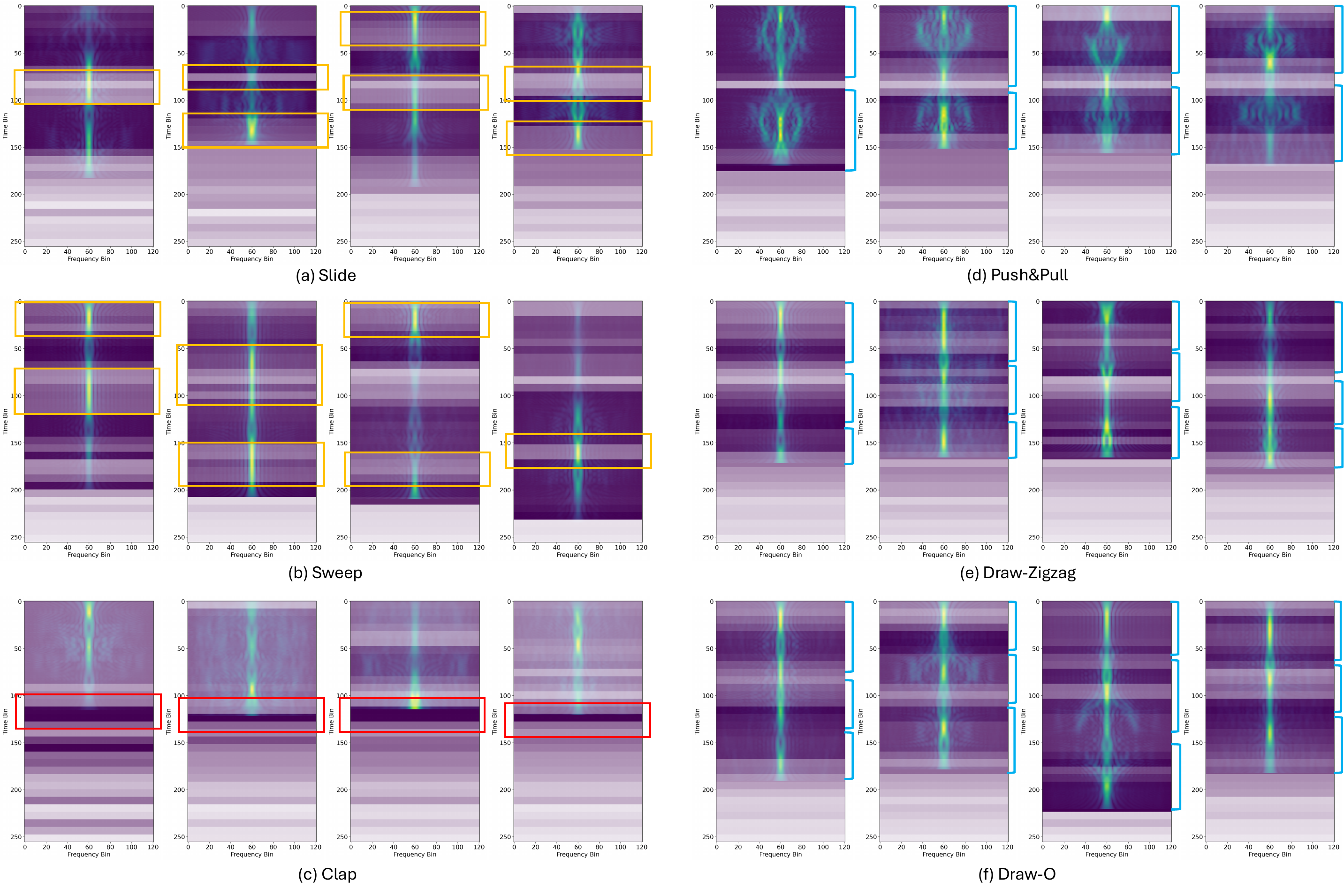}
    \vspace{-0.1in}
    \caption{\orev{Qualitative attention patterns on the DFS spectrum. \rm{Attention values are visualized as the alpha (transparency) channel; clear regions indicate higher attention. (a, b) The model suppresses static $0$~Hz components (yellow boxes). (c) For short-duration \textit{Clap} gestures, the model exploits the temporal boundary between the signal and zero-padding (red box). (d) \textit{Push \& Pull} exhibits a two-stage attention pattern aligned with physical motion phases. (e, f) \textit{Draw-Zigzag} and \textit{Draw-O} show similar three-stage signatures, explaining their semantic proximity in feature space.}}}
    \label{fig:attention_vis}
\end{figure*}

\orev{
\head{Attention and physical grounding}
To probe whether \sysname's learned attention exhibits physically meaningful patterns, we visualize the attention weights on the input DFS spectrum (\fig\ref{fig:attention_vis}).
The transparency (alpha) levels represent normalized attention values.
First, the model naturally suppresses the high-energy components around $0$~Hz (\fig\ref{fig:attention_vis}(a, b), yellow boxes), which correspond to static environmental reflections with no dynamic information. 
Second, for the \textit{Push \& Pull} gesture (\fig\ref{fig:attention_vis}(d)), the attention exhibits a clear two-stage pattern that matches the physical phases of the movement. 
For \textit{Draw-Zigzag} and \textit{Draw-O} (\fig\ref{fig:attention_vis}(e, f)), the model identifies similar three-stage temporal signatures, which explains the subtle feature-space overlaps noted earlier.
Interestingly, for the \textit{Clap} gesture (\fig\ref{fig:attention_vis}(c)), the model identifies a ``shortcut'' by attending to the transition between the signal and the zero-padded tail (red box). While effective for classification, this observation suggests that future research should explore more robust padding strategies or native variable-length sequence modeling to ensure the model remains focused strictly on the biomechanical signal. 
These visualizations should be interpreted with care. Attention maps are supportive qualitative evidence rather than definitive explanations of model reasoning; they do not by themselves establish causality, nor do they show that \sysname is uniquely more explainable than conventional transformer models with attention visualizations. 
Accordingly, we interpret these results as qualitative evidence of physical grounding, rather than as a definitive claim that \sysname provides superior explainability relative to standard transformer baselines.
Overall, these results suggest that \sysname's mathematically derived structure can yield attention patterns that are plausibly aligned with salient physical aspects of human motion.
}

\section{Discussion}
\label{sec:discussion}

This work introduces a mathematically grounded white-box framework for DWS.
Beyond performance gains, \sysname raises broader questions about how mathematically transparent models can reshape the design, deployment, and understanding of learning-based sensing systems.
We discuss key implications and future directions from computational, methodological, and application perspectives.

\head{Computational efficiency and real-world deployment}
Like transformer-based models, which have become the de facto backbone of modern learning-based systems, \sysname incurs computational costs that grow with input sequence length.
While our patching strategy substantially reduces the effective sequence length, real-time deployment on edge devices remains challenging.
This limitation is especially relevant for continuous sensing applications such as smart homes, healthcare monitoring, and interactive environments, where low latency and energy efficiency are paramount.
Future work may explore lightweight white-box variants, structured sparsification, and hardware-aware optimization to narrow the gap between theoretical interpretability and practical deployability.

\head{From black-box learning to physically grounded model design}
\sysname demonstrates that deep wireless sensing models can be derived from explicit complex-domain mathematical principles rather than empirical architectural heuristics.
This opens a new design paradigm in which model structures are co-designed with RF propagation theory, multi-path modeling, and classical signal processing methods.
Such physically grounded models may offer improved generalization across users, environments, and devices—an essential requirement for sensing systems operating in unconstrained real-world settings.
More broadly, this paradigm suggests that interpretability and performance need not be opposing goals in wireless sensing.

\head{Towards human-perceptible interpretability}
While \sysname provides architectural and mathematical transparency, its internal representations and decision processes are not yet directly interpretable at the human semantic level.
For human-centric sensing applications, interpretability should extend beyond equations to explanations that align with human concepts such as body motion, posture, activity semantics, and environmental context.
Future research may integrate visualization, causal analysis, and symbolic abstractions to bridge the gap between mathematical interpretability and human-understandable explanations.
Such capabilities are crucial for building trust in wireless sensing systems deployed in sensitive domains such as healthcare, elder care, and smart environments.

\head{Implications for wireless sensing research}
More broadly, \sysname suggests a shift in how DWS systems can be designed and understood.
Rather than treating learning-based wireless sensing as a purely data-driven pattern recognition problem, white-box models encourage a principled integration of physics, optimization, and sensing semantics.
This shift may lead to DWS systems that are not only accurate, but also more analyzable and adaptable to diverse human contexts.
We believe that such models represent an important step towards more transparent and sustainable DWS systems.

\section{Related Works}
\label{sec:related_works}

\head{Deep wireless sensing}
Early wireless sensing systems primarily relied on signal processing and classical machine learning techniques to extract handcrafted features from RF signals \cite{wuGaitWayMonitoringRecognizing2021, wuRFbasedInertialMeasurement2019, qianWidar2PassiveHuman2018, wangEeyesDevicefreeLocationoriented2014, zengWiWhoWifibasedPerson2016, puWholehomeGestureRecognition2013}. 
With the success of deep learning in vision and language, recent work has increasingly adopted neural models for RF-based human sensing \cite{zhengZeroEffortCrossDomainGesture2019, Lee_2023_WACV_hupr, zhao2018through, zhao2018rf, zhengMoReFiMotionrobustFinegrained2021, zhangCrossSenseCrossSiteLargeScale2018, jiangEnvironmentIndependentDevice2018, chenMoViFiMotionrobustVital2021, xiaoCsiGANRobustChannel2019, songRFURLUnsupervisedRepresentation2022, yaoSTFNetsLearningSensing2019, liUniTSShortTimeFourier2021, dingRFnetUnifiedMetalearning2020, yangSLNetSpectrogramLearning2023, guoContactlessFinegrainedCardiac2025, wangRFGymCareIntroducingRespiratory2024, shengMetaFormerDomainAdaptiveWiFi2024, liangMmStressDistillingHuman2023, renGoPose3DHuman2022}. 
Many existing DWS models adapt generic deep learning architectures with domain-specific preprocessing. 
For example, Widar3.0 \cite{zhengZeroEffortCrossDomainGesture2019} introduces domain-independent features for gesture recognition; RFPose and RFPose3D \cite{zhao2018through, zhao2018rf} employ CNNs for radar-based pose estimation; and MoRe-Fi \cite{zhengMoReFiMotionrobustFinegrained2021} applies autoencoders for respiration monitoring. 
Other works address practical challenges such as domain adaptation \cite{zhangCrossSenseCrossSiteLargeScale2018, jiangEnvironmentIndependentDevice2018, xiaoCsiGANRobustChannel2019} and unsupervised representation learning \cite{chenMoViFiMotionrobustVital2021, songRFURLUnsupervisedRepresentation2022}. 

A smaller body of work explicitly incorporates RF signal characteristics into network design. 
STFNets \cite{yaoSTFNetsLearningSensing2019} and UniTS \cite{liUniTSShortTimeFourier2021} embed time--frequency transforms into neural pipelines, while RF-Net \cite{dingRFnetUnifiedMetalearning2020} and SLNet \cite{yangSLNetSpectrogramLearning2023} integrate frequency-domain operators and phase information. 
Despite these advances, most existing approaches remain fundamentally black-box: interpretability is limited to individual modules, and the overall architecture and learned representations lack unified physical or mathematical meaning.
In contrast, \sysname provides model-level interpretability by deriving the entire architecture and representation learning objective from a principled mathematical framework, marking a departure from empirically engineered DWS models.

\head{Interpretable deep learning}
Interpreting deep learning models has received growing attention across high-stakes domains such as healthcare, autonomous systems, and large-scale language models \cite{kimViTNeTInterpretableVision2022, zhangNestedHierarchicalTransformer2022, zhangInterpretableConvolutionalNeural2018, chen2018fsrnet, sabourDynamicRoutingCapsules2017, mongaAlgorithmUnrollingInterpretable2020, chenInfoGANInterpretableRepresentation2016, zhouObjectDetectorsEmerge2015, bachPixelWiseExplanationsNonLinear2015}.
Approaches include making features more understandable \cite{zhangNestedHierarchicalTransformer2022, zhangInterpretableConvolutionalNeural2018, chenThisLooksThat2019}, integrating domain knowledge into the architecture \cite{chen2018fsrnet, tofighi2019prior, yaoSTFNetsLearningSensing2019, liUniTSShortTimeFourier2021, dingRFnetUnifiedMetalearning2020, yangSLNetSpectrogramLearning2023}, or replacing deep layers with interpretable structures like decision trees \cite{kimViTNeTInterpretableVision2022, wuInterpretableObjectDetection2019}.
Others propose entirely new interpretable models, including capsule networks \cite{sabourDynamicRoutingCapsules2017}, unrolled networks \cite{mongaAlgorithmUnrollingInterpretable2020}, and InfoGAN \cite{chenInfoGANInterpretableRepresentation2016}. Of these, only the last category achieves \emph{full} interpretability.
Recently, CRATE \cite{yuWhiteBoxTransformersSparse2023a} advances this line by offering a fully interpretable transformer-based model, linking layer behavior to an explicit optimization objective. For a broader overview, see surveys \cite{liInterpretableDeepLearning2022, mongaAlgorithmUnrollingInterpretable2020, zhangVisualInterpretabilityDeep2018, chakrabortyInterpretabilityDeepLearning2017}.

\head{Complex-valued neural networks}
Complex-valued neural networks (CVNNs) directly model amplitude and phase, yielding more informative representations than real-valued models that discard or decouple these features \cite{trabelsiDeepComplexNetworks2018, monning2018evaluation, hirose2006complex}.
Foundational techniques such as Wirtinger calculus \cite{hirose2006complex} and complex normalization layers \cite{trabelsiDeepComplexNetworks2018} enable deeper CVNNs.
Recent work extends mainstream architectures like CNNs into the complex domain for applications in RF and medical imaging \cite{eilersBuildingBlocksComplexValued2023, virtueBetterRealComplexvalued2017a}. However, major challenges persist. First, adapting advanced architectures like transformers to complex signals remains underexplored. Second, training CVNNs is difficult due to unstable optimization dynamics \cite{wuComplexvaluedNeuronsCan2023}.
Fortunately, \sysname addresses the first challenge by presenting a theoretically derived complex transformer and demonstrating its effectiveness on wireless sensing tasks. 
Furthermore, its white-box nature provides a principled framework for analyzing training dynamics, offering a path toward solving the broader optimization problem and unlocking the full potential of complex-valued models.

\section{Conclusion}
\label{sec:conclusion}

This paper presents \sysname, the first mathematically grounded white-box framework for deep wireless sensing.
Built on the principle of complex sparse rate reduction, \sysname introduces a fully complex-valued architecture that extends the white-box transformer CRATE to RF signals.
With mathematically grounded complex self-attention and a principled residual MLP, together with Subspace Regularization, \sysname unifies complex-domain modeling and deep representation learning within a transparent, optimization-driven framework.
We evaluate \sysname on five datasets spanning three RF modalities and diverse sensing tasks, showing that it remains competitive with strong black-box baselines while retaining mathematical transparency.
Beyond accuracy, \sysname demonstrates that principled complex-domain design can deliver robust and practically useful performance in wireless sensing, albeit with task-dependent gains.
More broadly, \sysname points to a promising direction for learning-based wireless sensing, shifting from empirically engineered black boxes toward more interpretable and mathematically grounded deep sensing systems.
We hope this work lays a foundation for future white-box DWS models that integrate signal processing theory, deep learning, and human semantics, advancing trustworthy and deployable wireless sensing in everyday environments.

\begin{acks}
This work is supported by the Hong Kong RGC GRF under grant No. 17211725 and CRF under grant No. C5002-23Y. 
\end{acks}

%% file: appendix.tex
\appendix

\section{Detailed Derivation of \sysname}
\label{sec:app}
The following subsections elaborate on the theoretical derivations that form the foundation of the RF Self-Attention and RF-MLP modules.

\subsection{Detailed Derivation of the RF Self-Attention Module} 
\label{sec: gradient_decent_rc}

This subsection presents the complete derivation of the RF Self-Attention module, \ie, the approximate gradient descent step for minimizing $R^{c}(\mathbf{Z}^{l+1/2} \mid \mathbf{U}_{[K]}^{l})$. 
We begin by deriving the gradient of \(R^c(\mathbf{Z} \mid \mathbf{U}_{[K]})\) under the $\mathbb{C}\mathbb{R}$-calculus framework. 
Next, we approximate the gradient using the Von Neumann expansion together with the complex softmax function, thereby reducing computational complexity. 
Finally, we derive the formulation of the multi-head subspace self-attention used in \sysname.

\head{Gradient derivation} 
Since \(R^c(\mathbf{Z} \mid \mathbf{U}_{[K]})\) is non-holomorphic, we compute its gradient under the $\mathbb{C}\mathbb{R}$-calculus framework \cite{kreutz_delgadoComplexGradientOperator2009}. 
Specifically, recall the definition in \eqn\ref{eq:CRcZ}:
\[
R^c(\mathbf{Z} \mid \mathbf{U}_{[K]}) = \frac{1}{2} \sum_{k=1}^{K} \log \det \left( \mathbf{I} + \beta (\mathbf{U}_k^H \mathbf{Z})^H (\mathbf{U}_k^H \mathbf{Z}) \right),
\]
where \(\mathbf{Z} \in \mathbb{C}^{d \times N}\) denotes the feature representation matrix, 
\(\mathbf{U}_k \in \mathbb{C}^{d \times p}\) are orthonormal bases for \(k \in \{1, 2, \ldots, K\}\), 
and \(\beta > 0\) is a scaling factor. 

Because this is a non-constant, real-valued complex function, it does not satisfy the Cauchy–Riemann equations and is therefore not complex analytic, \ie, non-holomorphic. 
In gradient-based optimization, the relevant quantity is the conjugate gradient \cite{brandwood1983complex}, defined as
\[
\nabla_{\overline{Z}} \triangleq \left( \frac{\partial}{\partial \overline{z}_1}, \ldots, \frac{\partial}{\partial \overline{z}_n} \right)^{T},
\]
where $\overline{z}_k = x_k - i y_k$ for $k = 1, \ldots, n$, and $T$ denotes matrix transpose. 
The operator $\frac{\partial}{\partial \overline{z}_k}$ is given by
\[
\frac{\partial}{\partial \overline{z}_k} \triangleq \frac{1}{2} \left( \frac{\partial}{\partial x_k} + j \frac{\partial}{\partial y_k} \right).
\]

To this end, the gradient of \(R^c(\mathbf{Z} \mid \mathbf{U}_{[K]})\) involved in the optimization procedure of \sec\ref{sec:design} is defined as follows. 
First, let 
\[
h(W_k) = \log \det \left( \mathbf{I} + W_k \right), \quad W_k \in \mathbb{C}^{n \times n},
\]
and 
\[
g(\mathbf{Z} \mid \mathbf{U}_{k}) = \beta (\mathbf{U}_k^H \mathbf{Z})^H (\mathbf{U}_k^H \mathbf{Z}).
\] 
Then,
\[
R^c(\mathbf{Z} \mid \mathbf{U}_{[K]}) = \tfrac{1}{2} \sum_{k=1}^{K} (h \circ g)(\mathbf{Z} \mid \mathbf{U}_{k}).
\]

The gradient is therefore given by
\begin{align*}
\nabla R^c(\mathbf{Z} \mid \mathbf{U}_{[K]}) 
&= \nabla_{\overline{\mathbf{Z}}} R^c(\mathbf{Z} \mid \mathbf{U}_{[K]}) \\
&= \tfrac{1}{2} \sum_{k=1}^{K} \frac{\partial h(g)}{\partial \overline{\mathbf{Z}}} \\
&= \tfrac{1}{2} \sum_{k=1}^{K} \left( \frac{\partial h}{\partial g} \frac{\partial g}{\partial \overline{\mathbf{Z}}} + \frac{\partial h}{\partial \overline{g}} \frac{\partial \overline{g}}{\partial \overline{\mathbf{Z}}} \right) \\
&= \tfrac{1}{2} \sum_{k=1}^{K} \left( \big( \mathbf{I} + g(\mathbf{Z} \mid \mathbf{U}_{k}) \big)^{-1} \frac{\partial g}{\partial \overline{\mathbf{Z}}} + 0 \cdot \frac{\partial \overline{g}}{\partial \overline{\mathbf{Z}}} \right) \\
&= \tfrac{1}{2} \sum_{k=1}^{K} \beta \mathbf{U}_k \mathbf{U}_k^H \mathbf{Z} \big( \mathbf{I} + g(\mathbf{Z} \mid \mathbf{U}_{k}) \big)^{-1} \\
&= \tfrac{1}{2} \beta \sum_{k=1}^{K} \mathbf{U}_k \mathbf{U}_k^H \mathbf{Z} \left( \mathbf{I} + \beta (\mathbf{U}_k^H \mathbf{Z})^H (\mathbf{U}_k^H \mathbf{Z}) \right)^{-1}.
\end{align*}

Notably, the second equality follows from the chain rule in $\mathbb{C}\mathbb{R}$-calculus, while the fourth equality exploits
\[
\frac{\partial g}{\partial \overline{\mathbf{Z}}} 
= \left. \frac{\partial}{\partial \overline{\mathbf{Z}}} \Big( (\mathbf{U}_k^H \mathbf{Z})^H (\mathbf{U}_k^H \mathbf{Z}) \Big) \right|_{\mathbf{Z}=\text{const.}}.
\]

\head{Von Neumann approximation} 
For computational efficiency, we approximate the inverse 
\[
\left( \mathbf{I} + \beta (\mathbf{U}_k^H \mathbf{Z})^H (\mathbf{U}_k^H \mathbf{Z}) \right)^{-1}
\]
using the first-order term of its von Neumann expansion, which generalizes the geometric series to matrices.  

Specifically, in von Neumann algebra, if a matrix $A$ converges under some norm, then $A$ is invertible and
\[
A^{-1} = \sum_{n=0}^{\infty} (\mathbf{I} - A)^n.
\]

Applying this result, the gradient can be written as
\begin{align*}
\nabla_{\mathbf{Z}} R^c(\mathbf{Z} \mid \mathbf{U}_{[K]}) 
&= \tfrac{1}{2} \beta \sum_{k=1}^{K} \mathbf{U}_k \mathbf{U}_k^H \mathbf{Z} 
\left( \mathbf{I} + \beta (\mathbf{U}_k^H \mathbf{Z})^H (\mathbf{U}_k^H \mathbf{Z}) \right)^{-1} \\
&\approx \tfrac{1}{2} \beta \sum_{k=1}^{K} \mathbf{U}_k \mathbf{U}_k^H \mathbf{Z} 
\left( \mathbf{I} - \beta (\mathbf{U}_k^H \mathbf{Z})^H (\mathbf{U}_k^H \mathbf{Z}) \right).
\end{align*}

\head{Complex softmax} 
The term $(\mathbf{U}_k^H \mathbf{Z})^H (\mathbf{U}_k^H \mathbf{Z})$ in the gradient captures the auto-correlation among projected tokens in the subspace defined by $\mathbf{U}_{[K]}$. 
Since our objective is to associate tokens only within the same subspace, we introduce a complex softmax operation to indicate subspace membership.  

Several approaches have been proposed for extending the softmax operator to complex inputs. 
For $z \in \mathbb{C}$ and a real-valued softmax $\sigma$, one can, for instance, define $\sigma(|z|)$, 
$\tfrac{\sigma(\Re(z)) + \sigma(\Im(z))}{2}$, $\sigma(\Re(z)\Im(z))$, or $\sigma(\sigma(\Re(z) + \Im(z)))$, among others.  

In our model, however, we adopt a definition that preserves the internal geometry of the complex vector space \cite{scardapaneComplexValuedNeuralNetworks2020}. 
Specifically, we define the complex softmax function $\mathcal{S}(\cdot)$ as
\[
\mathcal{S}(z_i) = \frac{e^{\Re(z_i)^2 + \Im(z_i)^2}}{\sum_{j=1}^{K} e^{\Re(z_j)^2 + \Im(z_j)^2}},
\]
where $z_i$ denotes the $i$-th element of the token matrix $\mathbf{Z}$.

Finally, this yields 
\begin{align*}
&\nabla_{\mathbf{Z}} R^c(\mathbf{Z} \mid \mathbf{U}_{[K]}) \\ &\approx \frac{1}{2} \beta \sum_{k = 1}^{K} \mathbf{U}_k\mathbf{U}_k^H\mathbf{Z} \left(\mathbf{I} - \beta(\mathbf{U}_k^H\mathbf{Z})^H(\mathbf{U}_k^H\mathbf{Z}) \right) \\
&\approx \frac{1}{2} \beta \sum_{k = 1}^{K} \mathbf{U}_k\mathbf{U}_k^H\mathbf{Z} \\ &- \frac{1}{2} \beta^2 \sum_{k = 1}^{K} \mathbf{U}_k \left( (\mathbf{U}_k^H\mathbf{Z} ) \mathcal{S} \left((\mathbf{U}_k^H\mathbf{Z})^H(\mathbf{U}_k^H\mathbf{Z}) \right) \right) \\
&\approx \frac{1}{2} \beta \mathbf{Z} - \frac{1}{2} \beta^2 \sum_{k = 1}^{K} \mathbf{U}_k \left( (\mathbf{U}_k^H\mathbf{Z} ) \mathcal{S} \left((\mathbf{U}_k^H\mathbf{Z})^H(\mathbf{U}_k^H\mathbf{Z}) \right) \right).
\end{align*}

\head{$\textit{RF-MSSA}$ from the gradient descent step} 
We utilize an \textit{approximate} gradient descent step to minimize the coding rate $R^{c}(\mathbf{Z}^{l+1/2} \mid \mathbf{U}_{[K]}^{l})$ with learning rate $\kappa > 0$, namely,
\begin{align*}
Z^{l + 1/2} &= \mathbf{Z}^{l}- \kappa \nabla R^c(\mathbf{Z}^{l} \mid \mathbf{U}_{[K]}^{l}) \\
&= \mathbf{Z}^{l}- \kappa \frac{1}{2} \beta \mathbf{Z}^l \\  &+ \kappa \frac{1}{2} \beta^2 \sum_{k = 1}^{K} \mathbf{U}^l_k \left( ((\mathbf{U}^l_k)^H\mathbf{Z}^l ) \mathcal{S} \left(((\mathbf{U}^l_k)^H\mathbf{Z}^l)^H((\mathbf{U}^l_k)^H\mathbf{Z}^l) \right) \right) \\
&= (1- \kappa \beta /2) \mathbf{Z}^{l} + \kappa \beta/2 \textit{ RF-MSSA}(\mathbf{Z}^{l} \mid \mathbf{U}_{[K]}^{l}).
\end{align*}
Here, $\textit{RF-MSSA}$ is defined as
\[
    \textit{RF-MSSA}(\mathbf{Z}^{l} \mid \mathbf{U}_{[K]}^{l}) = \beta \left[ \mathbf{U}_1^{l}, \cdots, \mathbf{U}_K^{l} \right] 
    \begin{bmatrix}
    \textit{RF-SSA}(\mathbf{Z}^{l} \mid \mathbf{U}_1^{l}) \\
    \vdots \\
    \textit{RF-SSA}(\mathbf{Z}^{l} \mid \mathbf{U}_K^{l})
    \end{bmatrix},
\]
where each subspace-level attention term is given by
\[
\textit{RF-SSA}(\mathbf{Z}^{l} \mid \mathbf{U}_{k}^{l}) = \underbrace{{\mathbf{U}^{l}_k}^H\mathbf{Z}^{l}}_{\textbf{Value}} \mathcal{S}( \underbrace{{(\mathbf{U}^{l}_k}^H\mathbf{Z}^{l})^H}_{\textbf{Query}} \underbrace{{(\mathbf{U}^{l}_k}^H\mathbf{Z}^{l})}_{\textbf{Key}}).
\]

\subsection{Detailed RF-MLP Module Derivation}

The derivation of the RF-MLP module arises from the optimization problem
\[
\min_{\mathbf{Z}^{l+1}} \; -R(\mathbf{Z}^{l+1}) + \lambda \|\mathbf{Z}^{l+1}\|_0.
\]
To address this sparse program, we first relax it to a non-negative LASSO formulation \cite{yuWhiteBoxTransformersSparse2023a}. 
We then derive a quadratic upper bound of the smooth component, \ie, $-R(\mathbf{Z}^{l+1})$, within the $\mathbb{CR}$-calculus framework. 
Next, we minimize this quadratic upper bound and apply a proximal step to the non-smooth term $\lambda \|\mathbf{Z}^{l+1}\|_0$. 
Finally, by treating the metric tensor $\Omega_Z$ as a learnable weight matrix and relaxing its Hermitian and positive-definite constraints, we arrive at the RF-MLP module.  
\noindent
\textbf{Notation.} 
We use [$\mathbb{CR}, P_{x}, y$] to denote an equation referenced from the $\mathbb{CR}$-calculus paper \cite{kreutz_delgadoComplexGradientOperator2009}, located on page $x$ with equation number $y$. 
Similarly, [$CRATE, P_{x}, y$] indicates an equation referenced from \cite{yuWhiteBoxTransformersSparse2023a}.

\head{Alternative optimization program} 
Following the approach in CRATE \cite{yuWhiteBoxTransformersSparse2023a}, we relax the $l^0$ "norm" to the $l^1$ norm and impose a nonnegativity constraint. 
This converts the original optimization program \eqn\ref{eq:csrr_optimize_2} to the following program:
\begin{align*}
    \min_{f_2 \in \mathcal{F}_2} \mathbb{E}_{\mathbf{Z}^{l+1}} \left[ -R(\mathbf{Z}^{l+1}) + \lambda \|\mathbf{Z}^{l+1}\|_1 + \chi_{\{\mathbf{Z}^{l+1} \geq 0\}}(\mathbf{Z}^{l+1}) \right],
\end{align*}
where $\mathbf{Z}^{l+1} = f_2(\mathbf{Z}^{l+1/2})$, $\chi_{\{\mathbf{Z} \geq 0\}}$ denotes the characteristic function for the set of element-wise nonnegative matrices $\mathbf{Z}^{l+1}$, and $\lambda > 0$.

\head{Quadratic upper bound of $-R(\mathbf{Z})$} 
We start with the second-order Taylor expansion of $R(\mathbf{Z})$ in a neighborhood of the current iterate $\mathbf{Z}^{l+1/2}$. Due to the non-holomorphic property of $R(\mathbf{Z})=\frac{1}{2} \log \det (\mathbf{I} + \alpha \mathbf{Z}^H \mathbf{Z})$, we have its second-order Taylor expansion under the $\mathbb{CR}$-Calculus framework [$\mathbb{CR}, P_{38}, 99$]: 
\begin{align*}
R(\mathbf{Z}) &= R(\mathbf{Z}^{l+1/2}) + 2 \Re \left\{ \nabla_{\mathbf{Z}} R(\mathbf{Z}^{l+1/2})^H \Delta \mathbf{Z} \right\} \\
&+ \Re \left\{ \Delta \mathbf{Z}^H \mathcal{H}_{\mathbf{ZZ}} \Delta \mathbf{Z} + \Delta \mathbf{Z}^H \mathcal{H}_{\mathbf{Z}\overline{\mathbf{Z}}} \Delta \overline{\mathbf{Z}} \right\} + \text{h.o.t.}
\end{align*}
where $\nabla_{\mathbf{Z}} R(\mathbf{Z}^{l+1/2})$ denotes the gradient, $\Delta \mathbf{Z} = \mathbf{Z} - \mathbf{Z}^{l+1/2}$, and $\mathcal{H}_{\mathbf{zz}}$ and $\mathcal{H}_{\mathbf{Z}\overline{\mathbf{Z}}}$ are the Hessian matrices.

Moreover, with the definition of the inner product in $\mathbb{CR}$-Calculus framework [$\mathbb{CR}, P_{30}, .$], the second-order Taylor expansion with  integral remainder is
\begin{align*}
R(\mathbf{Z}) &= R(\mathbf{Z}^{l+1/2}) + \left\langle \nabla_{\mathbf{z}} R(\mathbf{Z}^{l+1/2}), \mathbf{Z} - \mathbf{Z}^{l+1/2} \right\rangle \\
&+\frac{1}{2} \int_0^1 (1 - t) \left( \left\langle \mathbf{Z} - \mathbf{Z}^{l+1/2}, \mathcal{H}_{\mathbf{ZZ}} (\mathbf{Z}_t) \left( \mathbf{Z} - \mathbf{Z}^{l+1/2} \right) \right\rangle \right . \\
&\left . + \left\langle \mathbf{Z} - \mathbf{Z}^{l+1/2}, \mathcal{H}_{\mathbf{Z}\overline{\mathbf{Z}}} (\mathbf{Z}_t) \left( \overline{\mathbf{Z}} - \overline{\mathbf{Z}}^{l+1/2} \right) \right\rangle \right) dt,
\end{align*}

We first derive an upper bound for the quadratic residual. 
By applying the Cauchy–Schwarz inequality, we obtain
\begin{align*}
&\left\langle \mathbf{Z} - \mathbf{Z}^{l+1/2}, \mathcal{H}_{\mathbf{ZZ}} (\mathbf{Z}_t) \left( \mathbf{Z} - \mathbf{Z}^{l+1/2} \right) \right\rangle \\
&\leq \sup_{t \in [0,1]} \left\| \mathcal{H}_{\mathbf{ZZ}}(\mathbf{Z}_t) \right\|_{\ell_2 \to \ell_2} \left\| \mathbf{Z} - \mathbf{Z}^{\ell + 1/2} \right\|_F^2 
\end{align*}
and 
\begin{align*}
&\left\langle \mathbf{Z} - \mathbf{Z}^{l+1/2}, \mathcal{H}_{\mathbf{Z}\overline{\mathbf{Z}}} (\mathbf{Z}_t) \left( \overline{\mathbf{Z}} - \overline{\mathbf{Z}}^{l+1/2} \right) \right\rangle \\
&\leq \sup_{t \in [0,1]} \left\| \mathcal{H}_{\mathbf{Z}\overline{\mathbf{Z}}}(\mathbf{Z}_t) \right\|_{\ell_2 \to \ell_2} \left\| \mathbf{Z} - \mathbf{Z}^{\ell + 1/2} \right\|_F^2 
\end{align*}

Next, we derive an upper bound for the operator norms, namely 
$\left\| \mathcal{H}_{\mathbf{ZZ}}(\mathbf{Z}_t) \right\|_{\ell_2 \to \ell_2}$ and $\left\| \mathcal{H}_{\mathbf{ZZ}}(\mathbf{Z}_t) \right\|_{\ell_2 \to \ell_2}$.

\noindent \ding{182} \textit{Upper bound $\left\| \mathcal{H}_{\mathbf{ZZ}}(\mathbf{Z}_t) \right\|_{\ell_2 \to \ell_2}$}: Based on the definition of the operator norm, we have
\[
\left\| \mathcal{H}_{\mathbf{ZZ}}(\mathbf{Z}_t) \right\|_{\ell_2 \to \ell_2} \triangleq \sup_{\| \Delta \| \leq 1} \left\| \mathcal{H}_{\mathbf{ZZ}}(\mathbf{Z}_t) \Delta \right\|_F \text{ with } \Delta \in \mathcal{C}^{d \times N}.
\]
To calculate the Hessian matrix, we use the method in \cite{yuWhiteBoxTransformersSparse2023a} [$CRATE, P_{59}, 127$]:
if \(\Delta\) is any matrix with the same shape as \(Z\) and \(t > 0\),
\[
\mathcal{H}_{\mathbf{ZZ}}(\mathbf{Z})(\Delta) = \left. \frac{\partial}{\partial t} \right|_{t=0} \left[ t \mapsto \nabla_{\mathbf{Z}} R \left(\mathbf{Z} + t\Delta, \overline{\mathbf{Z}} \right) \right],
\]
which holds since $R$ is smooth.  
Here, we write $\mathbf{Z}$ in place of $\mathbf{Z}_t$ to avoid confusion, and we adopt the conjugate coordinate representation $(\mathbf{Z}, \overline{\mathbf{Z}})$ for $R$.
Based on the definition of gradient in [$\mathbb{CR}, P_{21}, 56$], we have 
\begin{align*}
     &\nabla_{\mathbf{Z}} R \left(\mathbf{Z} + t\Delta, \overline{\mathbf{Z}} \right) \\
     &\triangleq \left( \frac{\partial R}{\partial \mathbf{Z}} \left(\mathbf{Z} + t\Delta, \overline{\mathbf{Z}} \right) \right)^H \\
     &= \left( \frac{\partial R}{\partial \overline{\mathbf{Z}}} \left(\mathbf{Z} + t\Delta, \overline{\mathbf{Z}} \right) \right)^T \\
     &= \frac{1}{2} \alpha \left( \mathbf{Z} + t\Delta \right) \left( \mathbf{I} + \alpha \mathbf{Z}^H \left(\mathbf{Z} + t\Delta \right) \right)^{-1} \\
     &= \frac{1}{2} \alpha \left( \mathbf{Z} + t\Delta \right) \left( \mathbf{I} + \alpha \mathbf{Z}^H\mathbf{Z} + \mathbf{Z}^H t\Delta \right)^{-1} \\
     &= \frac{1}{2} \alpha \left( \mathbf{Z} + t\Delta \right) \left( \mathbf{I} + \alpha t \left(\mathbf{I} + \alpha \mathbf{Z}^H\mathbf{Z} \right)^{-1} \mathbf{Z}^H\Delta \right)^{-1}\\
     &\qquad\cdot \left(\mathbf{I} + \alpha \mathbf{Z}^H\mathbf{Z} \right)^{-1} \\
     &= \frac{1}{2} \alpha \left( \mathbf{Z} + t\Delta \right) \sum_{k=0}^{\infty} \left(-\alpha t \left(\mathbf{I} + \alpha \mathbf{Z}^H\mathbf{Z} \right)^{-1} \mathbf{Z}^H\Delta \right)^{k} \\ &\qquad\cdot \left(\mathbf{I} + \alpha \mathbf{Z}^H\mathbf{Z} \right)^{-1},
\end{align*}
where \(\alpha = \frac{d}{n \varepsilon^2}\) is the coefficient in $R$ and the partial derivatives are defined as cogradient operator in [$\mathbb{CR}, P_{14}, 20$]. In addition, the first equality is due to that $R$ is real-valued [$\mathbb{CR}, P_{16}, 30$] and the last equality is base on the Neumann series.

We, then, calculate the derivative of the above result against $t$ and set $t=0$ later. Thus, we have
\begin{align*}
    &\mathcal{H}_{\mathbf{ZZ}}(\mathbf{Z})(\Delta) \\
    &= \left. \frac{\partial}{\partial t} \right|_{t=0} \left[ t \mapsto \nabla_{\mathbf{Z}} R \left(\mathbf{Z} + t\Delta, \overline{\mathbf{Z}} \right) \right] \\
    &= \frac{\alpha}{2} \Delta \left(\mathbf{I} + \alpha \mathbf{Z}^H \mathbf{Z}\right)^{-1} \\
    &\qquad- \frac{\alpha^2}{2} \mathbf{Z} \left( \mathbf{I} + \alpha \mathbf{Z}^H \mathbf{Z} \right)^{-1} \left(\mathbf{Z}^H \Delta \right) \left( \mathbf{I} + \alpha \mathbf{Z}^H \mathbf{Z} \right)^{-1} \\
    &= \frac{\alpha}{2} \left( \Delta - \alpha \mathbf{Z} \left( \mathbf{I} + \alpha \mathbf{Z}^H \mathbf{Z} \right)^{-1} \left(\mathbf{Z}^H \Delta \right) \right) \left( \mathbf{I} + \alpha \mathbf{Z}^H \mathbf{Z} \right)^{-1}.
\end{align*}

Given the above formulation, we can now establish an upper bound for $\left\| \mathcal{H}_{\mathbf{ZZ}}(\mathbf{Z}_t) \right\|_{\ell_2 \to \ell_2}$, which is given by: 
\begin{align*}
    &\left\| \mathcal{H}_{\mathbf{ZZ}}(\mathbf{Z}_t) \right\|_{\ell_2 \to \ell_2} \\
    &\triangleq \sup_{\|\Delta \| \leq 1} 
    \left\| \mathcal{H}_{\mathbf{ZZ}}(\mathbf{Z}_t)\Delta \right\|_F \\[4pt]
    &= \sup_{\|\Delta \| \leq 1} 
    \Big\| \tfrac{\alpha}{2} \big( \Delta 
        - \alpha \mathbf{Z}_t 
        (\mathbf{I} + \alpha \mathbf{Z}_t^H \mathbf{Z}_t )^{-1} 
        (\mathbf{Z}_t^H \Delta) \big) \\
    &\hspace{3.5cm} \cdot (\mathbf{I} + \alpha \mathbf{Z}_t^H \mathbf{Z}_t )^{-1} 
    \Big\|_F \\[4pt]
    &\leq \tfrac{\alpha}{2} \sup_{\|\Delta \| \leq 1} 
    \left\| \Delta (\mathbf{I} + \alpha \mathbf{Z}_t^H \mathbf{Z}_t )^{-1} \right\|_F \\
    &\quad + \tfrac{\alpha^2}{2} \sup_{\|\Delta \| \leq 1} 
    \Big\| \mathbf{Z}_t (\mathbf{I} + \alpha \mathbf{Z}_t^H \mathbf{Z}_t )^{-1} 
        (\mathbf{Z}_t^H \Delta) 
     \cdot (\mathbf{I} + \alpha \mathbf{Z}_t^H \mathbf{Z}_t )^{-1} 
    \Big\|_F \\[4pt]
    &\leq \tfrac{\alpha}{2} \sup_{\|\Delta \| \leq 1} \|\Delta\|_F
      + \tfrac{\alpha}{2} \sup_{\|\Delta \| \leq 1} \|\Delta\|_F \\[4pt]
    &= \alpha,
\end{align*}
where the first inequality is based on the Cauchy-Schwarz inequality, and the second one is based on the equations in [$CRATE, P_{60}, 134$] and [$CRATE, P_{61}, 141$].

\noindent \ding{183} \textit{Upper bound $\left\| \mathcal{H}_{\mathbf{Z}\overline{\mathbf{Z}}}(\mathbf{Z}_t) \right\|_{\ell_2 \to \ell_2}$}: Follow the same method above, we have
\begin{align*}
    &\left\| \mathcal{H}_{\mathbf{Z}\overline{\mathbf{Z}}}(\mathbf{Z}_t) \right\|_{\ell_2 \to \ell_2} \triangleq \sup_{\| \Delta \| \leq 1} \left\| \mathcal{H}_{\mathbf{Z}\overline{\mathbf{Z}}}(\mathbf{Z}_t) \Delta \right\|_F \\
    &= \sup_{\| \Delta \| \leq 1} \left\| \left. \frac{\partial}{\partial t^{\prime}} \right|_{t^{\prime}=0} \left[ t^{\prime} \mapsto \nabla_{\mathbf{Z}_t} R \left(\mathbf{Z}_t, \overline{\mathbf{Z}_t} + t^{\prime} \Delta \right) \right] \right\|_F \\
    &= \sup_{\| \Delta \| \leq 1} \left\| \frac{\alpha}{2} \mathbf{Z} \left( - \alpha \left(\mathbf{I} + \alpha \mathbf{Z}_t^{H} \mathbf{Z}_t \right)^{-1} \Delta^{\top} \mathbf{Z}_t \right) \left( \mathbf{I}+\alpha \mathbf{Z}_t^{H}\mathbf{Z}_t \right)^{-1}  \right\|_F \\
    &= \frac{\alpha^2}{2} \sup_{\| \Delta \| \leq 1} \left\| \mathbf{Z} \alpha \left(\mathbf{I} + \alpha \mathbf{Z}_t^{H} \mathbf{Z}_t \right)^{-1} \Delta^{\top} \mathbf{Z}_t \left( \mathbf{I}+\alpha \mathbf{Z}_t^{H}\mathbf{Z}_t \right)^{-1}  \right\|_F \\
    & \leq \frac{\alpha^2}{2} \sup_{\| \Delta \| \leq 1} \frac{1}{4 \alpha} \left\| \Delta \right\|_F \\
    &= \frac{\alpha}{8},
\end{align*}
where the last inequality is based on [$CRATE, P_{61}, 143$].

To this end, we have the quadratic upper bound of $-R(\mathbf{Z})$:
\begin{align*}
-R(\mathbf{Z}) &\leq -R(\mathbf{Z}^{l+1/2}) + \left\langle -\nabla_{\mathbf{z}} R(\mathbf{Z}^{l+1/2}), \mathbf{Z} - \mathbf{Z}^{l+1/2} \right\rangle \\
&\quad+\frac{1}{2} \int_0^1 (1 - t) \left(\alpha + \frac{\alpha}{8} \right) \left\| \mathbf{Z} - \mathbf{Z}^{l + 1/2} \right\|_F^2  dt.
\end{align*}

\head{Minimize the quadratic upper bound} Let $Q(\mathbf{Z} \mid \mathbf{Z}^{l+1/2})$ denotes the quadratic upper bound of $-R(\mathbf{Z})$ in a neighborhood of $\mathbf{Z}^{l+1/2}$, \ie,
\begin{align*}
&Q(\mathbf{Z} \mid \mathbf{Z}^{l+1/2}) \\
&= -R(\mathbf{Z}^{l+1/2}) + \left\langle -\nabla_{\mathbf{z}} R(\mathbf{Z}^{l+1/2}), \mathbf{Z} - \mathbf{Z}^{l+1/2} \right\rangle \\
& \quad+\frac{1}{2} \int_0^1 (1 - t) \left(\alpha + \frac{\alpha}{8} \right) \left\| \mathbf{Z} - \mathbf{Z}^{l + 1/2} \right\|_F^2  dt \\
&= -R(\mathbf{Z}^{l+1/2}) + \left\langle -\nabla_{\mathbf{z}} R(\mathbf{Z}^{l+1/2}), \mathbf{Z} - \mathbf{Z}^{l+1/2} \right\rangle \\
& \quad+\frac{9 \alpha}{16} \left\| \mathbf{Z} - \mathbf{Z}^{l + 1/2} \right\|_F^2  \\
&= -R(\mathbf{Z}^{l+1/2}) + \operatorname{Tr} \left( 2\Re \left\{-\nabla_{\mathbf{z}} R(\mathbf{Z}^{l+1/2})^{H} ( \mathbf{Z} - \mathbf{Z}^{l+1/2} )  \right\} \right) \\
& \quad +\frac{9 \alpha}{16} \operatorname{Tr} \left( \Re \left\{\left(\mathbf{Z} - \mathbf{Z}^{l + 1/2} \right)^{H} \left(\mathbf{Z} - \mathbf{Z}^{l + 1/2}\right)\right\} \right),
\end{align*} 
where the last equality is based on the definition of the inner product in the $\mathcal{CR}$-Calculus framework [$\mathbb{CR}, P_{30}, .$].

Using the $\mathbb{CR}$-gradient definition [$\mathbb{CR}, P_{21}, 56$] with column-normalized $Z^{l+1/2}$ (via RF-MSSA's normalization layer), we have:
\begin{align*}
    &\nabla_{\mathbf{z}} R(\mathbf{Z}^{l+1/2}) \\
    &\triangleq \Omega_Z^{-1} \left( \frac{\partial R}{\partial \mathbf{Z}} \left(\mathbf{Z}^{l+1/2}, \overline{\mathbf{Z}}^{l+1/2} \right) \right)^H \\
     &= \Omega_Z^{-1} \left( \frac{\partial R}{\partial \overline{\mathbf{Z}}} \left(\mathbf{Z}^{l+1/2}, \overline{\mathbf{Z}}^{l+1/2} \right) \right)^T \\
     &= \Omega_Z^{-1} \frac{1}{2} \alpha \mathbf{Z} \left( \mathbf{I} + \alpha (\mathbf{Z}^{l + 1/2})^H \mathbf{Z}^{l + 1/2} \right)^{-1} \\
     &= \Omega_Z^{-1} \frac{1}{2} \alpha \mathbf{Z}^{l + 1/2} \left( \mathbf{I} + \alpha \mathbf{I} \right)^{-1}\\
     &= \frac{\alpha}{2 (1 + \alpha)} \Omega_Z^{-1} \mathbf{Z}^{l + 1/2},
\end{align*}
where $\Omega_z$ is a hermitian, positive-definite $d \times d$ metric tensor $\Omega_Z = \Omega_Z^H > 0$.

We now minimize the preceding quadratic upper bound $Q(\mathbf{Z} \mid \mathbf{Z}^{l+1/2})$ as a function of $Z$. Moreover, based on the conditions of the stationary points of real-valued functionals in the $\mathcal{CR}$-Calculus framework [$\mathcal{CR}, P_{17}, 37$], we first calculate the partial derivative of $Q((\mathbf{Z}, \overline{\mathbf{Z}}) \mid \mathbf{Z}^{l+1/2})$ against $Z$ where we use the conjugate coordinates. Then, obtain the optimal $\mathbf{Z}_{opt}$ by setting the partial derivative equal to $0$. Specifically, we have
\begin{align*}
    &\frac{\partial Q(\mathbf{Z} \mid \mathbf{Z}^{l+1/2})}{\partial Z} = \left. \frac{\partial Q((\mathbf{Z}, \overline{\mathbf{Z}}) \mid \mathbf{Z}^{l+1/2})}{\partial Z} \right|_{\overline{Z} = \text{const.}} \\
    &= \operatorname{Tr} \left( 2\Re \left\{-\nabla_{\mathbf{z}} R(\mathbf{Z}^{l+1/2})^{H}  \right\} \right) \\&+\frac{9 \alpha}{16} \operatorname{Tr} \left( \Re \left\{ \left(\mathbf{Z} - \mathbf{Z}^{l + 1/2} \right)^{H} \right\} \right).
\end{align*}

Thus, $\mathbf{Z}_{opt}$ can be obtained by solving the following equation:
\[
2\nabla_{\mathbf{z}} R(\mathbf{Z}^{l+1/2}) = \frac{9 \alpha}{16} \left(\mathbf{Z}_{opt} - \mathbf{Z}^{l + 1/2} \right).
\]

Substituting the result of the $\nabla_{\mathbf{z}} R(\mathbf{Z}^{l+1/2})$ into the equation, we have the final result as:
\[
\mathbf{Z}_{opt} = \left( 1+ \frac{16}{9(1+ \alpha)} \Omega_{Z} \right) \mathbf{Z}^{l + 1/2}.
\]

\head{Solving the non-smooth part $\lambda \|\mathbf{Z}^{l+1}\|_0$} As shown above, the non-smooth part is relaxed to \(\lambda \|\mathbf{Z}^{l+1}\|_1 + \chi_{\{\mathbf{Z}^{l+1} \geq 0\}}(\mathbf{Z}^{l+1})\). Following the method in \cite{yuWhiteBoxTransformersSparse2023a}, the proximal operator of the sum of \(\chi_{\{Z \geq 0\}}\) and \(\lambda \|\cdot\|_1\) is a one-sided soft-thresholding operator:
\[
\text{prox}_{\chi_{\{z \geq 0\}} + \lambda \|\cdot\|_1}(Z) = \max \{ Z - \lambda \mathbf{1}, \mathbf{0} \},
\]
where the maximum is applied element-wise. Moreover, one step of proximal majorization-minimization takes the form:
\[
Z^{l + 1} = \text{ReLU} \left( \left( 1 + \frac{16}{9(1 + \alpha)} \Omega_{Z} \right) \mathbf{Z}^{l + 1/2} - \frac{16 \lambda}{9 \alpha} \mathbf{1} \right),
\]
where we adopt the complex ReLU function \(\text{ReLU}\) to handle the complex input.

\head{$\textit{RF-MLP}$ module}
Finally, we derive the \(\textit{RF-MLP}\) operator for the MLP module in \sysname.

By treating the metric tensor $\Omega_Z$ as a learnable weight matrix and relaxing the constraints that it must be Hermitian and positive definite, we obtain the following form of the \textit{RF-MLP} operator:
\[
\textit{RF-MLP}(\mathbf{Z}^{l + 1/2}) = \text{ReLU} \left( \mathbf{Z}^{l + 1/2} + \eta \Omega_{Z} \mathbf{Z}^{l + 1/2} - \eta \lambda \mathbf{1} \right),
\]
where $\eta = \frac{16}{9(1 + \alpha)}$ is the learning rate.  
This operator is directly derived from a proximal majorization-minimization formulation.  
Importantly, the presence of the skip connection is not heuristic; it emerges naturally from the derivation, making the formulation of \textit{RF-MLP} mathematically complete.

However, if we follow the method in \cite{yuWhiteBoxTransformersSparse2023a}, which introduces an orthogonal dictionary $D \in \mathbb{C}^{d \times d}$ into the proximal majorization-minimization result and uses the fact that $R(DZ) = R(Z)$ for any $Z$, we arrive at:
\[
\mathbf{Z}^{l + 1} = \text{ReLU} \left( \mathbf{D} \mathbf{Z}^{l + 1/2} + \eta \mathbf{D} \mathbf{Z}^{l + 1/2} - \eta \lambda \mathbf{1} \right),
\]
where the skip connection is absent, leaving a gap between this form and the ISTA block used in CRATE.
Considering the complex domain, the original ISTA block (which is defined without direct derivation) can be extended to handle complex inputs. This leads to the $\textit{RF-ISTA}(\mathbf{Z}^{l + 1/2})$ operator:
\[
 \text{ReLU} \left( \mathbf{Z}^{l + 1/2} + \eta \mathbf{D}^H (\mathbf{Z}^{l + 1/2} - \mathbf{D} \mathbf{Z}^{l + 1/2}) - \eta \lambda \mathbf{1} \right).
\]
This form resembles the structure of a proximal gradient step and introduces a correction term based on the residual in the dictionary domain.  
However, it still lacks a formal derivation, and the absence of the skip connection limits its completeness.

In contrast, the \textit{RF-MLP} operator derived in \sysname provides a fully interpretable and mathematically grounded formulation. The inclusion of the skip connection is derived, not assumed, ensuring consistency with the underlying optimization principles.

\head{Complex ReLU}
We also require a version of ReLU that is compatible with complex-valued input. 
Again, several viable options already exist; for $z \in \mathbb{C}$ and real-valued ReLU $\rho$, cReLU outputs $\rho(\Re(z)) + i * \rho(\Im(z))$, modReLU \cite{arjovskyUnitaryEvolutionRecurrent2016a} outputs $\rho(|z|+b)*z/|z|$, and zReLU \cite{gubermanComplexValuedConvolutional2016} outputs $z$ if $\Re(z), \Im(z) > 0$ and 0 otherwise. Complex cardioid \cite{virtueBetterRealComplexvalued2017} outputs $(1 + \cos \varphi_z) * z/2$, which attenuates magnitude based on phase while preserving phase information. For real-valued inputs, this simplifies to the ReLU function.